%% file: main.tex
\documentclass[11pt]{book}
\usepackage[square,numbers]{natbib}
\usepackage[margin=1in]{geometry}

\usepackage{graphicx}
\usepackage[english]{babel}
\graphicspath{{images/}}
\usepackage{amssymb}
\usepackage{amsthm}
\usepackage{graphics}
\usepackage{adjustbox}
\usepackage{longtable}
\usepackage{amsmath}
\DeclareMathOperator*{\argmax}{argmax}
\DeclareMathOperator*{\argmin}{argmin}
\usepackage{amstext}
\usepackage{appendix}
\usepackage[nottoc]{tocbibind}
\usepackage{engrec}
\usepackage{microtype}
\usepackage{enumerate}
\usepackage{marginnote}
\usepackage{multirow}
\usepackage{fancyhdr}
\usepackage{setspace}
\usepackage{bm}
\usepackage[utf8x]{inputenc}
\usepackage{hyperref}
\usepackage{listings}
\usepackage[rightcaption]{sidecap}
\usepackage{subcaption}
\usepackage{wrapfig}
\usepackage{float}
\usepackage{tcolorbox}
\usepackage[ruled,vlined]{algorithm2e}
\usepackage{emptypage}
\newenvironment{abstract}
  {\cleardoublepage\thispagestyle{empty}
   \vspace*{\stretch{1}}
   \begin{center}\textbf{\abstractname}\end{center}\bigskip}
  {\par\vspace*{\stretch{2}}\cleardoublepage}

\begin{document}
    \begin{titlepage}
            
        \noindent
        \begin{minipage}[t]{0.19\textwidth}
            \vspace{-4mm}{\includegraphics[scale=1.15]{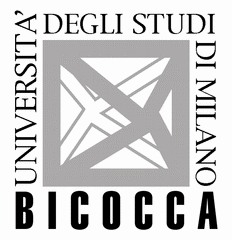}}
        \end{minipage}
        \begin{minipage}[t]{0.81\textwidth}
        {
            \setstretch{1.42}
                {\textsc{Università degli Studi di Milano - Bicocca}} \\
                \textbf{Scuola di Scienze} \\
                \textbf{Dipartimento di Informatica, Sistemistica e Comunicazione} \\
                \textbf{Corso di Laurea Magistrale in Informatica} \\
                \par
        }
        \end{minipage}
            
    	\vspace{35mm}
            
    	\begin{center}
            {\Huge{
                    \setstretch{1.2}
                    \textbf{Multi-Task Learning on Networks}
                    \par
                }
            }
        \end{center}
            
        \vspace{40mm}
    
        \noindent
            {\large \textbf{Relatore:} Prof. Antonio Candelieri} \\
    
        \noindent
            {\large \textbf{Co-relatore:} Prof. Francesco Archetti}
            
        \vspace{20mm}
    
        \begin{flushright}
            {\large \textbf{Relazione della prova finale di:}} \\
            \large{Andrea Ponti} \\
            \large{Matricola 816311} 
        \end{flushright}
            
        \vspace{30mm}
        \begin{center}
            {\large{\bf Anno Accademico 2020-2021}}
        \end{center}
    
        \restoregeometry
            
    \end{titlepage}
\clearpage
\thispagestyle{empty}
\frontmatter
\input{chapters/abstract.tex}
\tableofcontents
\listoffigures
\listoftables
\input{chapters.tex}
\bibliographystyle{unsrtnat}
\bibliography{bibliography}
\end{document}

%% file: chapters/abstract.tex
\begin{abstract}
The multi-task learning (MTL) paradigm can be traced back to an early paper of Caruana (1997) in which it was argued that data from multiple tasks can be used with the aim to obtain a better performance over learning each task independently. The rationale underlying this approach is that strong dependencies are ``hidden'' among seemingly unrelated tasks due to the shared data generating process. A natural way is to design a set of parametrized hypotheses that share some parameters across tasks which are learned solving an optimization problem that minimizes a weighted sum of the empirical risk for each task. Multi-task learning is a very common situation. For instance, in a recommender system, not only the accuracy of the rating prediction, but also the novelty and coverage of the recommendations should be optimized. Most recent Machine Learning (ML) applications require to optimize the ML algorithm’s hyperparameters not just for accuracy but also for fairness, interpretability, and energy consumption. For instance, tuning a Deep Neural Network’s hyperparameters depends on accuracy but also on latency and deployability on specific device (i.e., limited hardware resources of the target device on which making inference). Minimizing independently the empirical risks related to different tasks is the correct solution only when tasks are not competing with each other, which is rarely the case. ``Naive'' solutions like a linear combination of the single tasks are computationally expensive and lack credible metrics for evaluating the quality of the results. A solution of MTL with conflicting objectives requires modelling the trade-off among them which is generally beyond what a straight linear combination can achieve. Although constrained approaches, which optimize one single task while accounting the others as constraints, have been recently and successfully proposed to overcome these drawbacks in implementing specific real-life applications, a theoretically principled and computationally more effective strategy is finding solutions which are not ``dominated'' by others as it is addressed in the Pareto analysis. Multi-objective optimization problems arising in the multi-task learning context have specific features and require ad hoc methods. The analysis of these features, also in some specific instances, and the proposal of a new computational approach represent the focus of this work. Multi-objective evolutionary algorithms (MOEAs), specifically Non Sorting Dominated Genetic Algorithms (NSGAs) can easily include the concept of dominance and therefore the Pareto analysis. The major drawback of MOEAs is a low sample efficiency with respect to function evaluations which makes them hardly feasible when the evaluation of the objective functions is computationally very expensive. The key reason for this drawback is that most of the evolutionary approaches do not use models (surrogate models or metamodels) for approximating the objective function and therefore cannot make predictions over new candidate solutions. Bayesian Optimization (BO) takes a radically different approach based on a surrogate model, usually probabilistic, such as a Gaussian Process (GP). 
\newpage
\noindent
Most multi-objective BO approaches maintain different GPs, one for each task or objective: in general, the tasks show some underlying structure and cannot be treated as unrelated objectives. By making use of this structure, one might benefit significantly by learning the tasks simultaneously as opposed to learning them independently. In this thesis a different approach is considered. The solutions in the \textit{Input Space} are represented as probability distributions encapsulating the knowledge contained in the function evaluations. The focus is on discrete distributions and in particular histograms. These distributions are analyzed according to their distance. Among several distances the Wasserstein (WST) distance has been used. In this space of probability distributions, endowed with the metric given by the Wasserstein distance, a new algorithm MOEA/WST can be designed in which the model is not directly on the objective function but in an intermediate \textit{Information Space} where the objects from the input space are mapped into histograms and the genetic operators are built on the WST distance. Computational results show that both the sample efficiency and the quality of the Pareto set provided by MOEA/WST, as measured by the Hypervolume metric, are significantly better than in the standard MOEA implementation.
\end{abstract}

%% file: chapters.tex
\input{chapters/chapter01.tex}
\input{chapters/chapter02.tex}
\input{chapters/chapter03.tex}
\input{chapters/chapter04.tex}
\input{chapters/chapter05.tex}
\input{chapters/chapter06.tex}
\input{chapters/chapter07.tex}
\input{chapters/chapter08.tex}
\input{chapters/chapter09.tex}
\input{chapters/chapter10.tex}
\input{chapters/appendix.tex}

%% file: chapters/chapter01.tex
\mainmatter
\chapter{Introduction}
\label{ch:01}

In this chapter the Multi-Task Learning paradigm is introduced with a particular focus to optimization problems. In the following the main motivations are explained that lead to Multi-Objective Optimization and some of the principal strategies for solving it.

\section{Motivations}
The multi-task learning (MTL) paradigm can be traced back to an early paper of Caruana \cite{DBLP:journals/ml/Caruana97} in which it was argued that data from multiple tasks can be used with the aim to obtain a better performance over learning each task independently. The rationale underlying this approach is that strong dependencies are ``hidden'' among seemingly unrelated tasks due to the shared data generating process. A natural way is to design a set of parametrized hypothesis that share some parameters across tasks which are learned solving an optimization problem that minimizes a weighted sum of the empirical risk for each task. Multi-task learning is a very common situation. For instance, in a recommender system, not only the accuracy of the rating prediction, but also the novelty and coverage of the recommendations should be optimized. In general, hyperparameters should be optimized not just for accuracy but also for fairness, interpretability \cite{DBLP:conf/icml/AgarwalDW19} and energy consumption \cite{DBLP:conf/acl/StrubellGM19, DBLP:journals/soco/CandelieriPA21}. The mitigation of the ``contamination'' risk in networks, both physical and informational, depends both on detection time and its variance \cite{DBLP:conf/gecco/CandelieriPA21}. Tuning a Deep Neural Network’s hyperparameters depend on accuracy but also on latency and deployability \cite{DBLP:conf/icann/PeregoCAP20, DBLP:journals/mam/LoniSZDS20}. \\
Minimizing independently the empirical risks related to different tasks is the correct solution only when tasks are not competing with each other, which is rarely the case. ``Naive'' solutions like a linear combination of the single tasks are computationally expensive and lack credible metrics for evaluating the quality of the result. A solution of MTL with conflicting objectives requires modelling the trade-off between them which is generally beyond what a straight linear combination can achieve. A theoretically principled and computationally more effective strategy is finding solutions which are not ``dominated'' by others as it is addressed in the Pareto analysis. This solution has been recently advocated in several papers \cite{DBLP:conf/nips/SenerK18,DBLP:conf/nips/LinZ0ZK19}. \\
In this thesis the objective of multi-task learning is cast in terms of finding Pareto optimal solutions. The problem of finding Pareto optimal solutions given multiple criteria is called Multi-Objective Optimization (MOO). Multi-objective optimization problems arising in the multi-task learning context have specific features and require ad hoc methods \cite{DBLP:conf/icml/ZuluagaSKP13, DBLP:journals/jmlr/ZuluagaKP16, DBLP:conf/icml/ShahG16}. The analysis of these features, also in some specific instances, and the proposal of a new computational approach represent the focus of this work. \\
There are three main strategies to deal with multi-objective optimization problems occurring in MTL. \\
The first are gradient based methods \cite{DBLP:conf/nips/SenerK18,DBLP:journals/corr/abs-2010-06313}. These algorithms are quite effective but require the computation of gradients for each task which is not an easy task since the loss functions are usually black box and multimodal. \\
An alternative are evolutionary algorithms (EAs). EAs have developed over the last two decades along different approaches, which will be analyzed in Chapter \ref{ch02:pareto_evolutionary}. Their set-up is relatively simple, there are many software resources and do not require derivative information. Moreover, multi-objective evolutionary algorithms (MOEAs), specifically Non-dominated Sorting Genetic Algorithms (NSGAs) can easily include the concept of dominance and therefore the Pareto analysis. The major drawback of MOEAs is a low sample efficiency with respect to function evaluations which makes them hardly feasible when the evaluation of the objective functions is computationally very expensive, as it happens in machine learning problems in the case of large datasets and in simulation and optimization problems which is often the case in real word application as will be shown in Chapter \ref{ch04:instances}. The key reason for this drawback is that most of the evolutionary approaches do not use models (surrogate models or metamodels) for the objective function and therefore cannot make predictions over new candidate solutions. \\
Bayesian Optimization (BO) takes a radically different approach based on a surrogate model usually a Gaussian Process \cite{archetti2019bayesian, DBLP:journals/corr/abs-1807-02811, DBLP:conf/nips/BelakariaDD19, DBLP:journals/ijon/Garrido-Merchan19, DBLP:conf/gecco/RahatEF17}. Most multi-objective BO approaches maintain different Gaussian Processes (GPs), one for each task or objective: in general, the tasks show some underlying structure and cannot be treated as unrelated objects. By making use of this structure, one might benefit significantly by learning the tasks simultaneously as opposed to learning them independently. \\
An important line of research has been investigating the use of metamodels in MOEAs using in particular neural networks and gaussian processes. An early method is ParEGO \cite{DBLP:journals/tec/Knowles06, DBLP:journals/tec/ZhangLTV10} which use the uncertainty estimation allowed by GP to manage the exploration/exploitation (or diversification/intensification) trade-off. The importance of this issue, which is at the very basis of BO, has been recently recognized also in MOEAs in \cite{DBLP:journals/isci/ZhangSLZZ19}. \\
In this thesis a different approach is considered. The solutions in the input space are represented as probability distributions encapsulating the knowledge containing the function evaluations. The focus is on discrete distributions and in particular histograms. These distributions are analyzed according to their distance. Among several distances the Wasserstein (WST) distance has been used. In this space of probability distributions, endowed with the metric given by the Wasserstein distance, a new algorithm MOEA/WST can be designed in which the model is not directly on the objective function but in an intermediate information space where the objects from the input space are mapped into histograms and the genetic operators are built on the WST distance between histograms. Computational results show that both the sample efficiency and the quality of the Pareto set, as measured by the Hypervolume, are significantly better than in the standard MOEAs.

\section{Structure of the thesis}
The content of this work is organized as follows. \\
Chapter \ref{ch02:pareto_evolutionary} ``Pareto Analisys and Evolutionary Learning'' is devoted to the Pareto model and to the basic methods in MOEAs, analysing their structures and performance metric. \\
Chapter \ref{ch03:bo} ``Bayesian Optimization'' provides background material about Bayesian optimization methods focusing on their basic components: the surrogate model (metamodel) based on Gaussian processes and the acquisition function. \\
Chapter \ref{ch04:instances} ``Instances of Multi-Task Learning on Networks'' outlines the real world problems which are instances of learning on networks and have been inspirational for this work. They are contamination detection and resilience assessment in a Water Distribution Network (WDN), detection of fake-news in the blogosphere and recommender systems. These problems have a shared structure and have been targeted for computational experiments. All the problems considered are NP-hard so that solution methods are approximate metaheuristics. \\
Chapter \ref{ch05:wst} ``The Wasserstein Distance'' introduces the key concept in this work: solutions in the input space are mapped into a space of probability distributions. Among several distances between distributions, the Wasserstein distance is introduced. The focus is on discrete distributions and in particular histograms. \\
Chapter \ref{ch06:moeawst} ``MOEA with Wasserstein'' analyses MOEA/WST, an entirely new concept in which the model is not on the objective function but in an intermediate information space where the objects from the input space are mapped into histograms and the genetic operators are built on the WST distance between histograms. \\
Chapter \ref{ch07:water} ``Water Distribution Networks'' reports the experimental setting in the WDN target problems, a new data structure and the computational results. \\
Chapter \ref{ch08:recommender} ``Recommender Systems'' reports the experimental setting in the recommender target problems, and the computational results of MOEA/WST. \\
Chapter \ref{ch09:sw} ``Software resources'' describes the software used in this work. \\
Chapter \ref{ch10:conclusion} ``Conclusions'' contains a critical evaluation of the results along with indications of perspectives for future work.

\section{Contributions}
One of the key contributions of this work is the definition of a mapping from the search space (where each solution can be represented by a real, integer or binary vector) into an information space whose elements are probability distributions. This characterization of an information space enables the introduction of probabilistic distances, in particular the Wasserstein distance. In this way, it is possible to compute the distance between elements in the information space which are represented as histograms. \\
Another contribution of this thesis is the introduction of a novel data structure for archiving results of simulation in an efficient way and monitoring dynamical processes in networks. \\
The overall contribution is the formulation of a new evolutionary method called MOEA/WST in which combination operators are enabled by WST distance. Finally, it is proposed a critical analysis of the computational results obtained on target problems with MOEA/WST and other standard MOEAs, namely NSGA-II and ParEGO.

%% file: chapters/chapter02.tex
\chapter{Pareto Analisys and Evolutionary Learning}
\label{ch02:pareto_evolutionary}
Evolutionary learning refers to a set of algorithms and learning methods which draw inspiration from the process of natural evolution. The fundamental metaphor relates to an environment which is filled with a population of individuals that strive for survival and reproduction. The fitness of these individuals is determined by the environment and relates to how well they succeed in achieving their goals. In other words, it represents their chances of survival and of multiplying. In the context of learning, the environment is represented by the problem that has to be solved and the data available. The individuals refer to a set of candidate solutions and their quality determines the chance that they will be kept and used as seed for generating further candidate solutions. \\
This chapter considers the particular case of multi-objective optimizations and describes the two main evolutionary strategies used to solve this kind of problems.

\section{Pareto analysis}
\begin{figure}[h]
\centering
    \begin{subfigure}{.49\textwidth}
        \centering
        \includegraphics[width=1\linewidth]{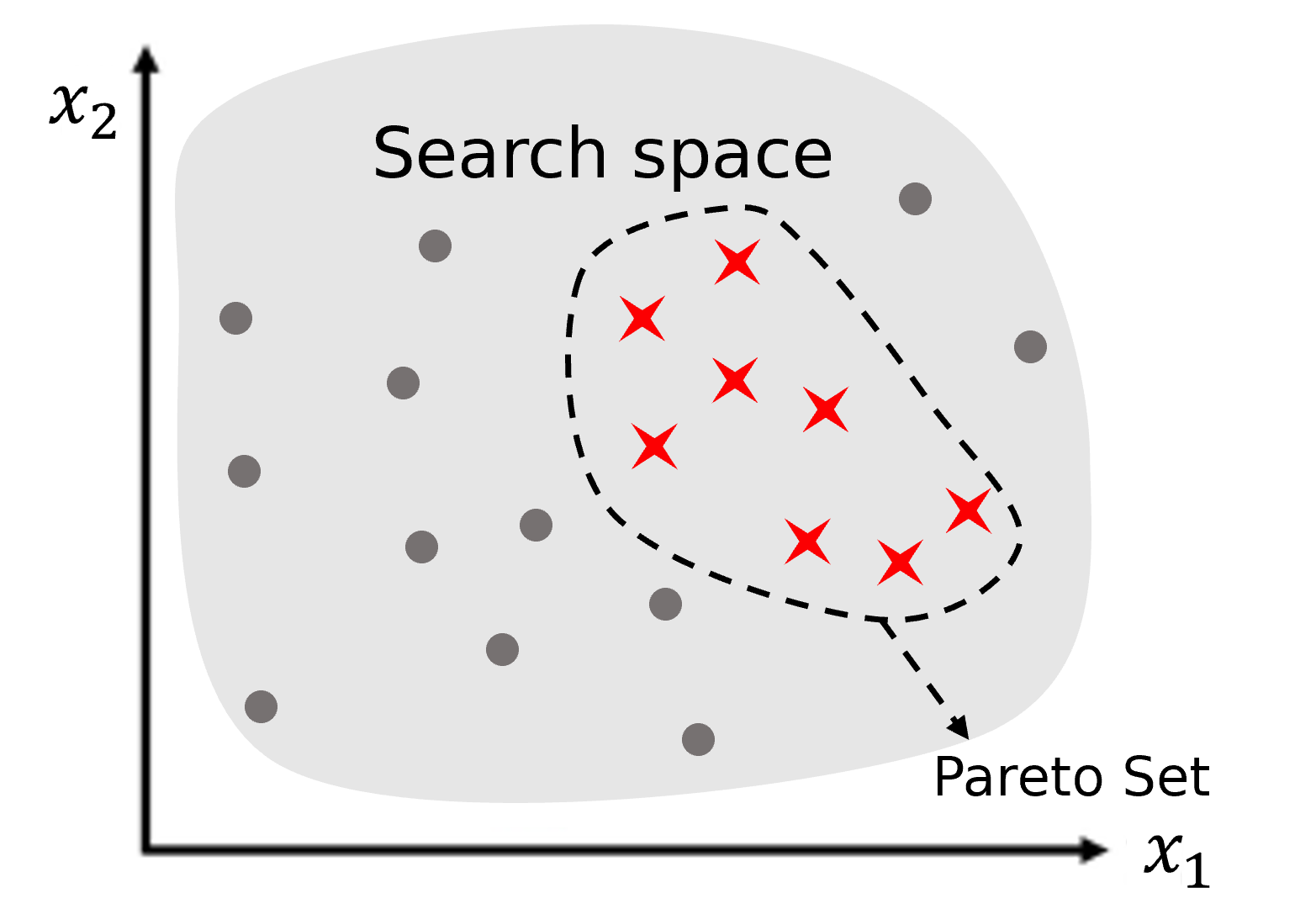}
        \caption{The search space with the Pareto set.}
        \label{fig:ch02_ps}
    \end{subfigure}
    \begin{subfigure}{.49\textwidth}
        \centering
        \includegraphics[width=1\linewidth]{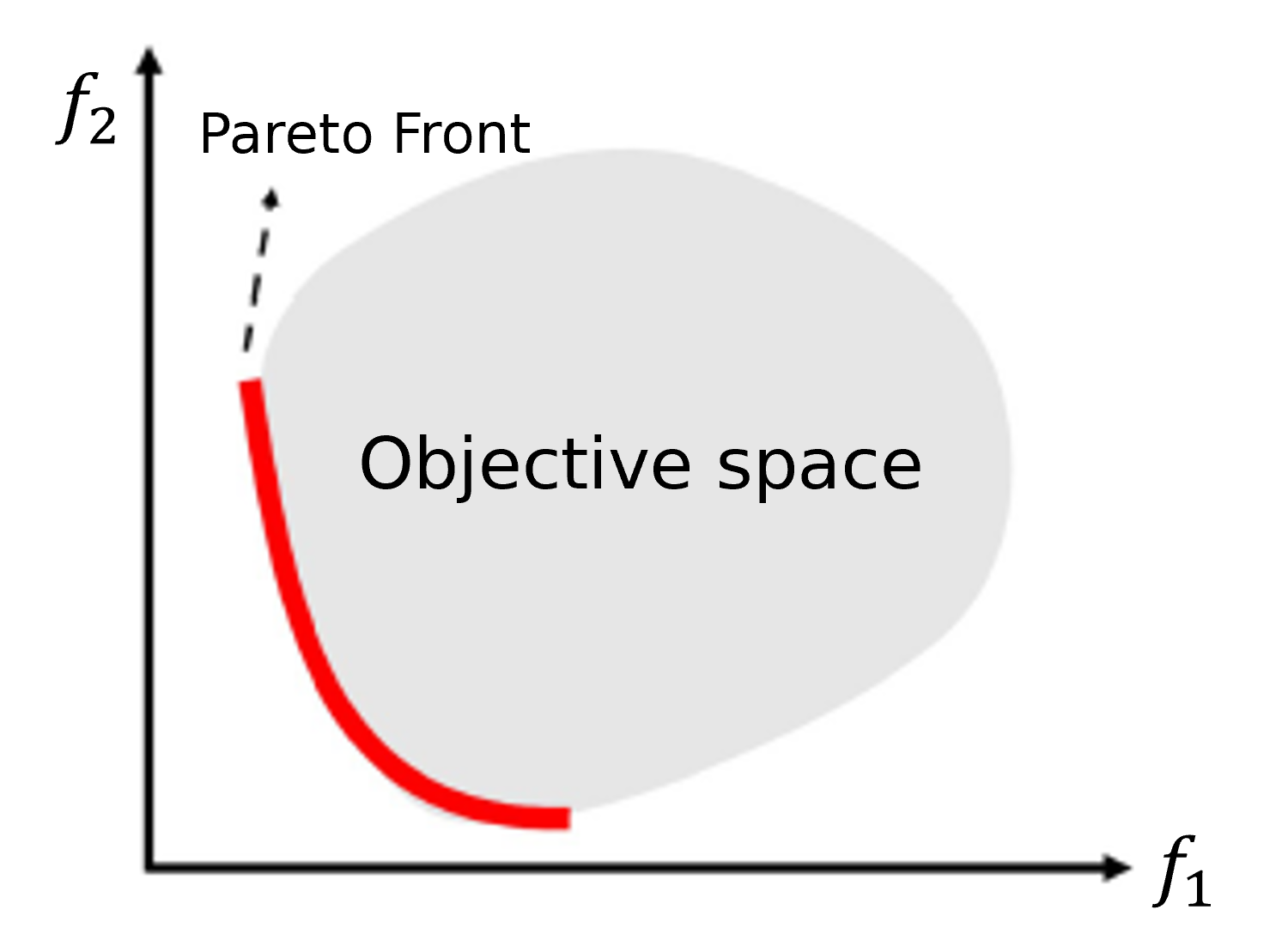}
        \caption{The objective space with the Pareto front.}
        \label{fig:ch02_pf}
    \end{subfigure}
\caption{An example of Pareto Set and Pareto Front.}
\label{fig:ch02_search_obj_space}
\end{figure}

\begin{figure}[h]
    \centering
    \includegraphics[scale = 0.7]{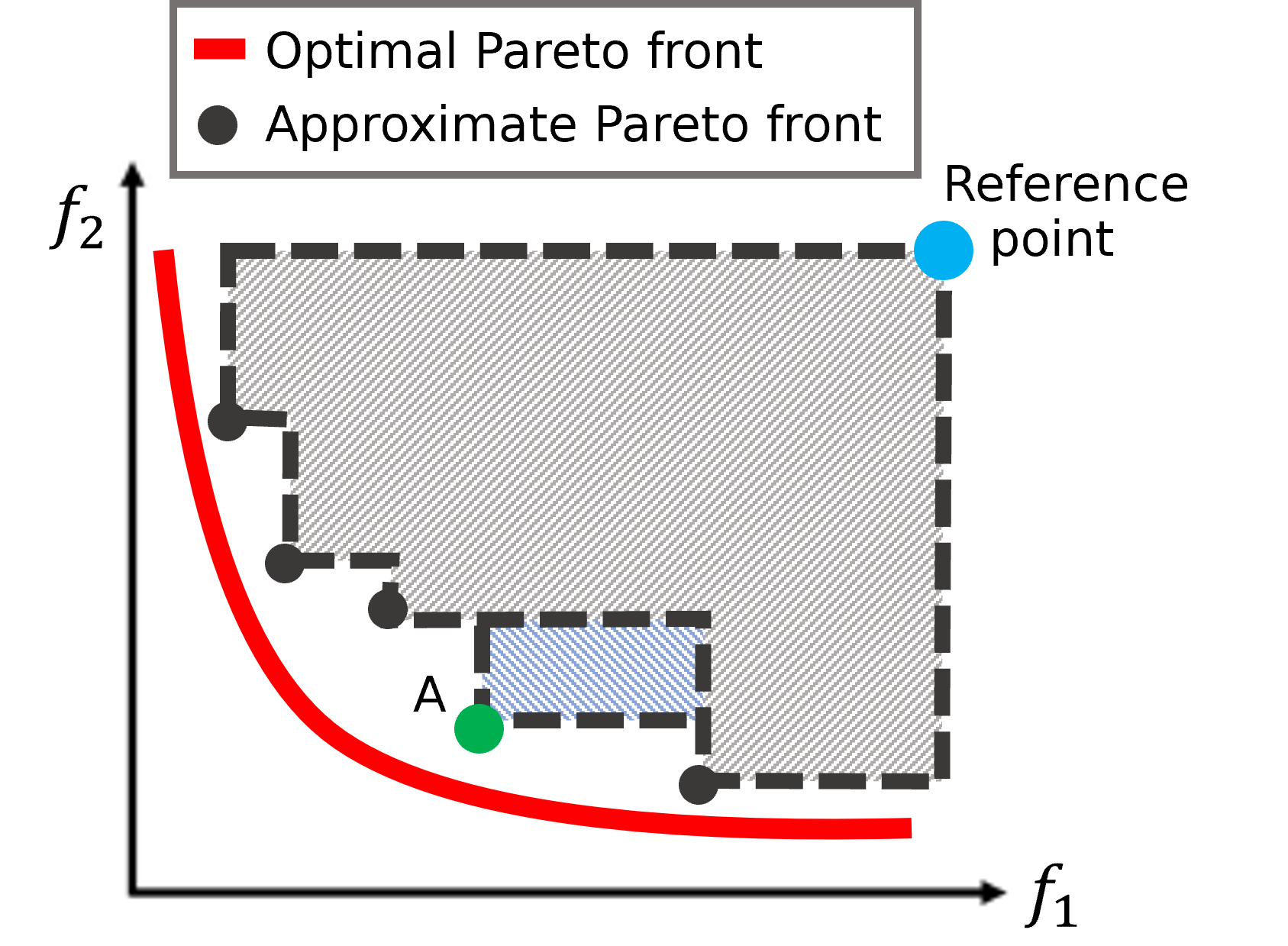}
    \caption{An example of hypervolume. Adding the point $A$ to the approximate Pareto front, leads to an improvement of the hypervolume.}
    \label{fig:ch02_hypervolume}
\end{figure}

Multi-Objective Optimization problem (MOO) can be stated as follows (Equation \ref{eq:ch02_mop}): 
\begin{equation}
    \min F(x) = (f_1 (x),\ldots,f_m(x))
    \label{eq:ch02_mop}
\end{equation}
Pareto rationality is the theoretical framework to analyse multi-objective optimization problems where $m$ objective functions $f_1 (x),\ldots,f_m (x)$, have to be simultaneously optimized in the search space $\Omega \subseteq \mathbb{R}$. \\
Let $u,v \in \mathbb{R}^m$, for a minimization problem, $u$ is said to dominate $v$ if and only if $u_i \leq v_i \; \forall i=1,\ldots,n$ and $u_j \le v_j$ for at least one index $j$. The goal in multi-objective optimization is to identify the Pareto frontier of $F(x)$. A point $x^*$ is Pareto optimal for the problem in Equation \ref{eq:ch02_mop} if there is no point $x$ such that $F(x)$ dominates $F(x^*)$. The set of all Pareto optimal points is the Pareto set (PS) (Figure \ref{fig:ch02_ps}) and the set of all Pareto optimal objective vectors is the Pareto front (PF) (Figure \ref{fig:ch02_pf}). \\
 \\
The interest in finding locations $x$ having the associated $F(x)$ on the Pareto frontier is clear: all of them represent efficient trade-offs between conflicting objectives and are the only ones, according to the Pareto rationality, to be considered by the decision maker. 
This implies that any improvement in a Pareto optimal point in one objective leads to a deterioration in another. \\
A fundamental difference between single and multi-objective optimization is that it is not obvious which metric to use to evaluate the solution quality. Moreover, the decision maker, post-optimization, must chose a point in the Pareto set according to his/her preference. \\
To measure the progress of the optimization, a natural and widely used metric is the Hypervolume (HV) indicator that measures the objective space between an approximate Pareto front and a predefined reference vector (Figure \ref{fig:ch02_hypervolume}). \\
A marginally or largely dominant Pareto set will result into a respectively low or high hypervolume value; thus, hypervolume is a reasonable measure for evaluating the quality of the optimization process. The hypervolume can be used also to guide the selection of solutions with good convergence and diversity properties \cite{DBLP:journals/eor/BeumeNE07, rodriguez2012new}. These advantages come at a computational cost as hypervolume calculation can be very expensive for many objective problems. \\ 
Another metric to compare different approximations of the Pareto front is the $C$-metric, also called coverage. Let $A$ and $B$ be two approximations of the PF, $C(A,B)$ gives the fraction of solutions in $B$ that are dominated by at least one solution in $A$. Hence, $C(A,B)=1$ means that all solutions in $B$ are dominated by at least one solutions in $A$ while $C(A,B)=0$ implies that no solution in $B$ is dominated by a solution in $A$.

\section{Multi-Objective Evolutionary Learning}
Multi-Objective Evolutionary Algorithms (MOEAs) developed over the last two decades along different strategies. The main two approaches are the non-dominated sorting based, as NSGA-II or NSGA-III, and the decomposition-based, as MOEA/D \cite{DBLP:journals/tec/ZhangL07}. Trivedi et al., \cite{DBLP:journals/tec/TrivediSSG17} gives a comprehensive review of the developments of MOEAs. The question of which strategy works better does not admit an easy answer: it depends on problem features like the shape of Pareto sets, disconnection, or degeneracy \cite{DBLP:journals/swevo/LiDZSC19}. A recent contribution to the decomposition approach is \cite{DBLP:journals/swevo/SunZZZZ19} where a clustering approach method is used to learn the Pareto optimal set structure. Recently it has been recognized the criticality of the exploration/exploitation dilemma in MOEAs \cite{DBLP:journals/tec/ZhangLTV10} which is widely studied in Bayesian optimization and whose solution is based on the predictive uncertainty enabled by the GP model. Most of the evolutionary approaches do not use models for the objective functions and therefore cannot make predictions about unevaluated designs: as a consequence, a large number of function evaluations is usually required. \\
A first solution to mitigate the  problem  of low sample efficiency of EA is the development of problem specific operators. Deb and Myburgh \cite{DBLP:journals/eor/DebM17} proposed problem specific recombination and repair operators \cite{DBLP:journals/eor/BeumeNE07}. Li et al., \cite{DBLP:journals/swevo/LiDZSC19} introduces new test problems with difficult problem features as objective scalability, complicated Pareto sets, bias, disconnection, and degeneracy. In \cite{DBLP:conf/cec/BlankD20} the authors move from the observation that the impact of the shape of Pareto sets has not been properly considered and introduce a set of test instances in order to compare the ability of algorithms to cope with complicated Pareto sets shapes. The issue of comparing and evaluating solution sets provided by different algorithms has been extensively analyzed also in \cite{DBLP:journals/csur/LiY19}. \\
Another solution is to endow the evolutionary strategy with a surrogate model usually a GP. A benchmark method is ParEGO \cite{DBLP:journals/tec/Knowles06} which combines the evolutionary approach with Gaussian based Bayesian optimization. ParEGO is based on the well-known EGO \cite{DBLP:journals/jgo/JonesSW98} algorithm which is an early industrial strength implementation of BO. ParEGO uses as acquisition function, the Expected Improvement, which is optimized by EA.  A similar GP based approach is in \cite{DBLP:journals/tec/EmmerichGN06} which uses Gaussian random fields metamodels to predict the values of the objective functions \cite{DBLP:journals/tec/ZhangLTV10}.

\section{Dominance}
\begin{figure}[ht]
\centering
    \begin{subfigure}{.49\textwidth}
        \centering
        \includegraphics[width=1\linewidth]{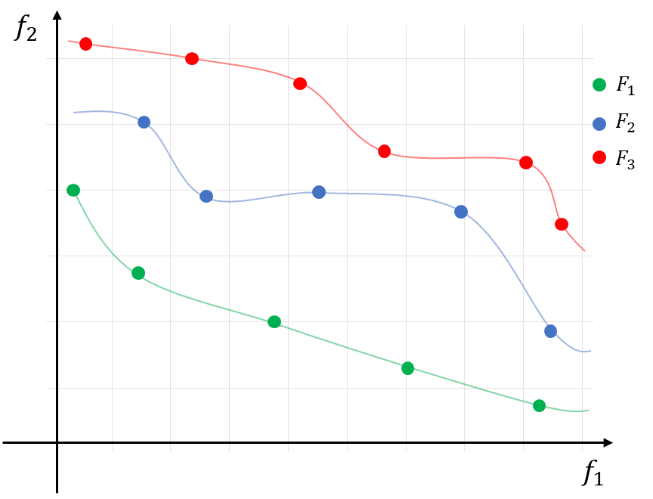}
        \caption{The different frontier identified by the non-dominated sorting procedure.}
    \end{subfigure}
    \begin{subfigure}{.49\textwidth}
        \centering
        \includegraphics[width=1\linewidth]{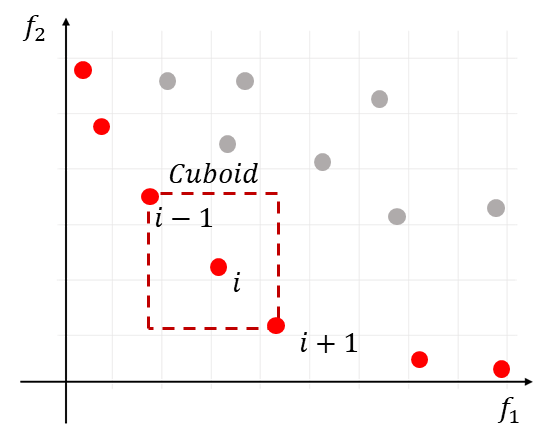}
        \caption{A visual representation of the crowding distance of a point $i$.}
    \end{subfigure}
\caption{The non-dominated sorting and the crowding distance in NSGA-II.}
\label{fig:ch02_sorting_crowding}
\end{figure}

\begin{figure}[h]
    \centering
    \includegraphics[scale = 0.7]{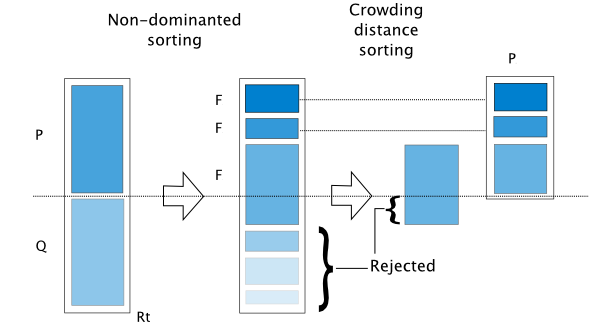}
    \caption{The general framework of NSGA-II.}
    \label{fig:ch02_nsga2}
\end{figure}

Non-dominated Sorting Genetic Algorithm (NSGA-II) is a well-known evolutionary algorithm based on the concept of Pareto dominance proposed in \cite{DBLP:journals/tec/DebAPM02}. NSGA-II is based on two key elements:
\begin{itemize}
    \item An elitist principle, i.e., the elites of a population are given the opportunity for their genes to be carried to the next generation.
    \item An explicit mechanism to preserve the diversity (crowding distance).
\end{itemize}
To identify the elites of a population, NSGA-II rank each individual considering the dominance of a solution on the others. Through the non-dominated sorting, NSGA-II defines different frontiers $F_1,\ldots,F_n$ as follow: first all the non-dominated individuals are assigned to $F_1$, then all the individuals that are dominated only by solutions in $F_1$ are assigned to $F_2$; this process is repeated until all the individuals belong to one frontier as showed in Figure \ref{fig:ch02_sorting_crowding}. \\
 \\
At each generation, the new population is created by picking the individuals in the best frontiers until the size of the original population is reached. To choose between individual with the same rank, the crowding distance ($CD$) is used. Considering an individual $i$ the crowding distance is defined as the perimeter of the cuboid defined by its neighbors $(i-1)$ and $(i+1)$ as shown in Equation \ref{eq:ch02_crowding}.
\begin{equation}
    CD(i) = \sum_{j=1}^{m}{\frac{f_j(i+1)-f_j(i-1)}{\max_{x \in F}{f_j(x)} }}
    \label{eq:ch02_crowding}
\end{equation}
Only the individuals with higher crowding distance will belong to the new population.
The general framework of NSGA-II can be schematized as in Figure \ref{fig:ch02_nsga2}.
\\
\section{Decomposition}
Reduction to a single-objective problem is largely used. This can be done in different ways, for instance considering convex combinations of the objective functions \cite{DBLP:books/cu/BV2014}. The scalarization approach is also followed by some evolutionary algorithms. For example, \cite{DBLP:journals/tec/ZhangL07} leads to the optimization of several single-objective problems. There are many approaches for converting the problem of approximation of the Pareto Front into a number of scalar optimization problems. In the following, two strategies are introduced.

\subsubsection{Weighted Sum Approach}
This approach considers a convex combination of the different objectives. Let $\lambda = (\lambda_1,...,\lambda_m)^T$ be a weight vector, i.e., $\lambda_i \geq 0$ for all $i - 1,...,m$ and $\sum_{i=1}^m \lambda_i = 1$. Then, the optimal solution to the following scalar optimization problem (Equation \ref{eq:ch02_weighted_sum}):  
\begin{equation}
    \begin{aligned}
        & \max g^{ws} (x|\lambda) = \sum_{i=1}^m \lambda_i f_i(x) \\
        & \text{subject to} \; x \in \Omega
    \end{aligned}
    \label{eq:ch02_weighted_sum}
\end{equation}
is a Pareto optimal point, where $g^{ws}(x|\lambda)$ is used to emphasize that $\lambda$ is a coefficient vector in this objective function, while $x$ is the variables to be optimized. To generate a set of different Pareto optimal vectors, one can use different weight vectors $\lambda$ in the above scalar optimization problem. If the PS is concave (convex in the case of minimization), this approach could work well. However, not every Pareto optimal point can be obtained by this approach in the case of non-concave PSs. To overcome these shortcomings, some effort has been made to incorporate other techniques into this approach \cite{DBLP:journals/soco/FanLCHFYMWG19}.

\subsubsection{Chebyshev Approach}
In this approach, the scalar optimization problem is in the form (Equation \ref{eq:tcheby}):
\begin{equation}
    \begin{aligned}
        & \min g^{te} (x|\lambda, z^*) = \max_{1 \leq i \leq m} \{ \lambda_i |f_i(x) - z_i^*| \} \\
        & \text{subject to} \; x \in \Omega
    \end{aligned}
    \label{eq:tcheby}
\end{equation}
where $z^* = (z_1^*,...,z_m^*)^T$ is the reference point (similarly to the hypervolume computation), i.e., $z_i^* = \max \{ f_i(x) | x \in \Omega \}$ for each $i = 1,...,m$. For each Pareto optimal point $x^*$ there exists a weight vector $\lambda$ such that $x^*$ is the optimal solution of Equation \ref{eq:tcheby} and each optimal solution of Equation \ref{eq:tcheby} is a Pareto optimal solution of Equation \ref{eq:ch02_mop}. Therefore, one is able to obtain different Pareto optimal solutions by altering the weight vector. One weakness with this approach is that its aggregation function is not smooth for a continuous MOP. However, it can still be used in the EA framework proposed in this thesis since the algorithm does not need to compute the derivative of the aggregation function.

\subsubsection{MOEA/D}
Multi-Objective Evolutionary Algorithm based on Decomposition (MOEA/D) is a genetic algorithm proposed in \cite{DBLP:journals/tec/ZhangL07}. MOEA/D decomposes the multi-objective problem into a number $N$ of single objective problems associated to an aggregation weight vector and to different points of Pareto Set (Front). Neighbourhood relations among sub-problems are based on the distance between aggregation vectors. As previously mentioned, there are several approaches for converting the problem of approximation of the Pareto front into a number of scalar optimization problems.
Let $\lambda^1,\ldots,\lambda^N$ be a set of even spread weight vectors and $z^*$ be the reference point. The problem of approximation of the PF can be decomposed into $N$ scalar optimization sub-problems by using the Chebyshev approach and the objective function of the $j$-th sub-problem is (Equation \ref{eq:ch02_moead}):
\begin{equation}
    \min g^{te} (x|\lambda^j, z^*) = \max_{1 \leq i \leq m} \{ \lambda_i^j |f_i(x) - z_i^*| \}
    \label{eq:ch02_moead}
\end{equation}
where $\lambda^j = (\lambda_1^j,\ldots,\lambda_m^j )^T$. MOEA/D minimizes all these $N$ objective functions simultaneously in a single run.\\
The general framework can be summarized as follows:
\begin{enumerate}
    \item \textbf{Initialization}:
    \begin{itemize}
        \item Compute the Euclidean distances between any two weight vectors and then work out the $T$ closest weight vectors to each weight vector. For each $i=1,\ldots,N$, set $B(i)={i_1,\ldots,i_T}$, where $\lambda^(i_1),\ldots,\lambda^(i_T)$ are the $T$ closest weight vectors to $\lambda^i$.
    \end{itemize}
    \item \textbf{Update}: 
    \begin{itemize}
        \item Reproduction: randomly select two indexes $k$,$l$ from $B(i)$, and then generate a new solution $y$ from $x^k$ and $x^l$ by using genetic operators. Two solutions have a chance to mate only when they are from neighboring sub-problems.
        \item Improvement: apply a problem-specific repair/improvement heuristic on $y$ to produce $y'$.
    \end{itemize}
\end{enumerate}
This process is repeated until a termination criteria is satisfied, such as the number of generations or the number of function evaluations. In initialization, $B(i)$ contains the indexes of the $T$ closest vectors of $\lambda_i$. The Euclidean distance is used to measure the closeness between any two weight vectors. Therefore, $\lambda_i$’s closest vector is itself, and then $i \in B(i)$. If $j \in B(i)$, the $j$-th sub-problem can be regarded as a neighbor of the $i$-th sub-problem.\\
MOEA needs to maintain diversity in its population for producing a set of representative solutions. Most, if not all, of non-decomposition MOEAs such as NSGA-II use crowding distances among the solutions in their selection to maintain diversity. However, it is not always easy to generate a uniform distribution of Pareto optimal objective vectors in these algorithms. In MOEA/D, a multi-objective problem is decomposed into several scalar optimization sub-problems. Different solutions in the current population are associated with different sub-problems. The ``diversity'' among these sub-problems will naturally lead to diversity in the population.

%% file: chapters/chapter03.tex
\chapter{Bayesian Optimization}
\label{ch03:bo}
Bayesian Optimization (BO) is a sequential strategy for global optimization of black-box functions (i.e., functions whose analytical form is unknown). Surrogate models are a key component of BO, in this thesis are considered the Gaussian Processes. Gaussian Processes are a powerful formalism for implementing both regression and classification algorithms. While most of the regression algorithms provides a deterministic output, GPs also offer a reliable estimate of uncertainty. This chapter presents the basic mathematics underlying this powerful tool.

\section{Gaussian Process Regression}
\begin{figure}[h]
    \centering
    \includegraphics[scale = 0.7]{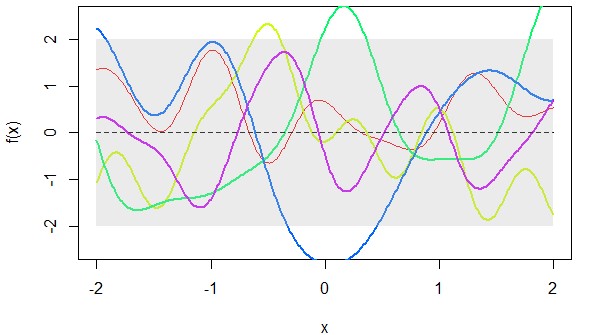}
    \caption{Five different samples from the prior of a GP with Squared Exponential kernel as covariance function.}
    \label{fig:ch03_sample_se_gp}
\end{figure}

\begin{figure}[h]
\centering
    \begin{subfigure}{.49\textwidth}
        \centering
        \includegraphics[width=1\linewidth]{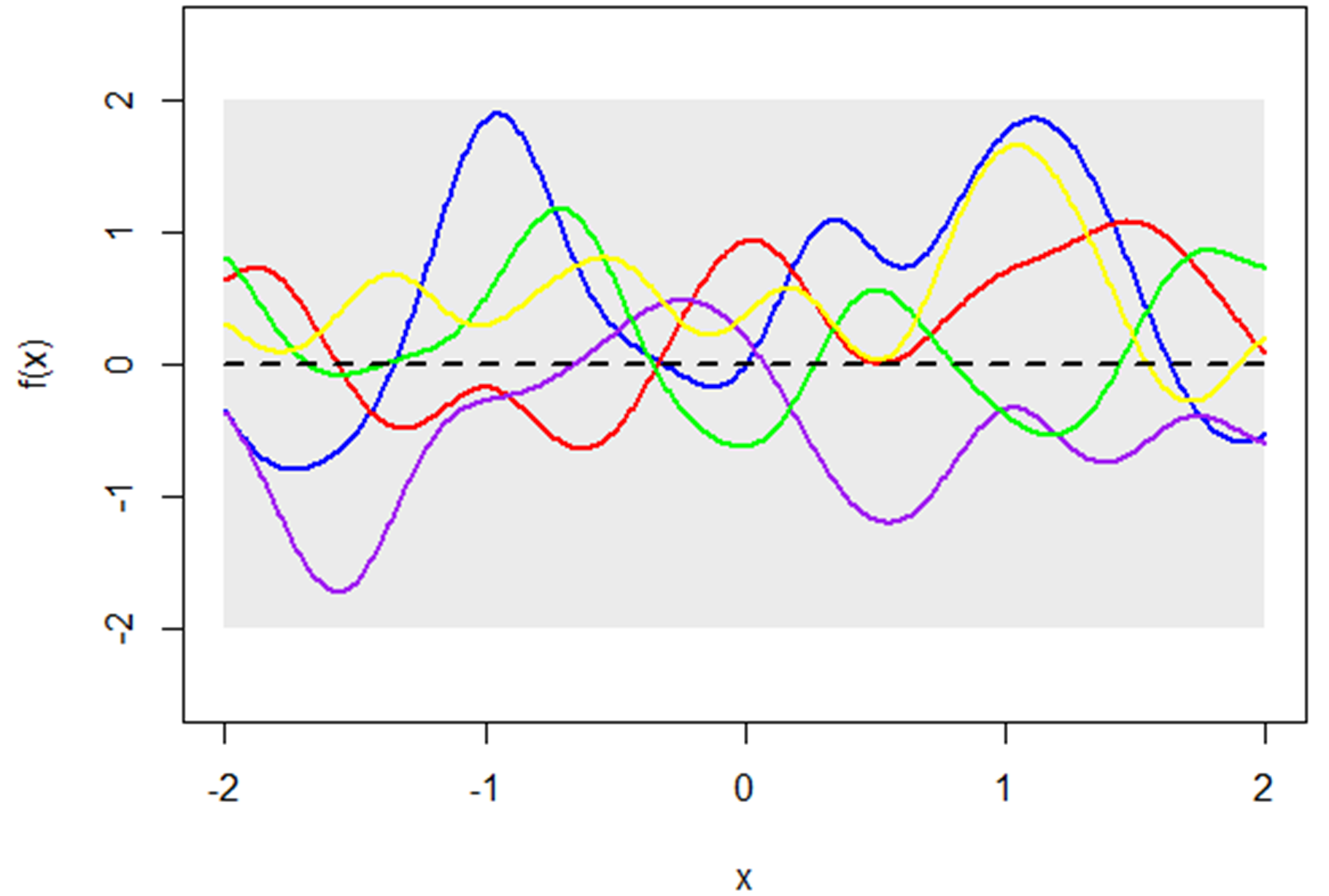}
        \caption{Sampling from prior.}
    \end{subfigure}
    \begin{subfigure}{.49\textwidth}
        \centering
        \includegraphics[width=1\linewidth]{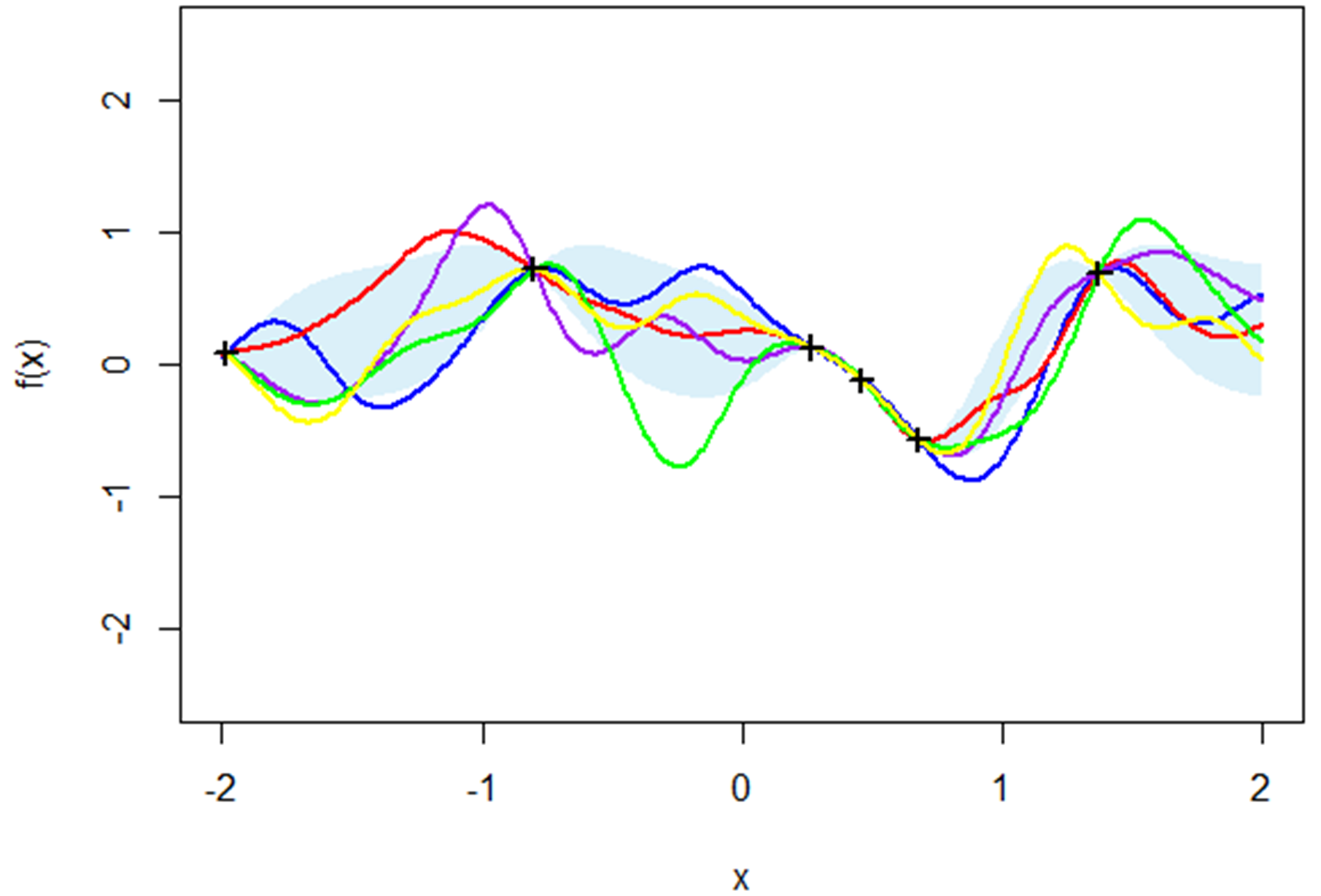}
        \caption{Sampling from posterior.}
    \end{subfigure}
\caption{Sampling from prior vs sampling from posterior (for the sake of simplicity, the noisy-free setting is considered).}
 \label{fig:ch03_prior_posterior}
\end{figure}

\begin{figure}[h]
    \centering
    \includegraphics[scale = 0.2]{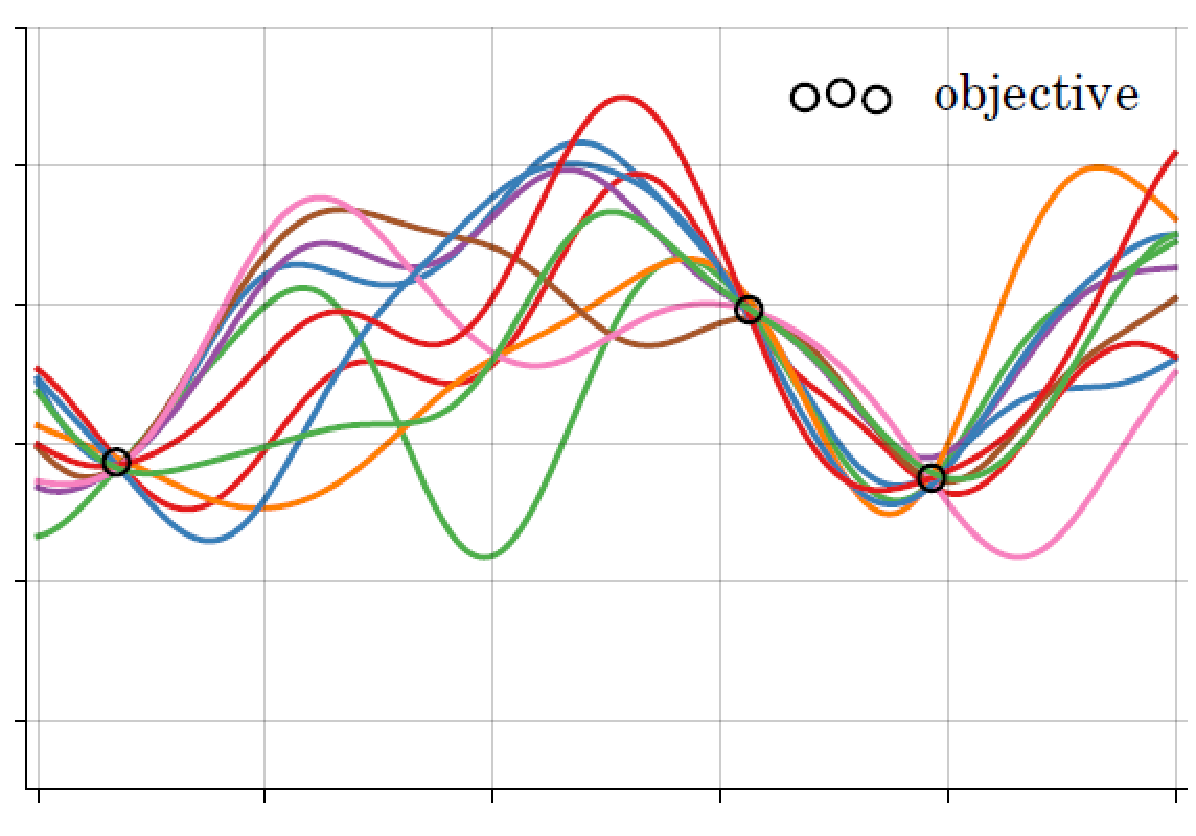}
    \caption{Different function compatible with the function observations $D_{1:3}$.}
    \label{fig:ch03_compatible_function}
\end{figure}

One way to interpret a Gaussian process (GP) regression model is to think of it as defining a distribution over functions, and with inference taking place directly in the space of functions (i.e., function-space view) \cite{DBLP:books/lib/RasmussenW06}. A GP is a collection of random variables, any finite number of which have a joint Gaussian distribution. A GP is completely specified by its mean function $\mu (x)$ (Equation \ref{eq:ch03_mean}) and covariance function $cov(f(x),f(x'))=k(x,x')$ (Equation \ref{eq:ch03_cov}):
\begin{gather}
    \mu(x)=\mathbb{E}[f(x)] \label{eq:ch03_mean}\\
    cov(f(x),f(x')) = k(x, x') = \mathbb{E}[(f(x) - \mu (x))(f(x') - \mu (x'))] \label{eq:ch03_cov}
\end{gather}
and will write the Gaussian process as (Equation \ref{eq:ch03_gp}):
\begin{equation}
    f(x) \sim GP(\mu (x), k(x, x'))
    \label{eq:ch03_gp}
\end{equation}
The covariance function assumes a critical role in the GP modelling, as it specifies the distribution over functions. To see this, consider samples from the distribution of functions evaluated at any number of points; in detail, a set of input points $X_{1:n} = (x_1,\ldots,x_n)^T$ is chosen and then the corresponding covariance matrix elementwise is computed. This operation is usually performed by using predefined covariance functions allowing to write covariance between outputs as a function of inputs (i.e., $cov(f(x),f(x' ))= k(x,x'))$. Finally, a random Gaussian vector can be generated as (Equation \ref{eq:ch03_random_gauss}):
\begin{equation}
    f(X_{1:n}) \sim \mathcal{N}(0,K(X_{1:n},X_{1:n})
    \label{eq:ch03_random_gauss}
\end{equation}
This is basically known as sampling from prior. The following is an example (Figure \ref{fig:ch03_sample_se_gp}) of five different GP samples drawn from the GP prior: the covariance function used is known as the Squared Exponential (SE) kernel. \\
 \\
Usually, the primarily interest is not in drawing random functions from the prior but is to incorporate the knowledge about the function obtained through the evaluations performed so far. Such a knowledge will be then used by the acquisition function in order to associate an informational utility to each point $x \in X$. Often, the function values are noisy and so consider $y = f(x) + \varepsilon$. Assuming additive independent and identically distributed Gaussian noise $\varepsilon$ with variance $\lambda^2$, the prior on the noisy observations becomes (Equation \ref{eq:ch03_prior_noisy}):
\begin{equation}
    cov(f(x),f(x')) = k(x, x') + \lambda^2 \delta_{xx'}
    \label{eq:ch03_prior_noisy}
\end{equation}
where $\delta_{xx'}$ is a Kronecker delta which is equal to $1$ if and only if $x = x'$. Thus, the covariance over all the function values $y=(y_1,\ldots,y_n)$ is (Equation \ref{eq:ch03_cov_noisy}):
\begin{equation}
    cov(y) = K(X_{1:n}, X_{1:n}) + \lambda^2 I
    \label{eq:ch03_cov_noisy}
\end{equation}
Therefore, the predictive equations for GP regression, that are $\mu (x)$ and $k(x, x')$, can be easily updated, by conditioning the joint Gaussian prior distribution on the observations (Equations \ref{eq:ch03_new_mean} and \ref{eq:ch03_new_std}):
\begin{gather}
    \mu (x) = \mathbb{E} [f(x) | D_{1:n}, x] = k(x, X_{1:n})[K(X_{1:n}, X_{1:n}) + \lambda^2 I]^{-1}y \label{eq:ch03_new_mean}\\
    \sigma^2 (x) = k(x,x) - k(x, X_{1:n})[K(X_{1:n}, X_{1:n}) + \lambda^2 I]^{-1}k(X_{1:n}, x) \label{eq:ch03_new_std}
\end{gather}
It follows a simple example (Figure \ref{fig:ch03_prior_posterior}) of five different samples drawn from a GP prior and posterior, respectively. Posterior is conditioned to six function observations.\\
 \\
These figures shows that even removing the noise, as it’s often assumed, the problem is still of ``structural uncertainty''. For instance, considering three noise-free evaluations of $f(x)$, $D_{1:3} = \{ (x_i, y_i) \}_{i=1,\ldots,3}$, there still are an infinite number of functions with different minima and minimizers, compatible with $D_{1:3}$, as depicted in Figure \ref{fig:ch03_compatible_function}. \\
It is easy to show that the mean prediction is a linear combination of $n$ functions, each one centered on an evaluated point. This allows to write $\mu (x)$ as (Equation \ref{eq:ch03_new_mu}):
\begin{equation}
    \mu (x) = \sum_{i=1}^{n} {\alpha_i k(x, x_i)}
    \label{eq:ch03_new_mu}
\end{equation}
where the vector $\alpha = [K(X_{1:n}, X_{1:n}) + \lambda^2 I]^{-1}y$ and $\alpha_i$ is the $i$-th component of the vector $\alpha$, given by the product between the $i$-th row of the matrix $[K(X_{1:n}, X_{1:n}) + \lambda^2 I]^{-1}$ and the vector $y$. This means that, to make a prediction at a given $x$, it is necessary only to consider the $(n+1)$-dimensional distribution defined by the $n$ function evaluations performed so far and the new point $x$ to evaluate. \\
Covariance functions are referred to in BO as kernels and some of the most widely used of them will be presented in the following section. \\
 \\
Every kernel has some hyperparameters to be set up, defining shape features of the GP, such as smoothness and amplitude. The values of the hyperparameters are usually unknown a priori and are approximated using maximum likelihood estimators on the basis of the observations $D_{1:n}$.

\section{Kernel: the data geometry of BO}
\begin{figure}[h]
\centering
    \begin{subfigure}{.49\textwidth}
        \centering
        \includegraphics[width=0.8\linewidth]{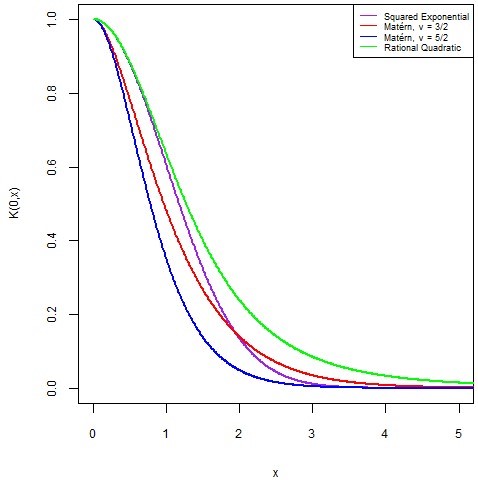}
        \caption{Value of four different kernels with $x$ moving away from $x'=0$.}
        \label{fig:ch03_different_kernel_1}
    \end{subfigure}
    \begin{subfigure}{.49\textwidth}
        \centering
        \includegraphics[width=0.8\linewidth]{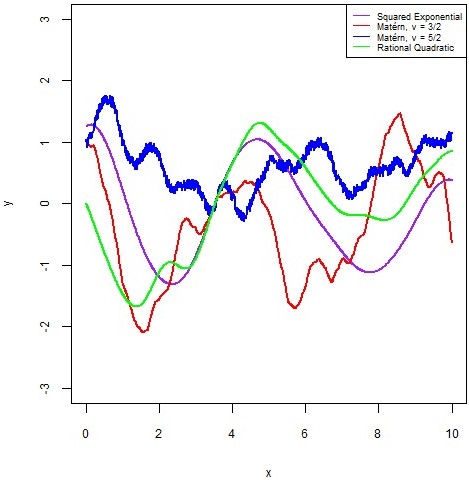}
        \caption{Four samples from GP prior, one for each kernel considered.}
        \label{fig:ch03_different_kernel_2}
    \end{subfigure}
\caption{The value of the characteristic length-scale is $\ell = 1$ for all the four kernels; $\alpha$ of the RQ kernel is set to $2.25$.}
\label{fig:ch03_different_kernel}
\end{figure}

A covariance function is the crucial ingredient in a GP predictor, as it encodes assumptions about the function to approximate. From a slightly different viewpoint, it is clear that, in learning, the notion of similarity between data points is crucial; it is a basic assumption that points which are close in $x$ are likely to have similar target values $y$, and thus function evaluations that are near to a given point should be informative about the prediction at that point. Under the GP view it is the covariance function that defines nearness or similarity.\\
A general name for a function $k$ of two arguments mapping a pair of inputs $x$ and $x'$ into a scalar is kernel. For a kernel to be a covariance function the following conditions must be satisfied:
\begin{itemize}
    \item The kernel has to be symmetric $k(x,x') = k(x',x)$.
    \item The matrix $K$ with entries $K_{ij} = k(x_i, x_j)$, also known as Gram matrix, must be positive semidefinite.
\end{itemize}
In the following, some examples of covariance (aka kernel) functions.\\ \\
\noindent
\textbf{Squared Exponential kernel} (Equation \ref{eq:ch03_se}):
\begin{equation}
    k_{SE}(x, x') = \exp^{-\frac{||x-x'||^2}{2\ell^2}}
    \label{eq:ch03_se}
\end{equation}
with $\ell$ known as characteristic length-scale. This kernel is infinitely differentiable, meaning that the GP’s sample functions are very ``smooth''. \\ \\
\noindent
\textbf{Matérn kernels} (Equation \ref{eq:ch03_matern}):
\begin{equation}
    k_{Mat} (x, x') = \frac{2^{1-v}}{\Gamma(v)} \left( \frac{|x - x'|\sqrt{2v}}{\ell} \right)^v K_v \left( \frac{|x - x'| \sqrt{2v}}{\ell} \right)
    \label{eq:ch03_matern}
\end{equation}
with two hyperparameters $v$ and $\ell$, and where $K_v$ is a modified Bessel function. Note that for $v \to \infty$ it is equivalent to the Squared Exponential kernel. The Matérn covariance functions become especially simple when $v$ is half-integer: $v=p+\frac{1}{2}$, where $p$ is a non-negative integer. In this case the covariance function is a product of an exponential and a polynomial of order $p$. The most widely adopted versions, specifically in the Machine Learning community, are $v=\frac{3}{2}$ (Equation \ref{eq:ch03_matern32}) and $v=\frac{5}{2}$ (Equation \ref{eq:ch03_matern52}).
\begin{gather}
    k_{v=\frac{3}{2}} (x, x') = \left( 1 + \frac{|x - x'|\sqrt{3}}{\ell} \right) \exp^{- \frac{|x - x'| \sqrt{3}}{\ell}} \label{eq:ch03_matern32}\\
    k_{v=\frac{5}{2}} (x, x') = \left( 1 + \frac{|x - x'|\sqrt{5}}{\ell} + \frac{(x - x')^2}{3 \ell^2} \right) \exp^{- \frac{|x - x'| \sqrt{5}}{\ell}} \label{eq:ch03_matern52}
\end{gather} \\ \\
\noindent
\textbf{Rational Quadratic Covariance function} (Equation \ref{eq:ch03_rqc}):
\begin{equation}
    k_{RQ} (x, x') = \left( 1 + \frac{(x - x')^2}{2 \alpha \ell^2} \right)^{- \alpha}
    \label{eq:ch03_rqc}
\end{equation}
where $\alpha$ and $\ell$ are two hyperparameters. This kernel can be considered as an infinite sum (scale mixture) of Squared Exponential kernels, with different characteristic length-scales. \\ \\
\noindent
Figure \ref{fig:ch03_different_kernel} summarizes how the value of the four kernels decreases with $x$ moving away from $x' = 0$ (Figure \ref{fig:ch03_different_kernel_1}) and which are possible resulting samples with different shape properties (Figure \ref{fig:ch03_different_kernel_2}). \\ \\
 \\

\section{The Acquisition Function}
The acquisition function is the mechanism to implement the trade-off between exploration and exploitation in Bayesian Optimization (BO). More precisely, any acquisition function aims to guide the search of the optimum towards points with potentially low values of objective function either because the prediction of $f(x)$, based on the probabilistic surrogate model, is low or the uncertainty, also based on the same model, is high (or both). Indeed, exploitation means to consider the area providing more chance to improve over the current solution (with respect to the current surrogate model), while exploring means to move towards less explored regions of the search space where predictions based on the surrogate model are more uncertain, with higher variance.

\subsubsection{Probability of Improvement}
\begin{figure}[h]
    \centering
    \includegraphics[scale = 0.5]{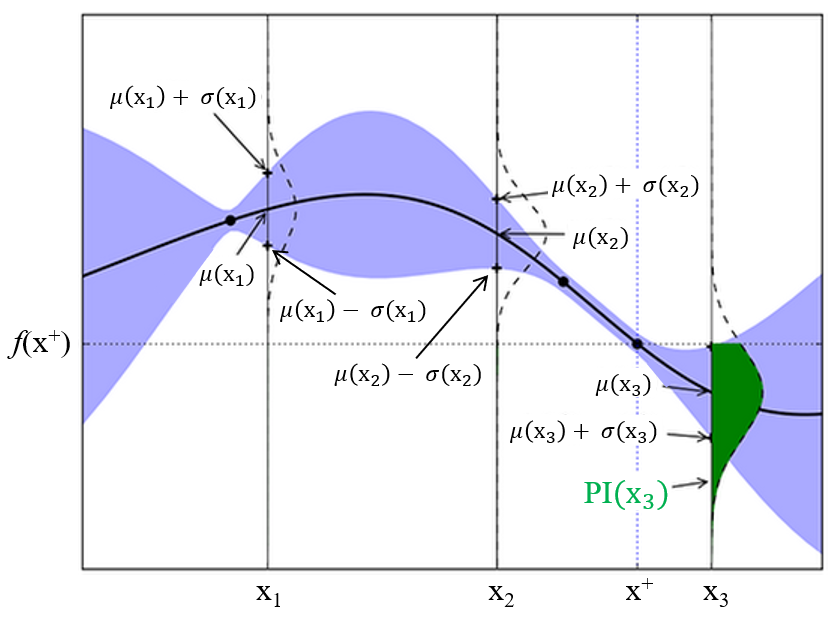}
    \caption{An example of how PI selects the new point.}
    \label{fig:ch03_prob_improvement}
\end{figure}

Probability of Improvement (PI) was the first acquisition function proposed in the literature \cite{kushner1964new}. One of the drawbacks of PI is that it is biased towards exploitation. To mitigate this effect, it can be introduced the parameter $\xi$ which modulates the balance between exploration and exploitation. The resulting equation is (Equation \ref{eq:ch03_pi}):
\begin{equation}
    PI(x) = P( f(x) \leq f(x^+) + \xi ) = \Phi \left( \frac{f(x^+) - \mu (x) - \xi}{\sigma(x)} \right)
    \label{eq:ch03_pi}
\end{equation}
More precisely, $\xi = 0$ is towards exploitation, while $\xi > 0$ is more towards exploration. \\
Finally, the next point to evaluate is chosen according to Equation \ref{eq:ch03_pi_next}.
\begin{equation}
    x_{n+1} = \argmax_{x \in  X} PI(x)
    \label{eq:ch03_pi_next}
\end{equation}
Figure \ref{fig:ch03_prob_improvement} shows that the value of PI in $x_3$, given the best $y$ value obtained in $x^+$ corresponds to the area depicted in green. \\
 \\
However, a weak point of PI is to assign a value to a new point irrespectively of the potential magnitude of the improvement. This is the reason why the next acquisition function was proposed.

\subsubsection{Expected Improvement}
Expected Improvement (EI) was initially proposed in \cite{mockus1978application} and then made popular in \cite{DBLP:journals/jgo/JonesSW98} which measures the expectation of the improvement on $f(x)$ with respect to the predictive distribution of the probabilistic surrogate model (Equation \ref{eq:ch03_ei}).
\begin{equation}
    EI(x) = \left\{ \begin{array}{rcl}
        (f(x^+) - \mu(x)) \Phi (Z) + \sigma (x) \phi (Z) & \text{if} & \sigma(x) > 0 \\
        0 & \text{if} & \sigma(x) = 0
    \end{array}\right.
    \label{eq:ch03_ei}
\end{equation}
In the equation $\phi (Z)$ and $\Phi(Z)$ represent the probability distribution and the cumulative distribution of the standardized normal, respectively, where (Equation \ref{eq:ch03_ei_z}):
\begin{equation}
    Z = \left\{\begin{array}{rcl}
        \frac{f(x^+) - \mu(x)}{\sigma(x)} & \text{if} & \sigma(x) > 0 \\
        0 & \text{if} & \sigma(x) = 0
    \end{array}\right.
    \label{eq:ch03_ei_z}
\end{equation}
EI is made up of two terms: the first is increased by decreasing the predictive mean; the second by increasing the predictive uncertainty. Thus, EI, in a sense, automatically balances, respectively, exploitation and exploration. It is possible to actively manage the trade-off between exploration and exploitation, introducing the parameter $\xi$. When exploring, points associated to high uncertainty of the probabilistic surrogate model are more likely to be chosen, while when exploiting, points associated to low value of the mean of the probabilistic surrogate model are selected.\\
Finally, the next point to evaluate is chosen according to Equation \ref{eq:ch03_ei_next}.
\begin{equation}
    x_{n+1} = \argmax_{x \in  X} EI(x)
    \label{eq:ch03_ei_next}
\end{equation}
EI has been largely used since 1998 and specialized to specific contexts. Astudillo and Frazier \cite{DBLP:conf/icml/AstudilloF19} propose a version for composite functions (EI-CF) which leads to a multi-output GP: the authors also note that constrained optimization can be regarded as a special case of the optimization of composite functions and that EI-CF reduces to the expected improvement for constrained optimization.

\subsubsection{Upper/Lower Confidence Bound}
Confidence Bound, where Upper and Lower are used, respectively for maximization and minimization problems, is an acquisition function that manage exploration-exploitation by being optimistic in the face of uncertainty, in the sense of considering the best-case scenario for a given probability value \cite{DBLP:journals/jmlr/Auer02}. For the case of minimization, LCB is given by Equation \ref{eq:ch03_lcb}:
\begin{equation}
    LCB(x) = \mu (x) - \xi \sigma (x)
    \label{eq:ch03_lcb}
\end{equation}
where $\xi \geq 0$ is the parameter to manage the trade-off between exploration and exploitation ($\xi = 0$ is for pure exploitation; on the contrary, higher values of $\xi$ emphasizes exploration by inflating the model uncertainty). For this acquisition function there are strong theoretical results, originated in the context of multi-armed bandit problems, on achieving the optimal regret \cite{DBLP:journals/tit/SrinivasKKS12}. For the candidate point $x_n$ instantaneous regret $r_n = f(x_n) - f(x^*)$ can be observed. The cumulative regret $R_N$ after $N$ function evaluations is the sum of instantaneous regrets $R_N = \sum_{n=1}^N {r_n}$. A desirable asymptotic property of an algorithm is to be no-regret $\lim_{N \to \infty} \frac{R_n}{N} = 0$. Bounds on the average regret $\frac{R_N}{N}$ translate to convergence rates: $f(x^+) = \min_{x_{n \leq N}} f(x_n)$ in the first $N$ function evaluations is no further from $f(x^*)$ than the average regret. Therefore, $f(x^+) - f(x^*) \to 0$, with $N \to \infty$. Finally, the next point to evaluate is chosen according to $x_{n+1} = \argmin_{x \in  X} LCB(x)$, in the case of a minimization problem, or $x_{n+1} = \argmax_{x \in  X} LCB(x)$ in the case of a maximization problem.

\section{Bayesian Optimization framework}
A BO framework consists of two main component:  a surrogate model for modelling the objective function, and an acquisition function for deciding where to sample next. After evaluating the objective according to an initial space-filling experimental design, they are used iteratively to allocate the remainder of a budget of $N$ function evaluations. Algorithm \ref{pc:ch03_bo} summarizes a general Bayesian Optimization process where the acquisition function, whichever it is, is denoted by $\alpha (x, D_{1:n})$. This function is generally maximized, except for the case of $\alpha = LCB$.
\begin{algorithm}[h]
\SetAlgoLined
 \CommentSty{Generate an initial set of $m$ points $X_{1:m}$ randomly sampled}\;
 \CommentSty{Evaluate the function in the initial set of points and obtain $D_{1:m}$}\;
 \CommentSty{Define a further budget $N$}\;
    \For{$n = m, \ldots, m + N$}{
        \CommentSty{Update the surrogate model obtaining the new estimates of $\mu(x)$ and $\sigma(x)$}\;
        \CommentSty{Select a new $x_{n+1}$ by optimizing an acquisition function $\alpha$, such that $x_{n+1} = \argmax_x \alpha (x | D_{1:n})$}\;
        \CommentSty{Evaluate the objective function to obtain $y_{n+1} = f(x_{n+1})$}\;
        \CommentSty{Update the dataset of observations $D_{1:n+1} = D_{1:n} \bigcup \{(x_{n+1},y_{n+1})\}$}\;
    }
 \KwResult{The best $y$ value observed over the entire optimization process}
 \caption{General Bayesian Optimization algorithm}
 \label{pc:ch03_bo}
\end{algorithm}

\section{Advanced topics in Bayesian and Evolutionary learning}
Early results are \cite{DBLP:conf/nips/SwerskySA13, DBLP:conf/icml/BardenetBKS13} in which BO is generalized for multiple related objectives using dependencies among the tasks (objectives) in order to share information.
One should anyway remark that these methods look for learning efficiently separate optimizers for each objective rather than approximating the Pareto frontier. Multi-objective Bayesian Optimization (MOBO) has been related to constrained optimization in \cite{DBLP:conf/uai/GelbartSA14} which proposes BO under unknown constraints as a way to obtain Pareto-optimal solutions. A significant contribution is \cite{DBLP:journals/jmlr/ZuluagaKP16} which uses LCB/UCB to model for each $x$ an uncertainty hyper-rectangle.  To select the next evaluation point instead of optimizing the acquisition function it samples in this hyper-rectangle in a way that favours exploration. Recent papers about this topic are \cite{DBLP:conf/gecco/RahatEF17, DBLP:conf/aaai/BelakariaDD20, DBLP:conf/aaai/BelakariaDJD20, DBLP:conf/icml/SuzukiTTSK20}. \\
Gaussian processes have been used in evolutionary algorithms as ParEGO and MOEA/D-GP \cite{DBLP:journals/tec/ZhangLTV10}. Both methods use a Chebyshev-based decomposition approach. Each resulting scalar aggregate objective is optimized by BO using Expected Improvement as acquisition function. A genetic algorithm is then used to maximize the expected improvement. The issue of balancing exploration and exploitation in multi-objective evolutionary optimization has been recently investigated in \cite{DBLP:journals/isci/ZhangSLZZ19}. Another approach is SMS-EMOA based on the efficient computation of the hypervolume applied as a selection criterion to discard the individuals which contribute the least hypervolume improvement \cite{DBLP:conf/cec/IshibuchiIMN18, DBLP:journals/eor/BeumeNE07}. \\
In the comparative analysis in Chapter \ref{ch07:water} a novel extension of ParEGO, that supports parallel evaluation and constraints, is used, namely \textit{q}ParEGO. Different to the classical implementation, \textit{q}ParEGO computes gradients via auto-differentiation for the optimization of acquisition functions. ParEGO is typically implemented by applying augmented Chebyshev scalarization and modelling the scalarized outcome. However, \textit{q}ParEGO uses a Monte Carlo-based Expected Improvement acquisition function, where the objectives are modelled independently and the augmented Chebyshev scalarization is applied to the posterior samples as a composite objective. This approach enables the use of sequential greedy optimization of $q$ candidates with proper integration over the posterior at the pending points. Importantly, the sequential greedy approach allows for using different random scalarization weights for selecting each of the $q$ candidates. \textit{q}ParEGO can also be extended to the constrained case by weighting the EI by the probability of feasibility. \textit{q}ParEGO is implemented using BoTorch, a very recent framework released by Facebook in 2019. For a description of BoTorch the reader is referred to Chapter \ref{ch09:botorch}.

%% file: chapters/chapter04.tex
\chapter{Instances of Multi-Task Learning on Networks}
\label{ch04:instances}
\section{From water networks to outbreak detection}
\begin{figure}[h]
    \centering
    \includegraphics[scale = 0.7]{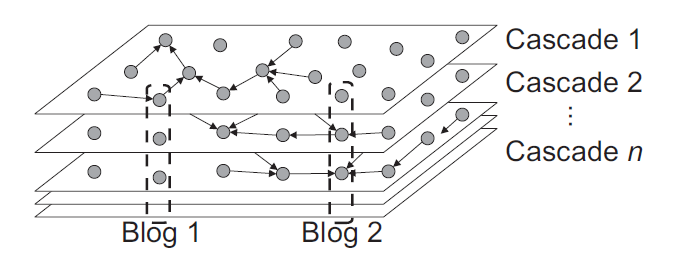}
    \caption{Spread of information between blogs. Each layer shows an information cascade. The objective is to find few blogs that quickly capture most cascades.}
    \label{fig:ch04_blogosphere}
\end{figure}

\begin{figure}[h]
    \centering
    \includegraphics[scale = 0.7]{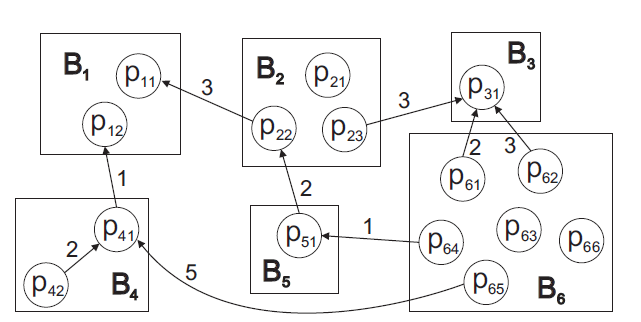}
    \caption{An example of a blogosphere. Blogs contain posts that are linked to the sources of informations. The cascades grow in the reverse direction of the edges.}
    \label{fig:ch04_blogs_posts}
\end{figure}

\begin{figure}[h]
\centering
    \begin{subfigure}{.49\textwidth}
        \centering
        \includegraphics[width=1\linewidth]{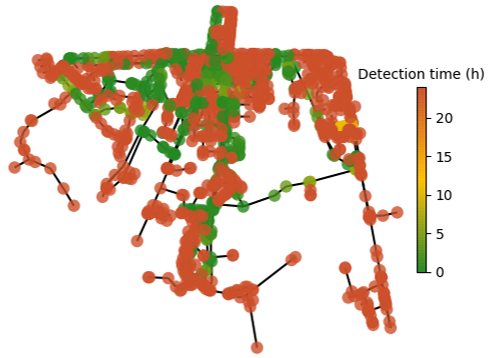}
        \caption{The impact over the events considering a sensor placement of 150 sensors. The mean detection time is 15 hours.}
    \end{subfigure}
    \begin{subfigure}{.49\textwidth}
        \centering
        \includegraphics[width=1\linewidth]{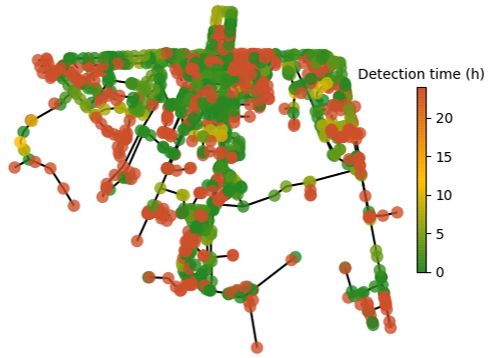}
        \caption{The impact over the events considering a different sensor placement of 150 sensors. The mean detection time is 9 hours.}
    \end{subfigure}
\caption{Impact of contamination events on a water distribution network consisting of 1213 nodes and 1393 edges. The color of each node indicates the impact before detection if a contaminant is introduced at the node and for a specific sensor placement. As impact the detection time is considered.}
\label{fig:ch04_wdn_impact}
\end{figure}

Consider a network and a dynamic process spreading over this network, it is possible to deploy a set of sensors at the nodes with the aim to select a set of nodes to detect the process as effectively as possible. Many real-world problems can be modelled under this setting. Consider a urban water distribution network, delivering water to consumers via pipes and junctions. Accidental or malicious intrusions can cause contaminants to spread over the network, and the objective is to select a few locations (pipes or junctions) to install sensors, in order to detect these contaminations as quickly as possible \cite{ponti2021new, ponti2021wasserstein}. Typical epidemics scenarios also fit into this outbreak detection setting: it is possible to early detect a disease outbreak by monitoring only a small set of people within a social network of interactions. \\
In the domain of web blogs, bloggers publish posts and use hyperlinks to refer to other bloggers’ posts and content on the web. Each post is time stamped, so the spread of information on the ``blogosphere''  can observed. In this setting, the aim is to select a set of blogs to read (or retrieve) which are most up to date, i.e., catch most of the stories that propagate over the blogosphere. Figure \ref{fig:ch04_blogosphere} illustrates this setting. Each layer plots the propagation graph of the information, also called information cascade. Circles correspond to blog posts, and all posts at the same vertical column belong to the same blog. Edges indicate the temporal flow of information: the cascade starts at some post (e.g., top left circle of the top layer of Figure \ref{fig:ch04_blogosphere}) and then the information propagates recursively by other posts linking to it. The goal is to select a small set of blogs (two in case of Figure \ref{fig:ch04_blogosphere}) which ``catch'' as many cascades (stories) as possible. There are several possible criteria one may want to optimize in outbreak detection. For example, one criterion seeks to minimize detection time (i.e., to know about a cascade as soon as possible, or avoid spreading of contaminated water). Similarly, another criterion seeks to minimize the population affected by an undetected outbreak (i.e., the number of blogs referring to the story, or the population consuming the contaminated water prior to detection). \\
Optimizing these objective functions over all possible sensor placements is NP-hard, so for large, real-world problems, it is not possible to find the optimal solution and meta-heuristic must be used. \\
 \\
The water distribution and blogosphere monitoring problems, even though in very different domains, share essential structure \cite{DBLP:conf/kdd/LeskovecKGFVG07}. In both problems, the objective is to select a subset of nodes (sensor locations, blogs) in a graph, which detect outbreaks (spreading of a virus/information) quickly (Figure \ref{fig:ch04_blogs_posts}). These outbreaks (e.g., information cascades) initiate from a single node of the network, and spread over the graph, such that the traversal of every edge takes a certain amount of time (indicated by the edge labels). As soon as the event reaches selected node, alarm is triggered. \\
 \\
Depending on the selected nodes, a different ``placement score'' is achieved. Figure \ref{fig:ch04_blogs_posts} illustrates several criteria one may want to optimize. If one only want to detect as many stories as possible, then reading just blog B6 is the best choice. However, reading B1 would miss one cascade, but would detect the other cascades immediately. In general, this placement score (representing, e.g., the fraction of detected cascades, or the population saved by placing a sensor) is a multi-value set function, mapping every placement to the values of each objective functions, which have to be maximized
Figure \ref{fig:ch04_wdn_impact} displays the impact over the detection time of different sensor placements in a water distribution network. \\
 \\

\section{Recommender Systems}
Recommender systems (RS) represent a critical component of B2C online services. They recommend items (movies, songs, books, etc.) that fit the user’s preferences, to help the user in selecting items from a large set of choices. Personalized recommendations have huge importance where the number of possible items is large, such as in e-commerce related to art (books, movies, music), fashion, food, etc. Some of the major participants in e-commerce (Amazon), movie streaming (Netflix), and music streaming (Spotify) successfully apply recommender systems to deliver automatically generated personalized recommendations to their customers. \\
Machine learning (ML) algorithms in Recommender systems are typically classified into two categories:
\begin{itemize}
    \item Content-based approaches profile users and items by identifying their characteristic features, such as demographic data for user profiling, and product information/descriptions for item profiling.
    \item Collaborative filtering approaches (CF) identify relationships between users and items and make associations using the past user activities information to predict user preferences on new items.
\end{itemize}
A drawback of the first approach is the necessity to collect information about users/items, and it is often tricky because the users must share their personal data for the creation of a database for profiling. 
The CF approach requires relatively fewer data, basically a list of tuples containing the user ID, the item ID, and the rating done by the user to that item. Therefore, the CF algorithms could be applied to RS independently of the domain of application. \\
In this thesis the focus is on CF, in which the basic data structure is the rating matrix, whose entries correspond to the rating of any possible user-item combination. Rating matrices are mostly sparse (many unknown entries): the key assumption is that the unknown ratings are predictable because the known ratings are often highly correlated across various users or items and once these correlations have been computed they can be used to fill the matrix. \\
The problem is also called the matrix completion problem. Two types of methods are commonly used to solve it: the memory-based methods and model-based methods. \\
The memory-based methods, or neighbourhood-based algorithms, were among the earliest collaborative filtering algorithms, in which the ratings of user-item combinations are predicted based on their neighbourhoods (users similar to a target user or items similar to a target item). They are based on the fact that similar users display similar patterns of rating behaviour (user-based) or similar items receive similar ratings (item-based). These methods are simple to implement, and the resulting recommendations are often easy to explain. It’s worth noting that the rating matrix is implicitly mapped into a graph, called ``$k$-nearest neighbors graph'' in which two users/items (vertices) $i$ and $j$ are connected by an edge if their distance is among the $k$-smallest distances from $i$ to the other users/items $j$. Clearly the choice of the distance and the value $k$ impact substantially the performance of the method. Memory-based algorithms do not work very well with sparse rating matrices: they scale poorly with the number of dimensions, and their predictions are not accurate for user/item matrix with few ratings. \\
The model-based methods are based on the assumptions that the preferences of a user can be inferred from a small number of hidden or latent factors. The most successful realizations of latent factor models are based on matrix factorization. This corresponds to a low-rank approximation of the rating matrix (Equation \ref{eq:ch04_matrix_factorization}), with the assumption of correlations between rows (or columns) guarantee the dimensionality reduction of the matrix itself.
\begin{equation}
    R \approx P \cdot Q
    \label{eq:ch04_matrix_factorization}
\end{equation}
where $P$ is a $m \times k$ matrix and $Q$ is a $k \times n$ matrix. The RS problems becomes a minimization problem, in which the decision variables are the elements of two low rank matrices whose product is as close as possible to the rating matrix. The number of optimization variables is still very high, $k\times (n+m)$ instead of $n\times m$, and the commonly used method is Stochastic Gradient Descent (SGD). The fine tuning of SGD raises the issue of hyperparameters optimization which is solved by BO in \cite{DBLP:journals/cms/GaluzziGCPA20}. \\
The main driver in the development of RSs has been so far the accuracy of recommendations. The increasing heterogeneity of users’ demands has led to multiple metrics such as diversity and novelty which might conflict with each other. Generally speaking, the increase of diversity and novelty will decrease accuracy. In addition, the increasing aware of ethical consideration has brawn to increasing importance of fairness. This brings to multi-objective optimization in which solutions are the elements of a Pareto set of non-dominated solutions. Matrix Factorization is not easily extended to take care of several objectives. The solutions proposed in \cite{DBLP:journals/corr/abs-1901-10757, DBLP:journals/tgrs/ZhuH16} that require the gradient, are based on linear sum scalarization, and do not guarantee a balanced approximation of the Pareto set.
To overcome the limitations of matrix factorization in Multi-Objectives Problem (MOP) Multi-Objectives Evolutionary Algorithms are considered in this thesis: many approaches have been proposed to this effect, focusing also on novelty and diversity \cite{DBLP:journals/complexity/LinWHMCLC18}. In this work a new method is proposed (Chapter \ref{ch08:recommender}) based on mapping the solution into a space of probability distributions.

%% file: chapters/chapter05.tex
\chapter{The Wasserstein Distance}
\label{ch05:wst}
There are many measures in the literature that can be used to compare probability distributions. Information theoretic based, like Kullback-Leibler and Jensen-Shannon, are the most used but can become undefined if the compared distributions do not have identical support. Other measures like the total variation or Hellinger distances do not provide a usable measure of distance for distributions without a significant overlap. \\
Wasserstein distances have a sound mathematical basis, they are generally well defined and provide an interpretable distance metric between distributions. Moreover, the Wasserstein distances are, at least in most conditions, differentiable which makes them more suitable for learning and optimization. These properties of the Wasserstein (WST) distances are built upon a deep mathematical framework which will be only hinted at in this work.

\section{Basic definitions}
\begin{figure}[h]
    \centering
    \includegraphics[scale = 0.8]{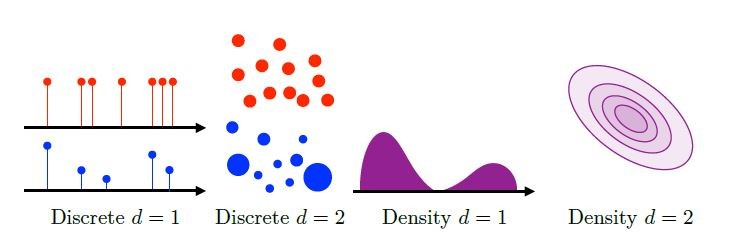}
    \caption{An example of discrete and continuous distributions in 1-dimensional and 2-dimensional case.}
    \label{fig:ch05_wst_discrete_density}
\end{figure}

\begin{figure}[h]
    \centering
    \includegraphics[scale = 0.7]{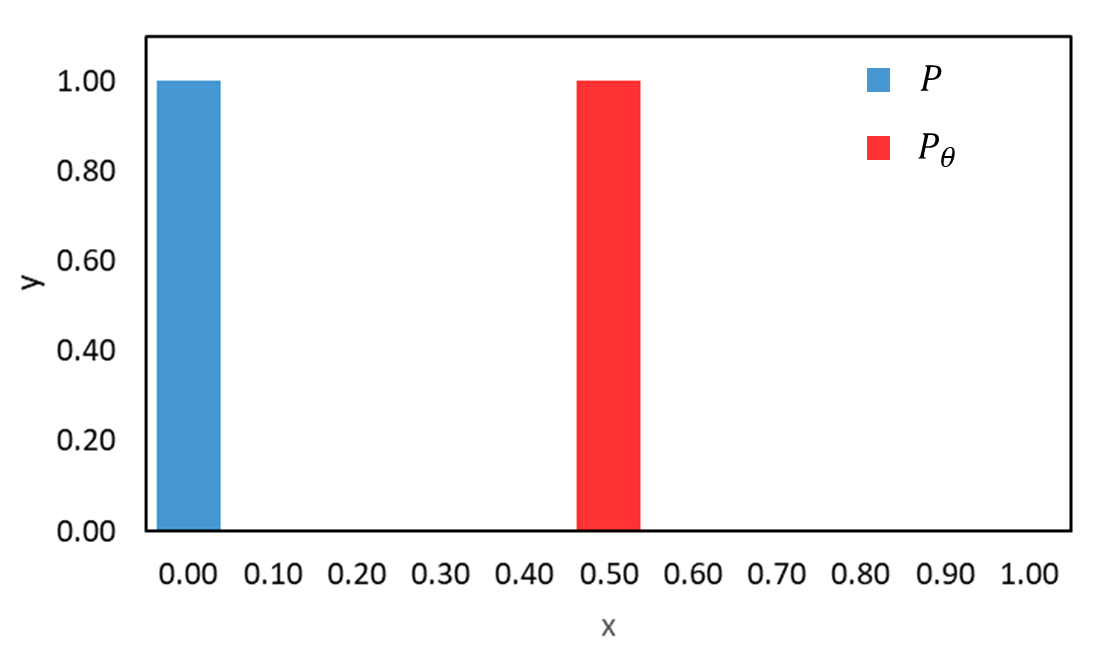}
    \caption{The two distributions $P$ and $P_{\theta}$. It is important to note that there is no overlap when $\theta \neq 0$.}
    \label{fig:ch05_example_wst}
\end{figure}

An important feature of the WST distance is that it can be applied to discrete, mixed and continuous distributions (Figure \ref{fig:ch05_wst_discrete_density}). \\
Consider first the continuous case. Given an exponent $p \ge 1$ let $f$ and $g$ be two probability distributions on $\mathbb{R}^d$ with finite $p$-moments, then the $p$-Wasserstein distance is (Equation \ref{eq:ch05_wst_cont}):
\begin{equation}
    \mathcal{W}_p (f, g) = \left( \inf_{\gamma \in \Gamma (f, g)} \int_{\mathcal{X} \times \mathcal{X}} d(x, y)^p d \gamma (x, y) \right)^{\frac{1}{p}}
    \label{eq:ch05_wst_cont}
\end{equation}
where $d(x, y)$ is also called ground distance (usually it is the Euclidean norm), $\Gamma(f,g)$ denotes the set of all joint distributions $\gamma (x,y)$ whose marginals are respectively $f$ and $g$, and $p \geq 1$ is an index.\\ 
 \\ \\ \\
There are some specific cases, very relevant in applications, where WST can be written in an explicit form. Let $F$ and $G$ be the cumulative distribution for the one-dimensional distributions $f$ and $g$ on the real line and $F^{-1}$ and $G^{-1}$ be their quantile functions (Equations \ref{eq:ch05_wst_quantile}).
\begin{equation}
    \mathcal{W}_p (f, g) = \left( \int_0^1 \left| F^{-1}(x) - G^{-1}(y) \right|^p \right)^{\frac{1}{p}}
    \label{eq:ch05_wst_quantile}
\end{equation}
Wasserstein distance is a measure of the distance between two probability distributions. It is also called Earth Mover’s Distance (EMD) from its informal interpretation as the minimum cost of moving and transforming a pile of sand in the shape of one probability distribution to the shape of the other distribution. The cost is quantified by the amount of sand moved times the moving distance. If the distribution domain is continuous the formula for the Earth Mover’s Distance is (Equation \ref{eq:ch05_emd}):
\begin{equation}
    \mathcal{W}_p (f, g) = \inf_{\gamma \in \Gamma (f, g)} \mathbb{E}_{(x, y) \in \gamma} \left[ \|x - y\|^p \right]^{\frac{1}{p}}
    \label{eq:ch05_emd}
\end{equation}
One joint distribution $\gamma(x, y) \in \Gamma(f, g)$ describes one transport plan: intuitively $\gamma(x, y)$ indicates how much mass must be transported from $x$ to $y$ in order to transform the distribution $f$ into the distribution $g$. Therefore, the marginal distribution over $x$ adds up to $\sum_x \gamma (x, y) = g(y)$ and analogously $\sum_y \gamma (x, y) = f(x)$. If $x$ is the starting point and $y$ the destination, the total amount of sand moved is $\gamma(x, y)$ and the traveling distance is $\|x - y\|$ and thus the total cost is $\gamma(x, y)\|x - y\|$. The expected cost averaged over all the $(x,y)$ pairs can be computed as (Equation \ref{eq:ch05_emd_cost}):
\begin{equation}
    \sum_{x, y} \gamma(x, y) \| x - y \| = \mathbb{E}_{(x, y) \sim \gamma}[\|x - y\|]
    \label{eq:ch05_emd_cost}
\end{equation}
The EMD is the cost of the optimal transport plan which is the minimum among the costs of all sand moving solutions.\\
The Wasserstein distance fits in the framework of optimal transport theory. It can be traced back to the work of Gaspard Monge (1781) \cite{monge1781memoire} and received its modern linear programming formulation by Lev Kantorovich (1958) \cite{custom:journals/Kantorovitch}. WST has been gaining increasing importance in several fields like Imaging \cite{DBLP:journals/tog/BonneelPC16}, Natural Language Processing (NLP) \cite{DBLP:conf/nips/HuangGKSSW16} and the generation of adversarial networks \cite{DBLP:conf/icml/ArjovskyCB17}. Some recent applications have been in the topic of Recommender Systems \cite{DBLP:conf/um/MengYLWC20, DBLP:conf/icml/BackursDIRW20, DBLP:journals/corr/abs-2102-03450}. The modelling flexibility and computational efficiency of the WST distance have been also shown in the design of neural architectural search \cite{DBLP:conf/nips/KandasamyNSPX18}. The formulation, computation and generalization of the WST distance require sophisticated mathematical models and raise challenging computational problems: important references are \cite{villani2008optimal, DBLP:journals/ftml/PeyreC19} which also give an up-to-date survey of numerical methods.\\
The Wasserstein distance has two key advantages. Even in the cases when the distributions are supported in different spaces, also without overlaps, WST can still provide a meaningful representation of the distance between distributions. Another advantage of WST is its differentiability. The former point will be exemplified in Chapter \ref{ch05:wst_discrete}; the latter point is illustrated in the following example (Figure \ref{fig:ch05_example_wst}). Let $Z=\mathcal{U}(0,1)$ be the uniform distribution on the unit interval. Let $P$ be the distribution of $(0,Z)$ ($0$ on the $x$-axis and the random variable $Z$ on the $y$-axis) and $P_\theta = (\theta, Z)$. \\
\begin{itemize}
    \item $KL(P, P_\theta) = +\infty$ if $\theta \neq 0$ and $0$ if $\theta = 0$
    \item $JS(P, P_\theta) = \log 2$ if $\theta \neq 0$ and $0$ if $\theta = 0$
    \item $\mathcal{W}(P, P_\theta) = \theta$ if $\theta \neq 0$ and $0$ if $\theta = 0$
\end{itemize}
Therefore, Wasserstein provides a smooth measure which is useful for any optimization and learning process using gradient descent \cite{DBLP:conf/icml/ArjovskyCB17}, in particular for generation of adversarial networks in deep learning \cite{li2020collaborative, DBLP:journals/prl/ZhangLZH20}.

\section{Wasserstein over discrete distributions}
\label{ch05:wst_discrete}
\begin{figure}[h]
    \centering
    \includegraphics[scale = 0.38]{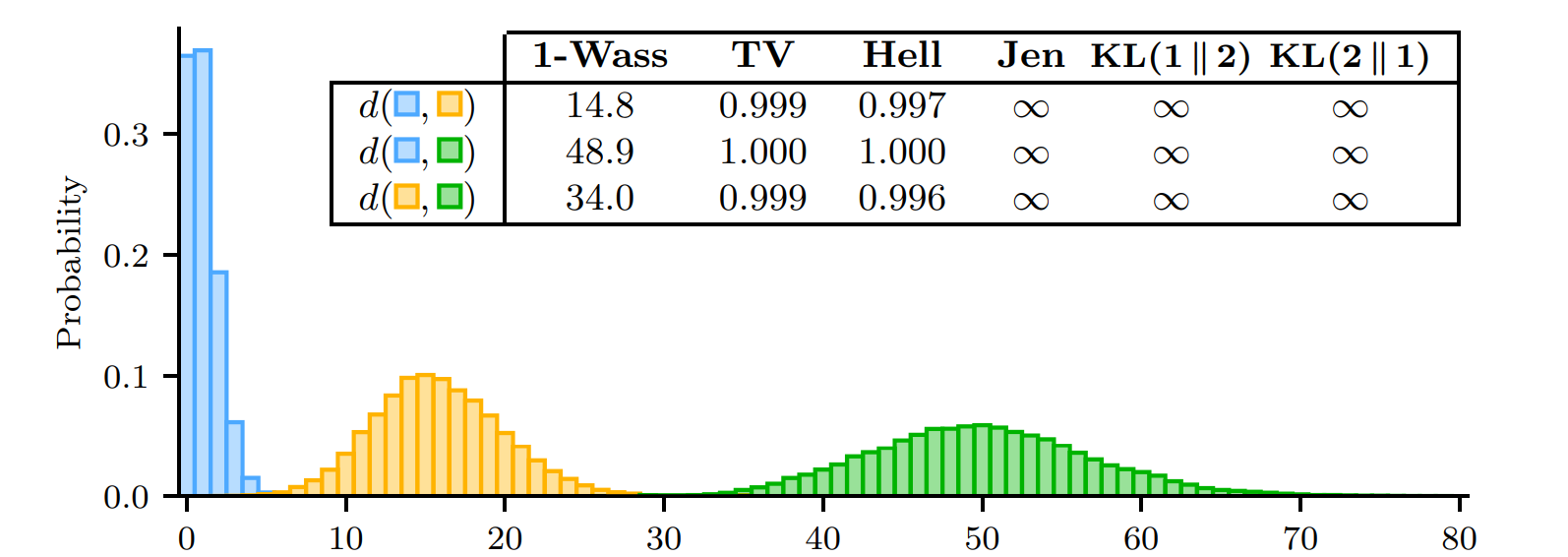}
    \caption{The table compares different probabilistic distances between the three histograms displayed in the figure. It is clear how the Wasserstein distance is the more interpretable one.}
    \label{fig:ch05_wst_vs_others}
\end{figure}

In the case of discrete distributions and specifically histograms, Wasserstein clearly displays its advantage over information theory based measures as shown in Figure \ref{fig:ch05_wst_vs_others} \cite{ocal2019parameter}. Each probability distribution is modelled as a histogram in which each bin has a weight and a coordinate in a multidimensional vector space. For instance, when measuring the distance between grey-scale images, the histogram weights are given by the pixel values and the coordinates are defined by the respective pixel positions.
 \\
The Earth Mover’s Distance (EMD) can be considered as the discrete version of the Wasserstein distance and can be used to quantify the affinity between discrete probability distributions. The distance between two histograms is calculated as the cost of transforming one into the other. Transforming a first histogram into a second one involves moving weights from the bins of the first histogram into the bins of the second, thereby constructing the second histogram from the first. The goal is to minimize the total distance travelled, where the pairwise distances between different histogram bins are computed based on their respective coordinates. This optimization problem is well studied in transportation theory. \\
Let $P$ be a discrete distribution specified by a set of support points $x_i$ with $i = 1,\ldots,m$ and their associated probabilities $w_i$ such that $\sum_{i=1}^m w_i = 1$ with $w_i \geq 0$ and $x_i \in M$ for $i = 1,\ldots,m$. Usually, $M=\mathbb{R}^d$ is the $d$-dimensional Euclidean space with the $L_p$ norm and $x_i$ are called the support vectors. $M$ can also be a symbolic set provided with a symbol-to-symbol similarity. \\
$P$ can also be written using the notation (Equation \ref{eq:ch05_p_discrete}):
\begin{equation}
    P(x) = \sum_{i=1}^m w_i \delta (x - x_i)
    \label{eq:ch05_p_discrete}
\end{equation}
where $\delta(\cdot)$ is the Kronecker delta. The WST distance between two distributions $P^{(1)} = \{ w_i^{(1)}, x_i^{(1)} \}$ with $i = 1,\ldots,m_1$ and $P^{(2)} = \{ w_i^{(2)}, x_i^{(2)} \}$ with $i = 1,\ldots,m_2$ is obtained by solving the following linear programming problem (Equation \ref{eq:ch05_emd_lp}):
\begin{equation}
    \mathcal{W}(P^{(1)}, P^{(2)}) = \min_{\gamma_{ij} \in \mathbb{R}^+} \sum_{i \in I_1, j \in I_2} \gamma_{ij} d \left( x^{(1)}_i, x^{(2)}_j \right)
    \label{eq:ch05_emd_lp}
\end{equation}
where $I_1 = \{1,\ldots,m_1\}$ and $I_2 = \{1,\ldots,m_2\}$ are two index sets such that (Equations \ref{eq:flow_constraint_1} and \ref{eq:flow_constraint_2}):
\begin{gather}
    \sum_{i \in I_1} \gamma_{ij} = w^{(2)}_j, \; \forall j \in I_2 \label{eq:flow_constraint_1}\\
    \sum_{j \in I_2} \gamma_{ij} = w^{(1)}_i, \; \forall i \in I_1 \label{eq:flow_constraint_2}
\end{gather}
The cost of transport between $x_i^{(1)}$ and $x_j^{(2)}$, $d(x_i^{(1)}, x_j^{(2)})$, is defined by the $p$-th power of the norm $\|x_i^{(1)}, x_j^{(2)}\|$, usually the Euclidean distance. \\
Equations \ref{eq:flow_constraint_1} and \ref{eq:flow_constraint_2} represent the in-flow and out-flow constraint, respectively. The terms $\gamma_{ij}$ are called matching weights between support points $x_i^{(1)}$ and $x_j^{(2)}$ or the optimal coupling for $P^{(1)}$ and $P^{(2)}$. The basic computation of Optimal Transport (OT) between two discrete distributions involves solving a network flow problem whose computation scales typically cubic in the sizes of the measure. The computation of EMD turns out to be the solution of a minimum cost flow problem on a bi-partite graph where the bins of $P^{(1)}$ are the source nodes and the bins of $P^{(2)}$ are the sinks while the edges between sources and sinks are the transportation costs. \\
In the case of one-dimensional histograms, the computation of WST can be performed by a simple sorting and the application of Equation \ref{eq:ch05_edm_1d}:
\begin{equation}
    \mathcal{W}(P^{(1)}, P^{(2)}) = \left( \frac{1}{n} \sum_{i = 1}^n \left| x_i^{(1)^*} - x_i^{(2)^*} \right| \right)
    \label{eq:ch05_edm_1d}
\end{equation}
where $x_i^{(1)^*}$ and $x_i^{(2)^*}$ are the sorted samples.

\section{Barycenters and Wasserstein clustering}
Probability distributions can be viewed as over an underlying feature domain. Multidimensional distributions are defined as histograms by partitioning the underlying domain into bins with a mass associate to each bin. Defining a distance between distributions requires a notion of distance between points in the underlying domain: this is called the ground distance. If the ground distance is a metric and the distributions have the same mass (which is true in the case of PDF), EMD is a metric as well. EMD does not suffer from arbitrary quantization problems due to rigid binning strategies. Therefore, it is robust to errors in the transformations that take raw data into the feature space.\\
Consider a set of $N$ discrete distributions, $P = \{ P^{(1)},\ldots,P^{(N)} \}$, with $P^{(k)} = \{ (w_i^{(k)}, x_i^{(k)}) : i = 1,\ldots,m_k \}$ and $k = 1,\ldots,N$, then, the associated barycenter, denoted with $ \bar{P} = \{ (\bar{w}_1, x_1),\ldots,(\bar{w}_m, x_m ) \}$, is computed as follows (Equation \ref{eq:barycenter}:
\begin{equation}
    \bar{P} = \argmin_{P} \frac{1}{N} \sum_{k=1}^N \lambda_k \mathcal{W} \left( P, P^{(k)} \right)
    \label{eq:barycenter}
\end{equation}
where the values $\lambda_k$ are used to weight the different contributions of each distribution in the computation. Without loss of generality, they can be set to $\lambda_k = \frac{1}{N} \; \forall \; k = 1,\ldots,N$.\\
The Wasserstein barycenter, also called the Frechet mean of distributions, appears to be a meaningful feature to represent the mean variation of a set of distributions and offers a useful synthesis of the structure of probability distributions, in particular:
\begin{itemize}
    \item It is sensitive to the underlying geometry. Consider three distributions $P^{(1)} = \delta_0$, $P^{(2)} = \delta_\varepsilon$ and $P^{(3)} = \delta_{100}$, then $\mathcal{W}(P^{(1)}, P^{(2)}) \approx 0$, $\mathcal{W}(P^{(1)}, P^{(3)}) \approx \mathcal{W}(P^{(2)}, P^{(3)}) \approx 100$. The distances Total variation, Hellinger and Kullback-Leibler take the value $1$, thus they fail to capture the intuition that $P^{(1)}$ and $P^{(2)}$ are close to each other while they are far away from $P^{(3)}$.
    \item It is shape preserving. Denote $P^{(1)},\ldots,P^{(N)}$ and assume that each $P^{(j)}$ can be written as a location shift of any other $P^{(i)}$, with $i \neq j$. Suppose that each $P^{(j)}$ is defined as $P^{(j)} = \mathcal{N}(\mu_j,\Sigma)$, then the barycenter has the closed form (Equation \ref{eq:ch05_bary_example}):
    \begin{equation}
        \bar{P} = \mathcal{N} \left( \frac{1}{N} \sum_{j = 1}^N \mu_j, \Sigma \right)
        \label{eq:ch05_bary_example}
    \end{equation}
    in contrast to the (Euclidean) average of the distributions $\frac{1}{N} \sum_{j = 1}^N P^{(j)}$.
\end{itemize}
Therefore, the concept of barycenter enables clustering among distributions, in a space whose metric is the Wasserstein distance. More simply, the barycenter in a space of distributions is the analogue of the centroid when the clustering takes place in a Euclidean space. The most common and well-known algorithm for clustering data in the Euclidean space is $k$-means. Since it is an iterative distance-based (aka representative based) algorithm, it is easy to propose variants of $k$-means by simply changing the distance adopted to create clusters, such as the Manhattan distance (leading to $k$-medoids) or any kernel allowing for non-spherical clusters (i.e., kernel $k$-means). The crucial point is that only the distance is changed, while the overall iterative two-step algorithm is maintained. This is also valid in the case of the Wasserstein $k$-means, where the Euclidean distance is replaced by WST and centroids are replaced by barycenters:
\begin{itemize}
    \item \textbf{Step 1 - Assign.} Given the current $k$ barycenters at iteration $t$, namely $\bar{P}_t^{(1)},\ldots,\bar{P}_t^{(k)}$, clusters $C_t^{(1)},\ldots,C_t^{(k)}$ are identified by assigning each one of the distributions $P^{(1)},\ldots,P^{(N)}$ to the closest barycenter (Equation \ref{eq:ch05_assign}):
    \begin{equation}
        C_t^{(i)} = \left\{ P^{(j)} \in P : \bar{P}_t^{(i)} = \argmin_{Q = \{ \bar{P}_t^{(1)},\ldots, \bar{P}_t^{(k)} \}} \mathcal{W}(Q, P^{(j)}) \right\}, \; \forall i = 1,\ldots,k
        \label{eq:ch05_assign}
    \end{equation}
    \item \textbf{Step 2 - Optimize.} Given the new composition of the clusters, update the barycenters (Equation \ref{eq:ch05_optimize}):
    \begin{equation}
        \bar{P}_{t + 1}^{(i)} = \argmin_Q \frac{1}{\left| C_t^{(i)} \right|} \sum_{P \in C_t^{(i)}} \mathcal{W}(Q, P)
        \label{eq:ch05_optimize}
    \end{equation}
    that comes directly from Equation \ref{eq:barycenter}.
\end{itemize}
As in $k$-means, a key point of Wasserstein $k$-means is the initialization of the barycenters. In the case that all the distributions in $P$ are defined on the same support, then they can be randomly initialized, otherwise, a possibility is to start from $k$ distributions randomly chosen among those in $P$. Finally, termination of the iterative procedure occurs when the result of the assignment step does not change any longer or a prefixed maximum number of iterations is achieved.

\section{Approximations and computational issues}
The issue of computational complexity of WST has been investigated also in the context of $k$-Nearest Neighbourhood ($k$-NN) first in \cite{indyk2003fast} then in \cite{DBLP:conf/icml/BackursDIRW20} where an approximate Nearest Neighbour Search is proposed for the $\mathcal{W}_1$ distance in the context of image retrieval. The ``ground'' set is a finite set of $\mathbb{R}^d$ and each distribution over $X$ has a finite support. Given a data set of $n$ distributions, the $k$-distributions closer to a target $v$ in the Wasserstein space are looked for. \\
Recently, Atasu and Mittelholzer \cite{DBLP:conf/icml/AtasuM19} proposed new approximate algorithms resulting in more accurate estimates that can be computed in linear time.

%% file: chapters/chapter06.tex
\chapter{MOEA with Wasserstein}
\label{ch06:moeawst}
This chapter presents the key algorithmic element of this thesis, a new evolutionary algorithm, namely MOEA/WST (Multi-Objective Evolutionary Algorithm with Wasserstein), based on a distributional representation of the individual.

\section{General framework}
\begin{figure}[h]
    \centering
    \includegraphics[width=1\linewidth]{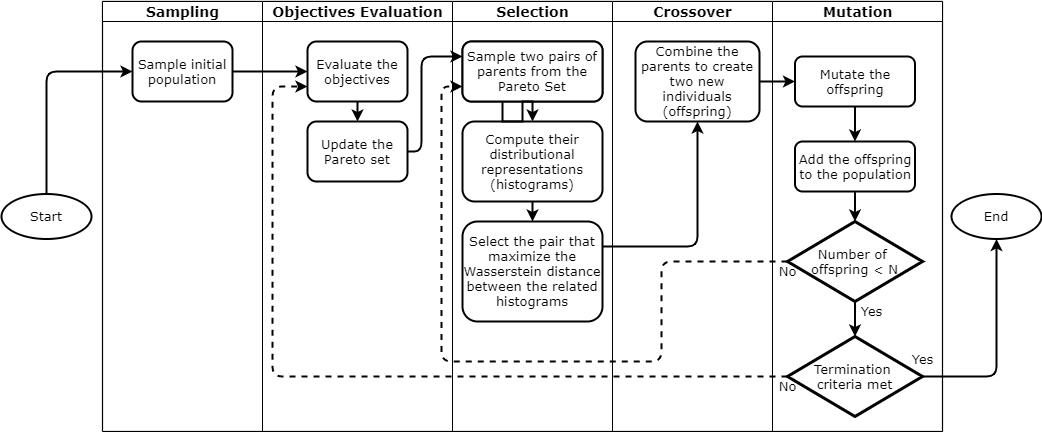}
    \caption{The general framework of MOEA/WST.}
    \label{fig:ch06_moea_wst}
\end{figure}

MOEA/WST starts sampling the initial population, i.e., a set of candidate solutions of the problem, also called individuals. For instance, in the case of sensor placement problem the individuals are the binary vectors encoding the sensor placements as explained in Chapter \ref{ch07:water}; in the case of recommender systems, the population is composed by a set of top-$L$ rating matrices as explained in Chapter \ref{ch08:recommender}. The initial individuals of the population are randomly sampled from the entire search space. The only constraint is that all the individuals have to be different (sampling without replacement). \\
In the next step for each candidate solution in the population all the objective functions are evaluated. Among this population only the non-dominated solutions are selected, i.e., the Pareto set as in Chapter \ref{ch02:pareto_evolutionary}. After that, the selection, crossover and mutation operators are used to generate the new offspring until a given number of new individuals are created. The newly generated individuals are added to the population and the next generation starts. The individuals that ``survive'' between a generation and the next one are chosen based on the dominance and the crowding distance as in NSGA-II. This process is repeated until a termination criteria is met, as a given number of generation or function evaluations. \\
Figure \ref{fig:ch06_moea_wst} schematize the optimization process of MOEA/WST. In particular, the innovation of MOEA/WST is given by the new selection operator that will be discussed in the next section.

\section{Wasserstein based selection operator}
\label{ch06:selection}
\begin{figure}[h]
    \centering
    \includegraphics[width=1\linewidth]{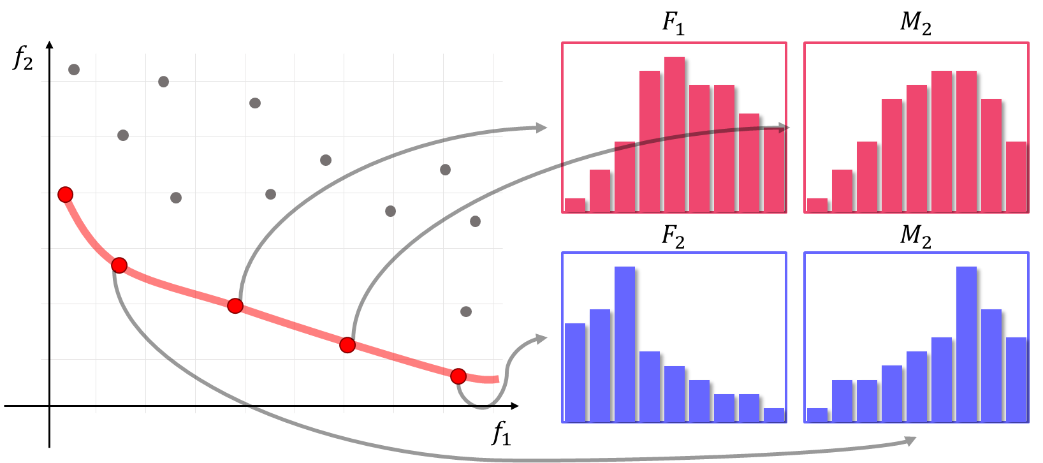}
    \caption{Two pairs of individuals are sampled from the Pareto front. As parents of the new offspring, the most different pair is chosen. In this case $F_2$ and $M_2$ will be the parents of the new offspring.}
    \label{fig:ch06_selection}
\end{figure}

In order to select the pairs of parents to be mated using the crossover operation, in MOEA/WST a new selection method has been introduced which takes place into a space of probability distributions. The idea that enables this operator is that, in some problem, the information in the input space can be represented in an intermediate space of distributions, before computing the objective functions. Two particular instances of this kind of problem will be discussed in Chapter \ref{ch07:water} and \ref{ch08:recommender} respectively. \\
This new selection operator, first, randomly sample from the actual Pareto set two pairs of individuals $(F_1,M_1)$ and $(F_2,M_2)$. To choose between this two pairs of candidate parents a binary tournament is computed. Differently from NSGA-II this tournament is not based on the non-dominated sorting and the crowding distance, but on the distance between parents in each pair (Figure \ref{fig:ch06_selection}). \\
Since in some situation a distance in the search space can be misleading, as deeply discussed in Chapter \ref{ch07:info_space}, in MOEA/WST, a distance between the distributional representation of the candidate parents has been considered. Therefore, assume that $h(F_i)$ and $h(M_i)$ are the distributions associated to $F_i$ and $M_i$ respectively. Then the pair $(F_i,M_i)$ that is chosen as the parents of the new offspring is the one in which (Equation \ref{eq:ch06_selection}):
\begin{equation}
    i = \argmax_{i \in \{ 1, 2 \}} d(h(F_i), h(M_i))
    \label{eq:ch06_selection}
\end{equation}
This favours exploration and diversification. Any probabilistic distance could be considered, but for the reason discussed in Chapter \ref{ch05:wst}, in MOEA/WST the Wasserstein distance has been used. \\
In the case in which the problem is constrained, if at least one individual of the pair of parents is not feasible the Constraint Violation (CV) is considered instead. \\
 \\
Then the pair of parents $(F_i,M_i)$ is chosen according to Equation \ref{eq:ch06_selection_cv}.
\begin{equation}
    i = \argmin_{i \in \{ 1, 2 \}} \left\{ CV(F_i), CV(M_i) \right\}
    \label{eq:ch06_selection_cv}
\end{equation}

\section{Problem specific crossover operator}
\label{ch06:crossover}
\begin{figure}[ht]
    \centering
    \includegraphics[width=1\linewidth]{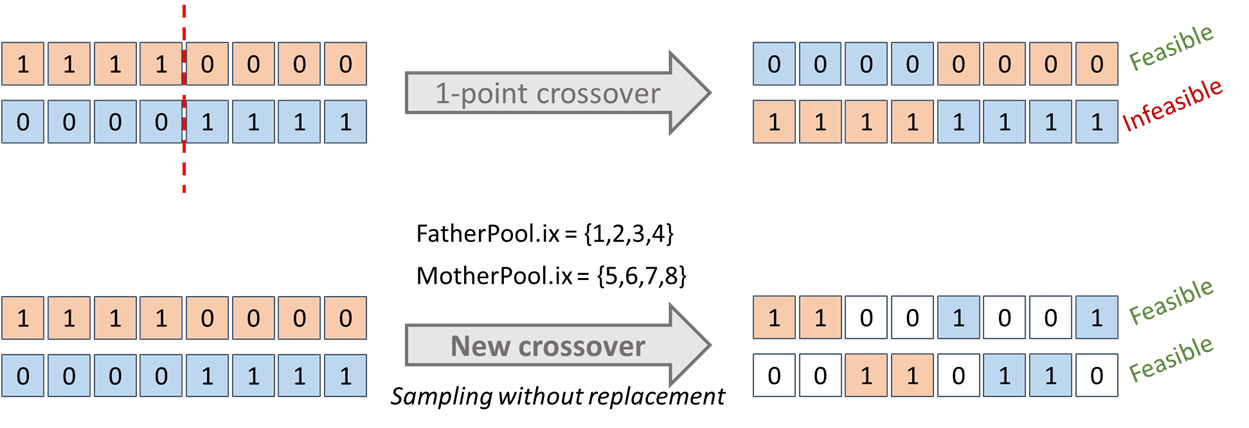}
    \caption{The comparison between one-point crossover and problem specific crossover.}
    \label{fig:ch06_crossover}
\end{figure}

The crossover is executed in the input space and for this reason it depends on the encoding of the individuals. Problem tailored crossover operators can substantially improve the performance \cite{DBLP:journals/eor/DebM17}. For the sensor placement problem, a problem specific crossover operator has been introduced which generates two ``feasible-by-design'' children from two feasible parents chosen according to the previous selection.\\
Denote with $x,x' \in \{0,1\} ^d$ two feasible parents and with $J$ and $J'$ the two associated sets $J=\{i: x_i=1\}$ and $J'=\{i: x'_i=1\}$. Finally, denote with $c,c' \in \{0,1\} ^d$ the two children of $x$ and $x'$. In turn, every child $c$ samples an index from $J$ and $c'$ from $J'$, respectively FatherPool and MatherPool without replacement. This guarantees to have no children with more than $p$ non-zero components. \\ \\ \\
Figure \ref{fig:ch06_crossover} shows an example comparing the behaviour of this problem specific crossover compared to a typical 1-point crossover. \\
 \\

%% file: chapters/chapter07.tex
\chapter{Water Distribution Networks}
\label{ch07:water}
This chapter focus on the analysis of Water Distribution Networks (WDNs), in terms of resilience and vulnerability, and on the problem of detecting contamination in the water flow. \\
The key element of this chapter is given by the representation of WDNs as discrete distributions which allows the definition of a new framework through the Wasserstein distance. This enables the use of MOEA/WST.

\section{Water networks}
\begin{figure}[h]
\centering
\begin{subfigure}{.32\textwidth}
  \centering
  \includegraphics[width=1\linewidth]{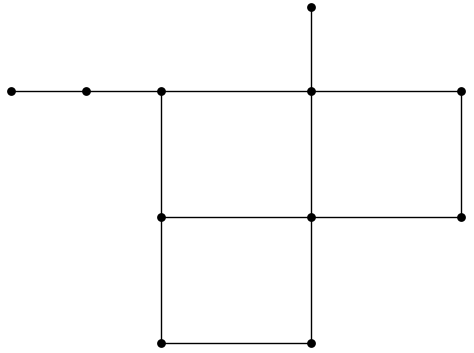}
  \caption{Net1}
  \label{fig:ch07_net1}
\end{subfigure}
\begin{subfigure}{.32\textwidth}
  \centering
  \includegraphics[width=1\linewidth]{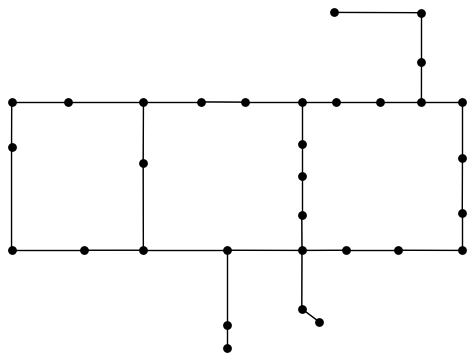}
  \caption{Hanoi}
  \label{fig:ch07_hanoi}
\end{subfigure}
\begin{subfigure}{.32\textwidth}
  \centering
  \includegraphics[width=1\linewidth]{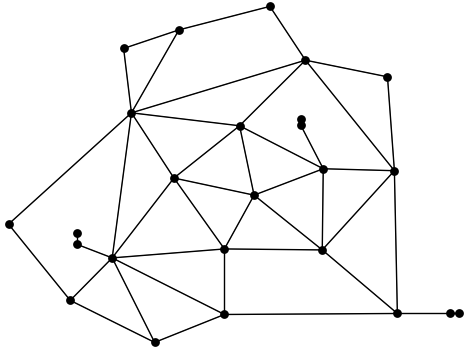}
  \caption{Anytown}
  \label{fig:ch07_anytown}
\end{subfigure}
\caption{A schematic representation of the three benchmark WDNs.}
\label{fig:bench_wdn_model}
\end{figure}

\begin{figure}[h]
\centering
\begin{subfigure}{.5\textwidth}
  \centering
  \includegraphics[width=.93\linewidth]{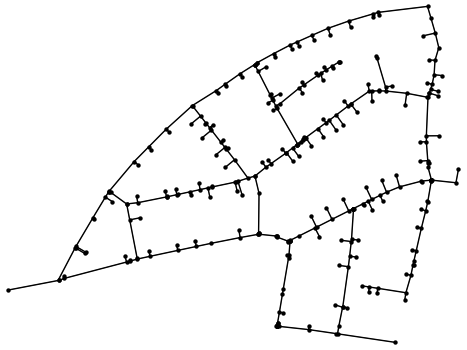}
  \caption{Neptun}
  \label{fig:ch07_neptun}
\end{subfigure}%
\begin{subfigure}{.5\textwidth}
  \centering
  \includegraphics[width=.93\linewidth]{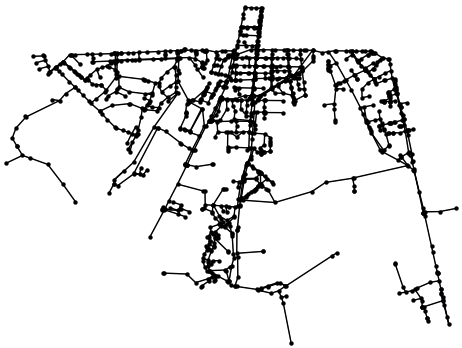}
  \caption{Abbiategrasso}
  \label{fig:ch07_abbiategrasso}
\end{subfigure}
\caption{A schematic representation of the two real word-size WDNs}
\label{fig:real_wdn_model}
\end{figure}

A Water Distribution Network (WDN) is a complex system aims to carry potable water from a water treatment plant to consumers in order to deliver water to satisfy residential, commercial, and industrial needs. \\
The main components of a WDN are pumps, junctions, and pipes. Additional components are tanks to store water and valves to isolate equipment, buildings, and other areas of the water system for repair as well as to control the direction and rate of flow.\\ \\
In this thesis five WDN models are considered. \\ \\
The following three are benchmarks models: 
\begin{itemize}
    \item \textbf{Net1} (Figure \ref{fig:ch07_net1}) is a small WDN provided by EPANET and WNTR, whose associated graph consists of 11 nodes (1 reservoir, 1 tank and 9 junctions) and 13 pipes.
    \item \textbf{Hanoi} (Figure \ref{fig:ch07_hanoi}) is a benchmark used in the literature. It has 32 nodes (1 reservoir and 31 junctions) and 34 pipes.
    \item \textbf{Anytown} (Figure \ref{fig:ch07_anytown}) is another benchmark WDN, whose graph consists of 25 nodes (1 reservoir, 2 tanks and 22 junctions) and 46 edges (3 pumps and 43 pipes).
\end{itemize}
The other two networks are real world-size WDNs:
\begin{itemize}
    \item \textbf{Neptun} (Figure \ref{fig:ch07_neptun}) is the WDN of the Romanian city of Timisoara, with an associated graph of 333 nodes (1 reservoir and 332 junctions) and 339 edges (27 valves and 312 pipes).
    \item \textbf{Abbiategrasso} (Figure \ref{fig:ch07_abbiategrasso}) refers to a pressure management zone in Milan (namely, Abbiategrasso) with an associated graph consisting of 1213 nodes (1 reservoir and 1212 junctions) and 1393 edges (4 pumps, 4 valves and 1385 pipes).
\end{itemize}
These lasts two networks have been a pilot in the European project \textit{Icewater} \cite{fantozzi2014ict}.

\section{Wasserstein for resilience evaluation}
\begin{figure}[h]
\centering
    \begin{subfigure}{.49\textwidth}
        \centering
        \includegraphics[width=0.88\linewidth]{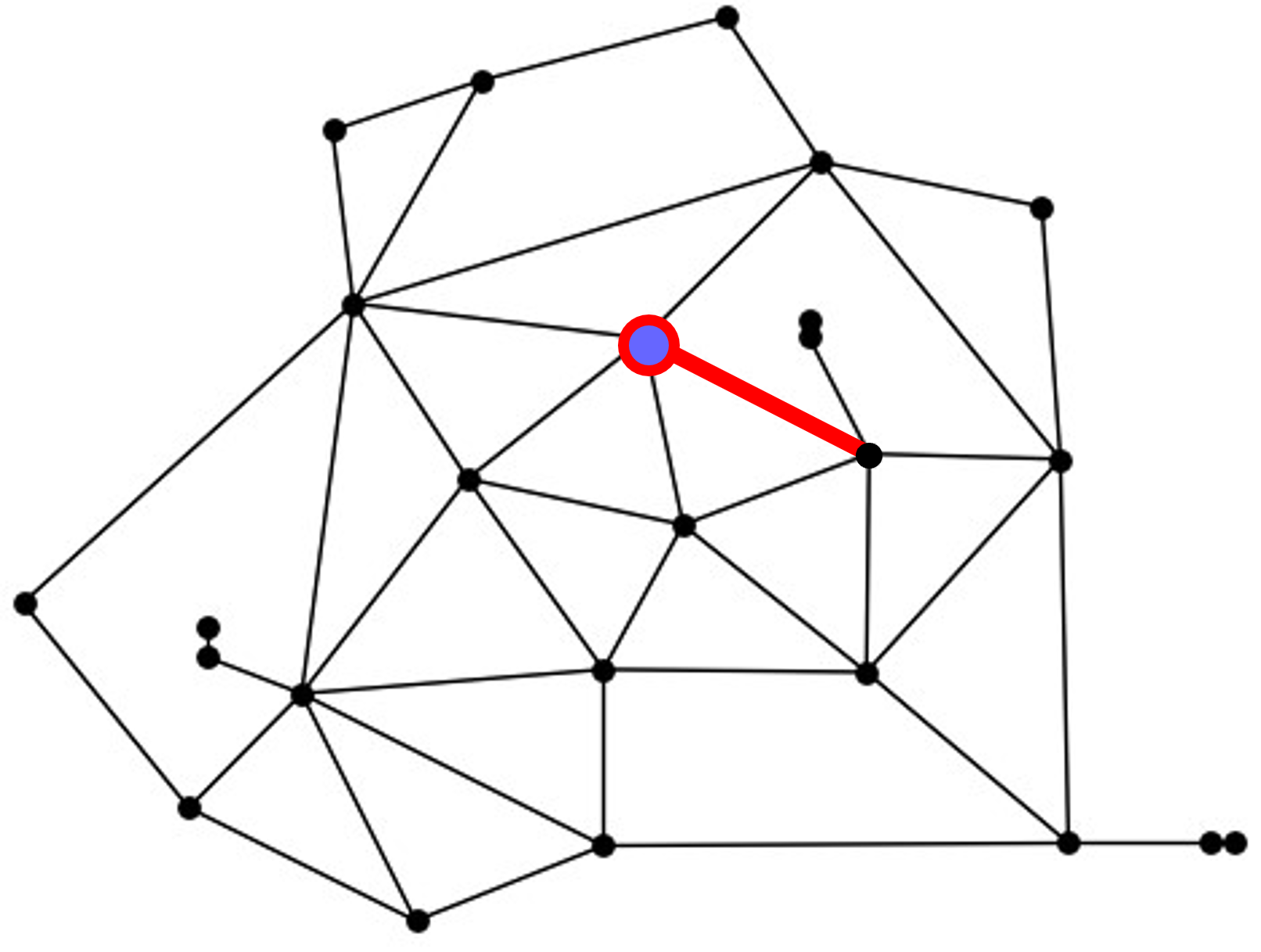}
        \caption{$G'$ is obtained removing the red edge of Anytown ($G$).}
    \end{subfigure}
    \begin{subfigure}{.49\textwidth}
        \centering
        \includegraphics[width=1\linewidth]{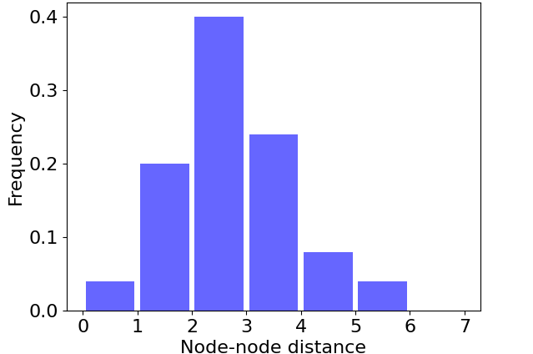}
        \caption{The node-node distances distribution of the highlighted node.}
    \end{subfigure}
\caption{The Anytown water distribution network and an example of node-node distances distribution.}
\label{fig:ch07_anytown_removed_edge}
\end{figure}

\begin{figure}[h]
    \centering
    \includegraphics[width=1\linewidth]{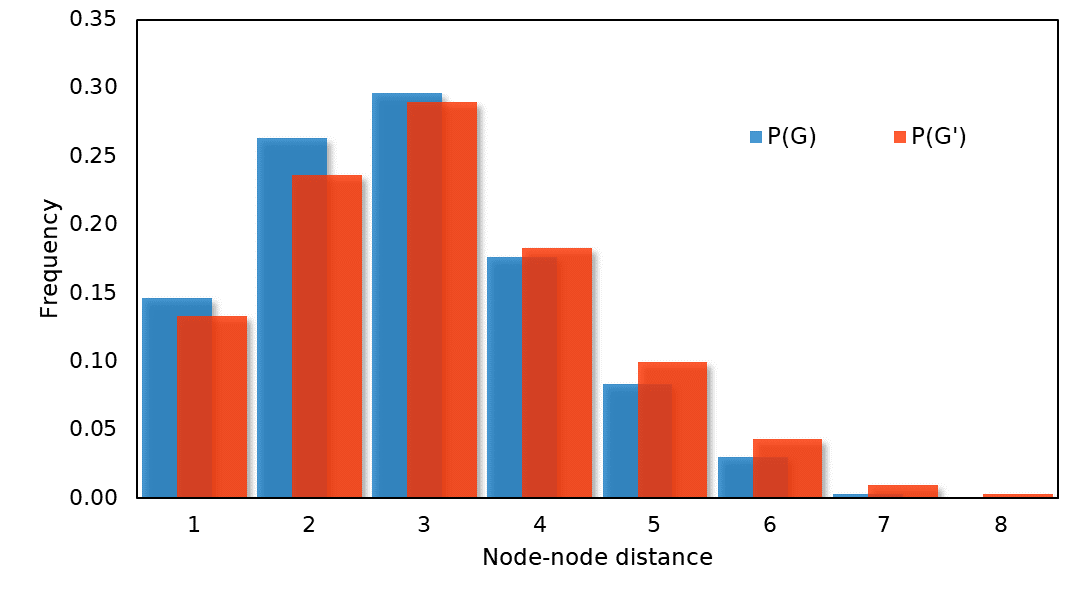}
    \caption{The node-node distance distributions at network level of $G$ and $G'$.}
    \label{fig:ch07_different_distr}
\end{figure}

Given a graph $G=(V,E)$ it is possible to associate to each node $i = 1,\ldots,n$ a discrete probability distribution (Equation \ref{eq:ch07_node_distr}) as the fraction of nodes which are connected to $i$ at a distance $k$:
\begin{equation}
    P_k (i) = \frac{n_{i,k}}{n - 1}
    \label{eq:ch07_node_distr}
\end{equation}
Figure \ref{fig:ch07_anytown_removed_edge} displays the distribution of the highlighted node of Anytown. \\
 \\
The support of this distribution is $1,\ldots,D(G)$ where $D(G)$ is the diameter of $G$. The distance distribution over the whole network is shown in Equation \ref{eq:ch07_net_distr}.
\begin{equation}
    P_k (G) = \frac{1}{n} \sum_{i = 1}^{n} \frac{n_{i,k}}{n - 1} = \frac{1}{n} \sum_{i = 1}^n P_k(i)
    \label{eq:ch07_net_distr}
\end{equation}
Let $G'$ be the graph without the red edge and consider the distributions $P = P_k (G)$ (Equation \ref{eq:ch07_res_ex1}) and $P' = P_k (G')$ (Equation \ref{eq:ch07_res_ex2}). In the case of Anytown $D(G)=8$ and the two histograms are displayed in Figure \ref{fig:ch07_different_distr}
\begin{gather}
    P(G) = [0.147, 0.263, 0.297, 0.177, 0.083, 0.030, 0.003, 0] \label{eq:ch07_res_ex1} \\
    P(G') = [0.133, 0.237, 0.290, 0.183, 0.100, 0.043, 0.010, 0.003] \label{eq:ch07_res_ex2}
\end{gather}
The most widely used distance measure is the Kullback-Leibler (KL) divergence (Equation \ref{eq:ch07_kl}) which has the drawback of being asymmetric and possibly infinite when there are points $x$ such that $P(x)=0$ and $P'(x) \ge 0$.
\begin{equation}
    KL \left( P \left| \frac{P + P'}{2} \right. \right) = \int P \log \frac{2P}{P + P'} dx
    \label{eq:ch07_kl}
\end{equation}
The Jensen-Shannon (JS) divergence (Equation \ref{eq:ch07_js}) is built on KL and is symmetric and always definite.
\begin{equation}
    JS(P, P') = \frac{1}{2} KL \left( P \left| \frac{P + P'}{2} \right. \right) + \frac{1}{2} KL \left( P' \left| \frac{P + P'}{2} \right. \right)
    \label{eq:ch07_js}
\end{equation}
 \\
The use of Jensen-Shannon divergence in computing the dissimilarity between networks has been considered in \cite{schieber2017quantification}. To obtain a metric is often used the following form (Equation \ref{eq:ch07_js_metric}):
\begin{equation}
    \mathcal{JS} (P, P') = \sqrt{JS(P, P')} \in [0, 1] 
    \label{eq:ch07_js_metric}
\end{equation}
In the considered problem the network space is given by the basic network $G$ and the subgraphs obtained by the removal of one or more edges. The elements of this space are represented as probability distributions of node-node distances (Equation \ref{eq:ch07_node_distr}) and aggregated into a distributional representations of the basic network and of the subgraphs (Equation \ref{eq:ch07_net_distr}). Given the drawbacks of KL and JS illustrated in Chapter \ref{ch05:wst}, in the space of distributional representation the Wasserstein distance is used as introduced in \cite{ponti2021novel}. The result of this computation is mapped back into the network space as measures of network dissimilarity and labels of criticality of individual components.

\subsubsection{Results on real-word WDN}
\begin{figure}[h]
\centering
    \begin{subfigure}{.49\textwidth}
        \centering
        \includegraphics[width=1\linewidth]{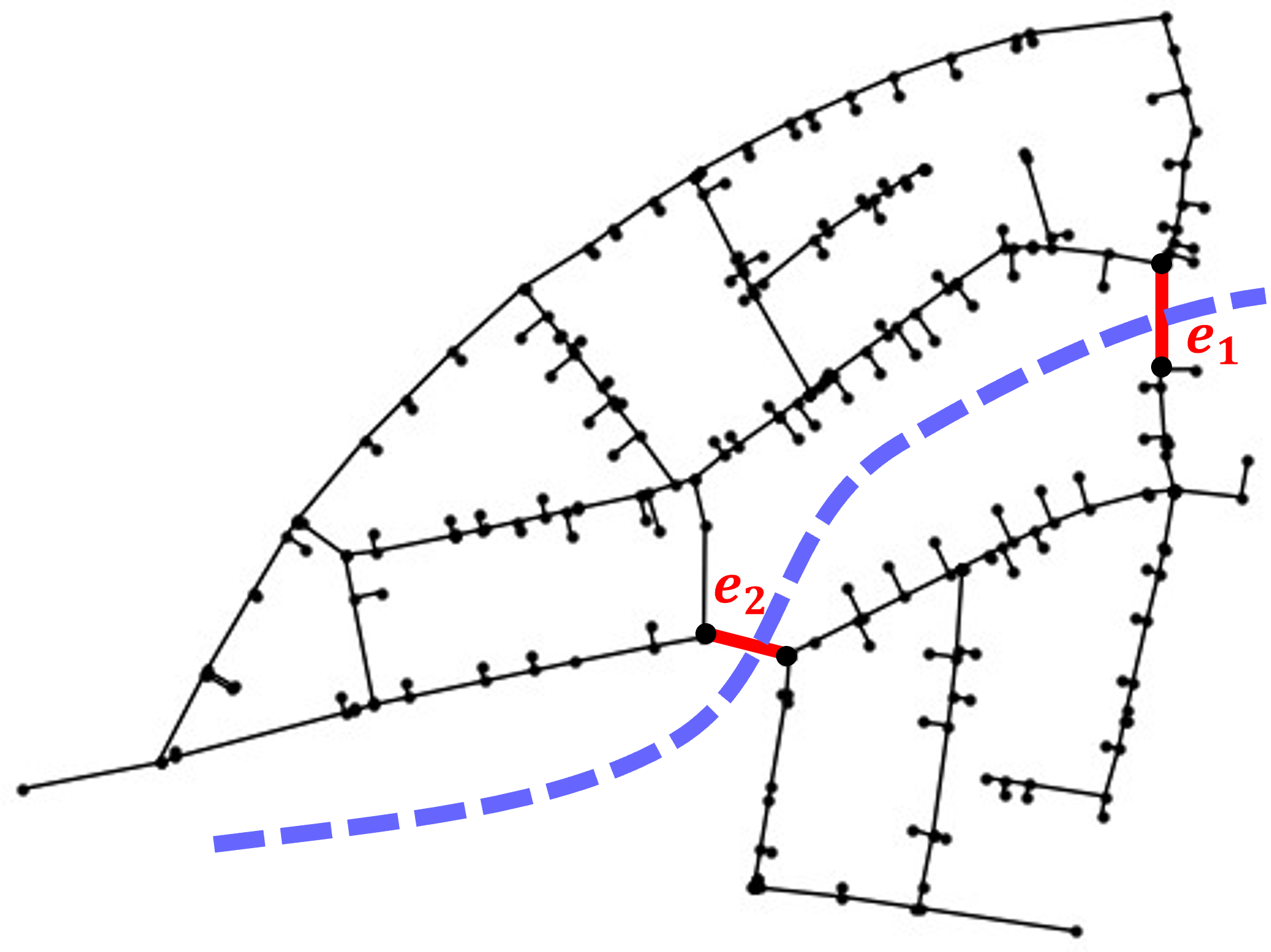}
        \caption{The two critical edges (red) whose simultaneous removal generates a disconnection in Neptun.}
    \end{subfigure}
    \begin{subfigure}{.49\textwidth}
        \centering
        \includegraphics[width=1\linewidth]{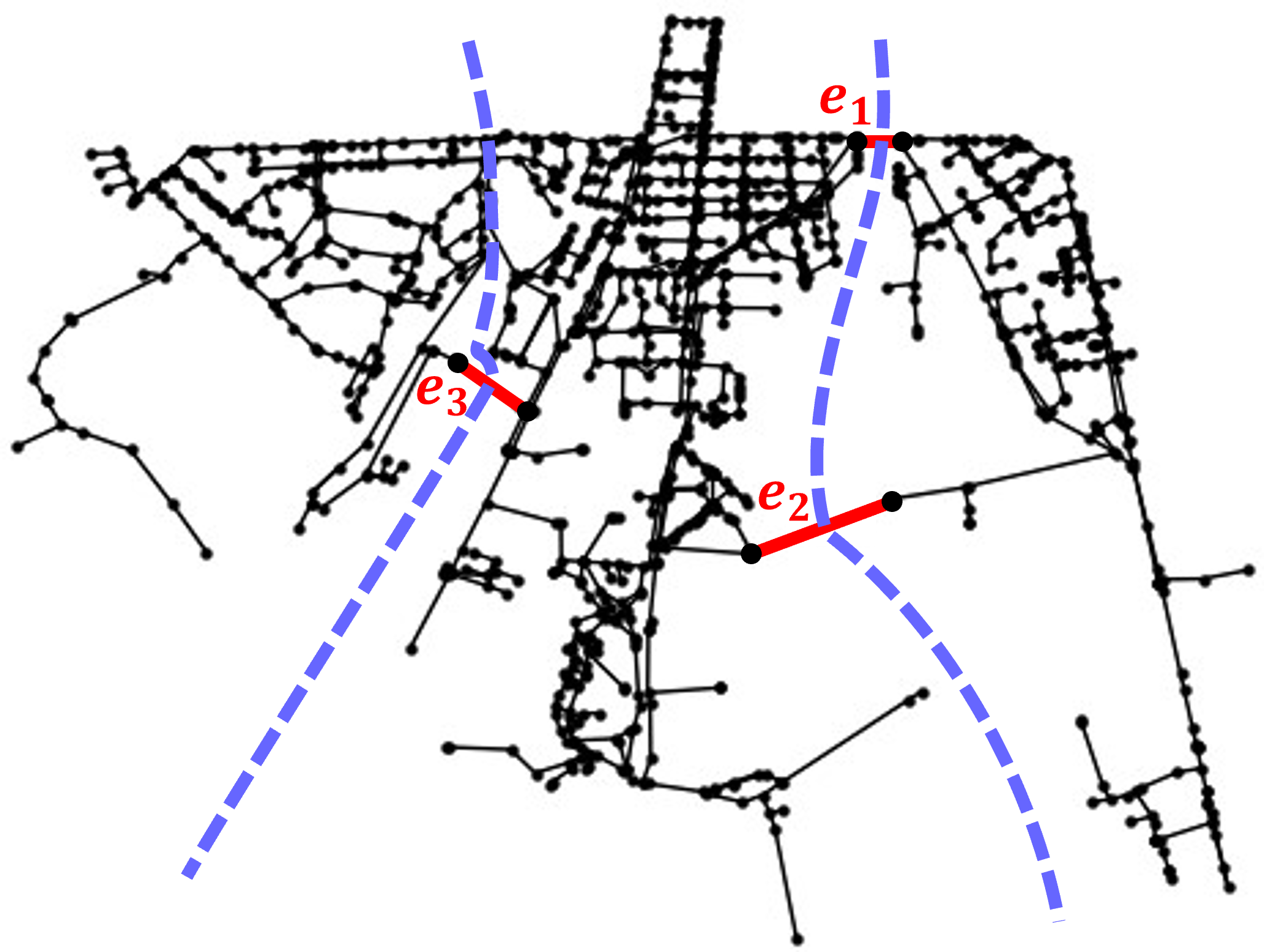}
        \caption{The three critical edges (red) whose simultaneous removal generates a disconnection.}
    \end{subfigure}
\caption{The spectral clustering results over the two WDN, Neptun (with $k = 2$) and Abbiategrasso (with $k=3$).}
\label{fig:ch07_wdn_cluster}
\end{figure}

\begin{figure}[h]
\centering
    \begin{subfigure}{.49\textwidth}
        \centering
        \includegraphics[width=1\linewidth]{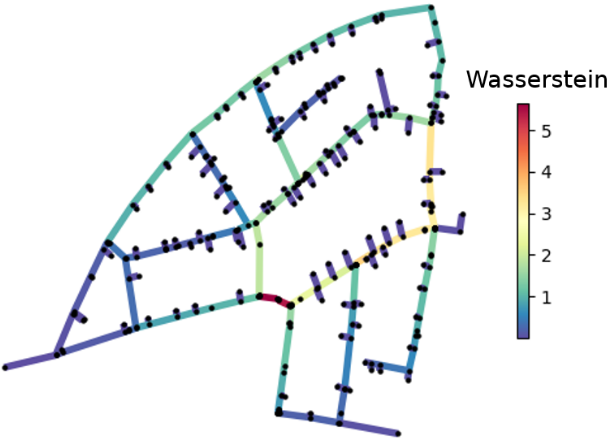}
        \caption{Heatmap of the Wasserstein distance for all edges in Neptun.}
    \end{subfigure}
    \begin{subfigure}{.49\textwidth}
        \centering
        \includegraphics[width=1\linewidth]{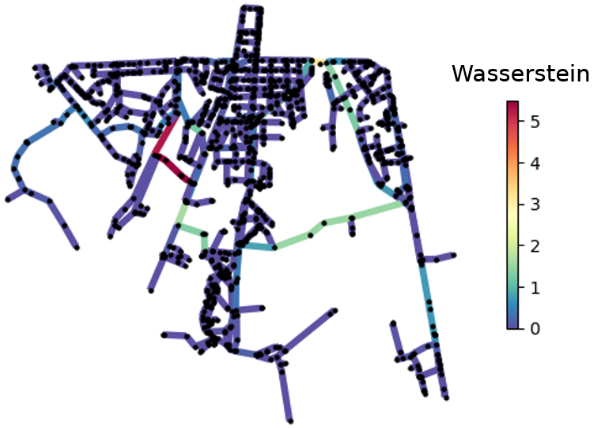}
        \caption{Heatmap of the Wasserstein distance for all edges in Abbiategrasso.}
    \end{subfigure}
\caption{The color of each edge $e$ depends on $\mathcal{W}(G, G\setminus \{e\})$, i.e., the Wasserstein distance between the original graph $G$ and the one obtained removing the edge $G\setminus \{e\}$.}
\label{fig:ch07_wdn_heatmap}
\end{figure}

Firstly, the centrality measures of the networks, Neptun and Abbiategrasso, are analysed (Table \ref{tab:ch07_centrality}).  This two real-world WDNs are effectively planar and ``almost'' regular. This can be due to the fact that their structure is strongly constrained by spatial characteristics making a classification based on nodal degree distribution less meaningful. \\
\begin{table}[ht]
\centering
\caption{Centrality measure (defined in Appendix \ref{apx:centrality}) of the two WDN, Neptun and Abbiategrasso.}
\begin{tabular}{lcc} 
    \hline
    \textbf{Measure} & \textbf{Neptun} & \textbf{Abbiategrasso} \\ 
    \hline
    \hline
    Diameter & 57 & 83 \\ 
    Characteristic path length & 23.7613 & 30.6126 \\ 
    Density & 0.0061 & 0.0019 \\ 
    Link-per-node ratio & 0.0019 & 1.1467 \\ 
    Central point dominance & 0.2432 & 0.3100 \\ 
    Clustering coefficient & 0.0000 & 0.0055 \\ 
    \hline
\end{tabular}
\label{tab:ch07_centrality}
\end{table} \\
The second step in the analysis is clustering in order to identify the specific edges whose removal induces a disconnection of the network (Figure \ref{fig:ch07_wdn_cluster}). \\
The number of clusters $k$ is set according to context information about the districtualization adopted by the water utility. Failures affecting also only one pipe may imply a reduction in the efficiency of the network and an increase in vulnerability.
 \\
Once the edges identified by the clustering are removed the efficiency and vulnerability metrics are computed (Table \ref{tab:ch07_efficiency}). \\
Then, the probabilistic distances between the original networks and the once without the edges identified by cluster are computed (Table \ref{tab:ch07_distances}). \\
The results reported in Table \ref{tab:ch07_efficiency} and \ref{tab:ch07_distances}, which are quite unique in the literature given the size of the networks analysed, demonstrate that probabilistic distance measures show better capacity to discriminate between different networks not only globally but also edge-wise. \\
\begin{table}[ht]
\centering
\caption{Efficiency and vulnerability metrics (defined in Appendix \ref{apx:vulnerability}) of the two WDN, Neptun and Abbiategrasso. The algebraic connectivity $\lambda_2$ is defined as the second smallest eigenvalue of the Laplacian associated to the graph; $\lambda_2 = 0$ means that the graph is disconnected.}
\begin{tabular}{lcccc} 
    \hline
    \textbf{Neptun} & $\pmb{E}$ & $\pmb{V_{MEAN}}$ & $\pmb{V_{MAX}}$ & \textbf{Algebraic connectivity} \\ 
    \hline
    \hline
    $G$                             &    0.068608 &	0.018927 &	0.072646 &	0.0018 \\ 
    $G \setminus \{e_1\}$           &    0.065390 &	0.024181 &	0.211362 &	0.0007 \\ 
    $G \setminus \{e_2\}$           &    0.064486 &	0.024796 &	0.194813 &	0.0006 \\ 
    $G \setminus \{e_1, e_2\}$      &    0.051924 &	0.016642 &	0.068246 &	0.0000 \\ 
    \hline
    \textbf{Abbiategrasso} & $\pmb{E}$ & $\pmb{V_{MEAN}}$ & $\pmb{V_{MAX}}$ & \textbf{Algebraic connectivity} \\ 
    \hline
    \hline
    $G$                                 &   0.047557 &	0.003436 &	0.150390 &	0.0004 \\ 
    $G \setminus \{e_1\}$               &   0.045019 &	0.003935 &	0.181174 &	0.0003 \\ 
    $G \setminus \{e_2\}$               &   0.046385 &	0.003642 &	0.205294 &	0.0004 \\ 
    $G \setminus \{e_3\}$               &   0.040405 &	0.002628 &	0.060728 &	0.0000 \\ 
    $G \setminus \{e_1, e_2, e_3\}$     &   0.031077 &	0.002251 &	0.057007 &	0.0000 \\ 
    \hline
\end{tabular}
\label{tab:ch07_efficiency}
\end{table} \\
\begin{table}[ht]
\centering
\caption{Probabilistic distances and loss of efficiency (defined in Appendix \ref{apx:vulnerability}) between the original networks (Neptun and Abbiategrasso) and the once obtained removing some edges.}
\begin{tabular}{lccc} 
    \hline
    \textbf{Neptun} & $\pmb{\mathcal{JS}}$ & $\pmb{\mathcal{W}}$ & \textbf{Loss of efficiency} \\ 
    \hline
    \hline
    $G \setminus \{e_1\}$           &    0.1677 &	3.3183 &	0.0469 \\ 
    $G \setminus \{e_2\}$           &    0.2456 &	5.4704 &	0.0601 \\ 
    $G \setminus \{e_1, e_2\}$      &    0.3286 &	6.5542 &	0.2432 \\ 
    \hline
    \textbf{Abbiategrasso} & $\pmb{\mathcal{JS}}$ & $\pmb{\mathcal{W}}$ & \textbf{Loss of efficiency} \\ 
    \hline
    \hline
    $G \setminus \{e_1\}$               &   0.0935 &	3.1040 &	0.0534 \\ 
    $G \setminus \{e_2\}$               &   0.0528 &	1.5170 &	0.0246 \\ 
    $G \setminus \{e_3\}$               &   0.1843 &	5.4871 &	0.1504 \\ 
    $G \setminus \{e_1, e_2, e_3\}$     &   0.3633 &	8.8845 &	0.3465 \\ 
    \hline
\end{tabular}
\label{tab:ch07_distances}
\end{table} \\
To highlight this fact the heatmap of edge wise criticality, as Wasserstein distance, are shown in Figure \ref{fig:ch07_wdn_heatmap}. \\
This analysis framework supports decision making at design stage, to simulate alternative network layouts of different robustness, and also at operational stage where the decision to be taken can be, which nodes/edges are to temporarily be removed for maintenance and rehabilitation. Indeed, critical tasks of WDN management can be supported by just using topological and geometric information. The analysis framework also helps for the efficient and automatic definition of district metered areas and to facilitate the localization of water losses through the definition of an optimal network partitioning.\\
The modelling and algorithmic framework platform developed can be straightforwardly translated to many networked infrastructures among which power grids, transit networks but also global supply chains networks whose vulnerability has been exposed in the recent COVID crisis.

\section{Hydraulic and quality simulation}
Water Distribution Networks face multiple challenges such as aging infrastructure, natural disasters, terrorist attack and much more. All these problem can potentially disrupt a large portion of a water system causing damage to infrastructure and outages to customers. Simulation and analysis tools can help to explore the capacity of a WDN to handle disruptive events and guide the necessary operations to make a system more resilient over time.\\
The Water Network Tool for Resilience (WNTR) \cite{klise2018overview} is a Python package designed to simulate and analyse resilience of WDNs. WNTR is based on EPANET 2.0, which is a tool to simulate flowing of drinking water constituents within a WDN. WNTR contains an hydraulic and water quality simulator that tracks the flow over time of a contaminant injected in a specific location. \\
The simulation is computationally costly and scales linearly with the inverse of the simulation timestep. To consider different location in which the contaminant can be injected, it is necessary to run one simulation for each contamination event. The result of these simulations is a matrix, the so-called \textit{Trace Matrix}, which contains the percentage of contaminant in a specific location for each event and simulation time. From the \textit{Trace Matrix} is possible to extract the detection times matrix that reports, for each node considered as a possible sensor location, the detection times of all the different contamination events. The detection takes place when the percentage of contaminant exceeds a given threshold (in the computation case $\tau \geq 10\%$).

\section{The problem of sensors placement}
\begin{figure}[h]
    \centering
    \includegraphics[width=0.6\linewidth]{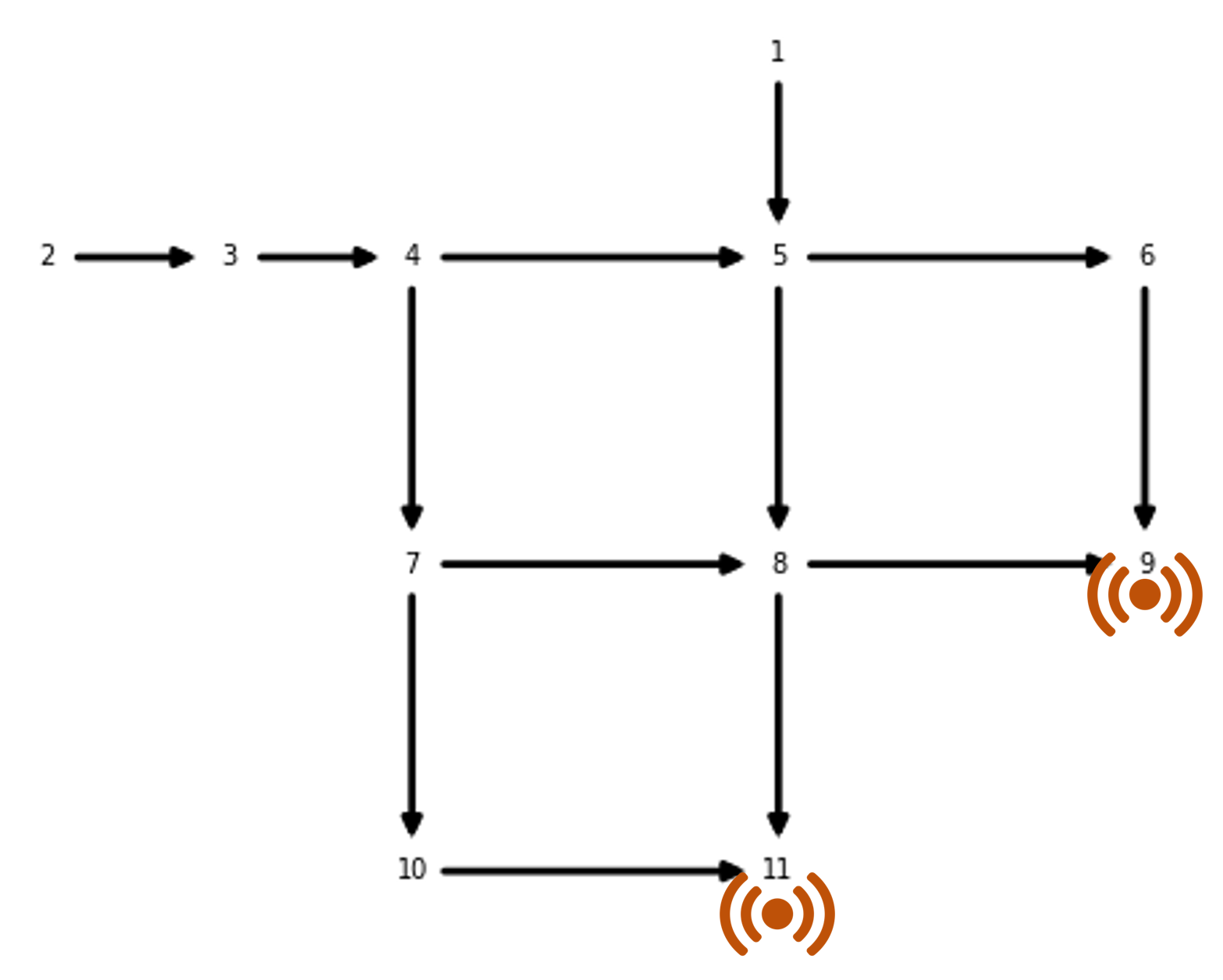}
    \caption{Example of a sensor placement in Net1. Sensors are placed at nodes $9$ and $11$.}
    \label{fig:ch07_sp_net1}
\end{figure}

Consider a graph $G = (V, E)$ and a set of possible locations for placing sensors $L \subseteq V$ . Thus, a Sensor Placement (SP) is a subset of sensor locations, with the subset’s size less or equal to $p$ depending on the available budget. A SP is represented by a binary vector $s \in \{0, 1\}^{|L|}$  whose components are $s_i = 1$ if a sensor is located at node $i$, $s_i = 0$ otherwise. Therefore, a SP is given by the nonzero components of $s$. An example, considering Net1, is given by $s = [0,0,0,0,0,0,0,0,1,0,1]$ which means that two sensors are placed respectively at nodes $9$ and $11$, as shown in Figure \ref{fig:ch07_sp_net1}. \\
 \\
For a WDN the vertices in $V$ represent junctions, tanks, reservoirs or consumption points, and edges in $E$ represent pipes, pumps, and valves. \\
Let $A \subseteq V$ denote the set of contamination events $a \in A$ (i.e., the locations in which the contaminant can be injected) which must be detected by a sensor placement $s$, and $d_{ai}$ the impact measure associated to a contamination event $a$ detected by the $i$-th sensor. A probability distribution is placed over possible contamination events associated to the nodes. In the computations assume – as usual in the literature – a uniform distribution, but in general different distributions are also possible. \\
In this thesis a general model of sensor placement is considered (Equation \ref{eq:sp_formulation_generic}).
\begin{equation}
    \begin{aligned}
    & \min_{s \in \{0,1\}^{|L|}}
    & & f_1(s) = \sum_{a \in A}{\alpha_a \sum_{i=1}^{|L|}{d_{ai}x_{ai}}} \\
    & \text{s.t.}
    & & \sum_{i=1}^{|L|}{s_i \leq p}, \; s_i \in \{0,1\}
    \end{aligned}
\label{eq:sp_formulation_generic}
\end{equation}
where:
\begin{itemize}
    \item $\alpha_a$ is the probability for the contaminant to enter the network at node $a$.
    \item $d_{ai}$ is the impact for a sensor located at node $i$ to detect the contaminant introduced at node $a$.
    \item $x_{ai}$ is an indicator variable assuming value $1$ if $s_i=1$ and $i$ is the first sensor in the placement $s$ detecting the contaminant injected at node $a$, i.e., the one with the minimum possible impact; $0$ otherwise.
\end{itemize}
In this study assume that all the events have the same chance of happening, that is $\alpha_a = \frac{1}{|A|}$, therefore $f_1(s)$ can be rewritten as (Equation \ref{eq:f1_simple}):
\begin{equation}
    f_1(s) = \frac{1}{|A|}\sum_{a \in A}{\hat{t}_a}
\label{eq:f1_simple}
\end{equation}
where $\hat{t}_a = \sum_{i = 1}^{|L|}{d_{ai}x_{ai}}$.\\
As a measure of risk, the standard deviation is considered (Equation \ref{eq:sp_std}).
\begin{equation}
    f_2(s) = STD_{f_1}(s) = \sqrt{\frac{1}{|A|} \sum_{a \in A}{(\hat{t}_a - f_1(s))^2}}
\label{eq:sp_std}
\end{equation}
This model can be specialized to different objective functions as:
\begin{itemize}
    \item \textbf{Detection time:} the impact $d_{ai}$ is the minimum detection time ($MDT$). For each event $a$ and sensor placement $s$ the $MDT$ is defined as $MDT_a = \min_{i : s_i = 1}{d_{ai}}$.
    \item \textbf{Volume of contaminated water:} the impact $d_{ai}$ represents the amount of contaminated water consumed prior to detection for scenario $a$ and sensor located in $i$.
    \item \textbf{Probability of detection:} the impact $d_{ai}$  represents the probability for a sensor in node $i$ to detect an event in $a$ during the simulation period. $d_{ai}$ is modelled as a Bernoulli random variable assuming value $1$ if the detection takes place and $0$ otherwise, i.e., no detection at time $t$ of a concentration larger than $\tau$.
\end{itemize}
In this study the detection time has been considered as the measure of impact in Equation \ref{eq:sp_formulation_generic}.

\section{Distributional representation of sensor placements}

\subsubsection{Sensor Matrices}
\begin{figure}[h]
\centering
    \begin{subfigure}{.49\textwidth}
        \centering
        \includegraphics[width=1\linewidth]{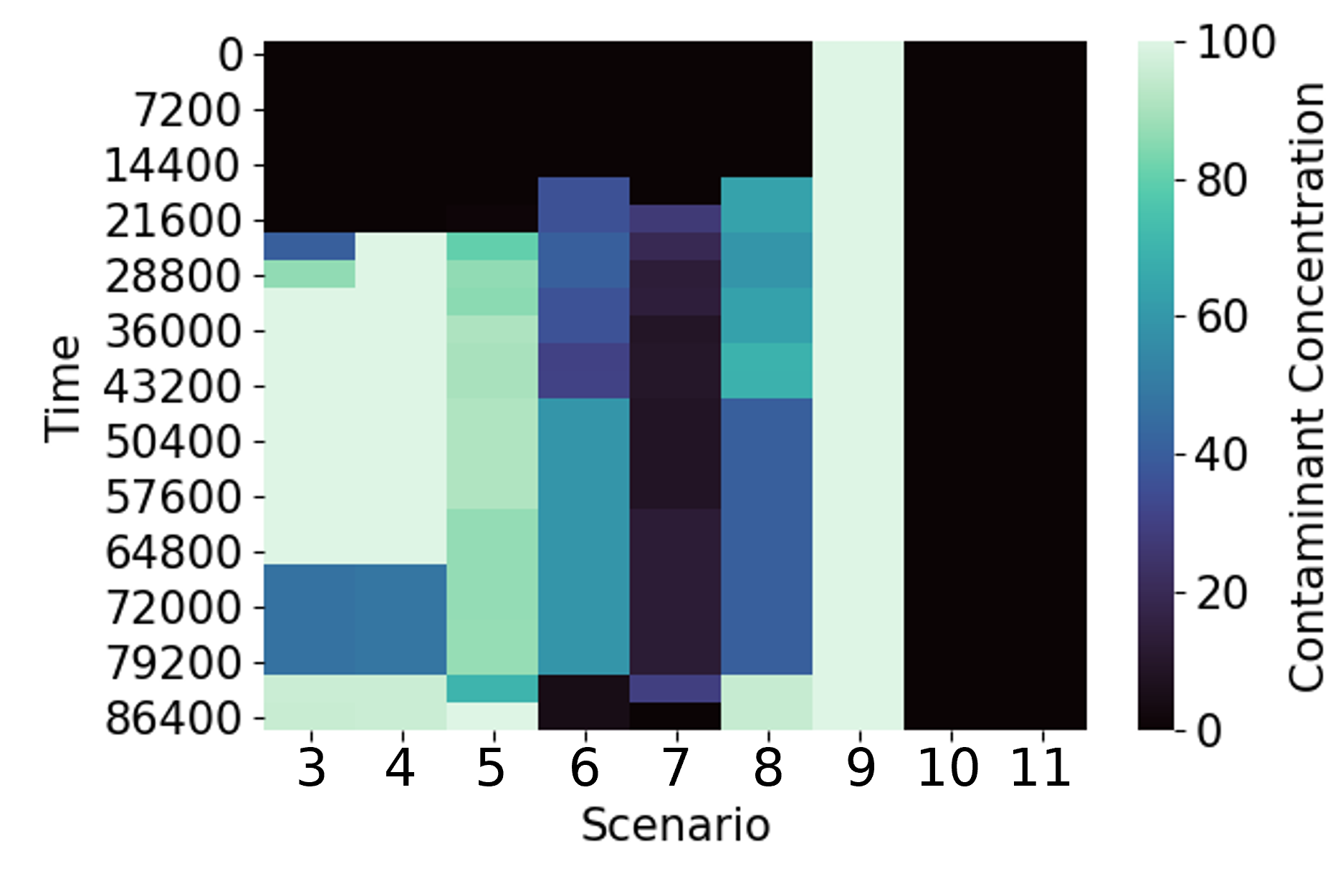}
        \caption{Sensor placed at node $9$.}
    \end{subfigure}
    \begin{subfigure}{.49\textwidth}
        \centering
        \includegraphics[width=1\linewidth]{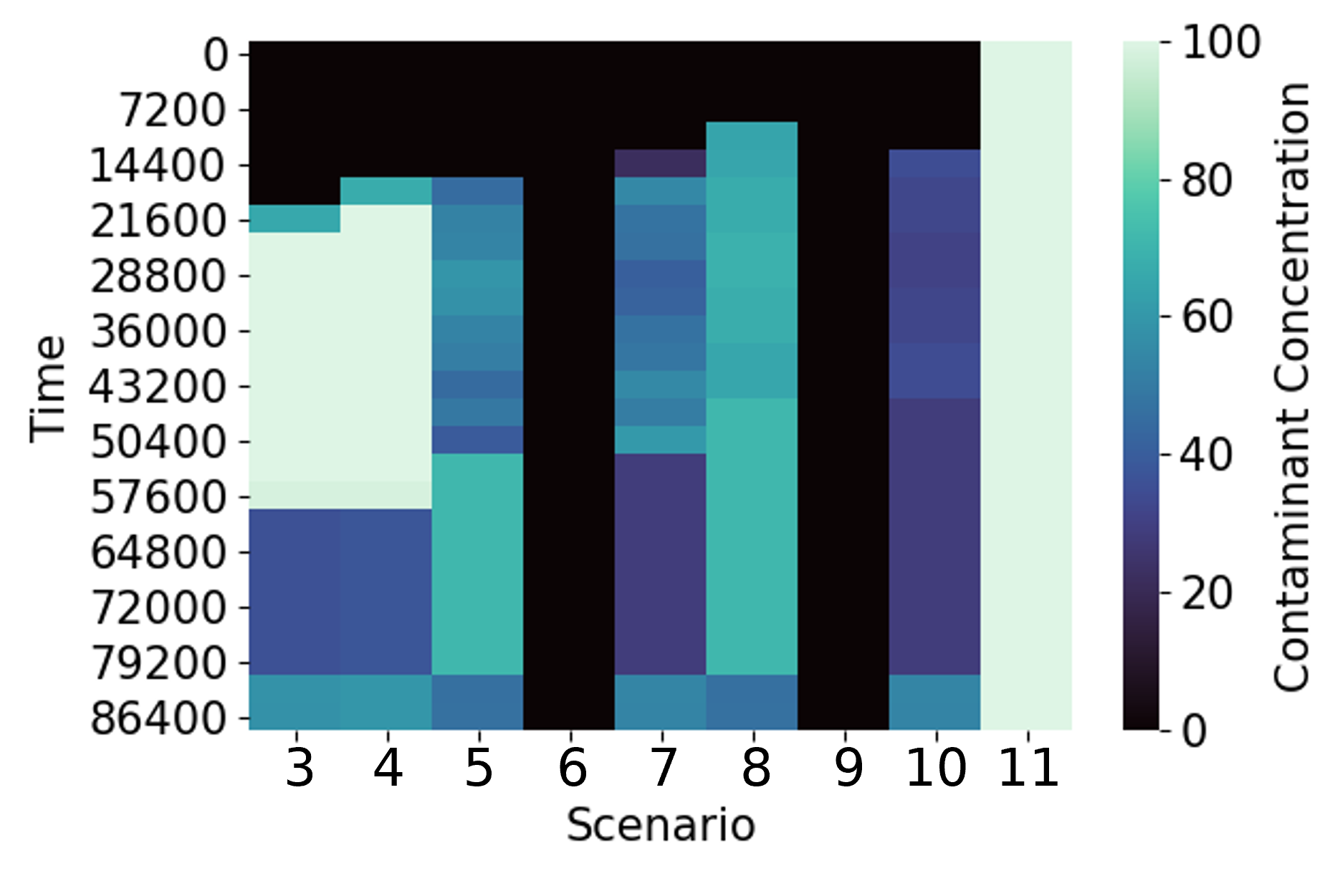}
        \caption{Sensor placed at node $11$.}
    \end{subfigure}
\caption{Examples of sensor matrices considering Net1 WDN.}
\label{fig:ch07_sensor_matrix}
\end{figure}

\begin{figure}[h]
    \centering
    \includegraphics[width=0.5\linewidth]{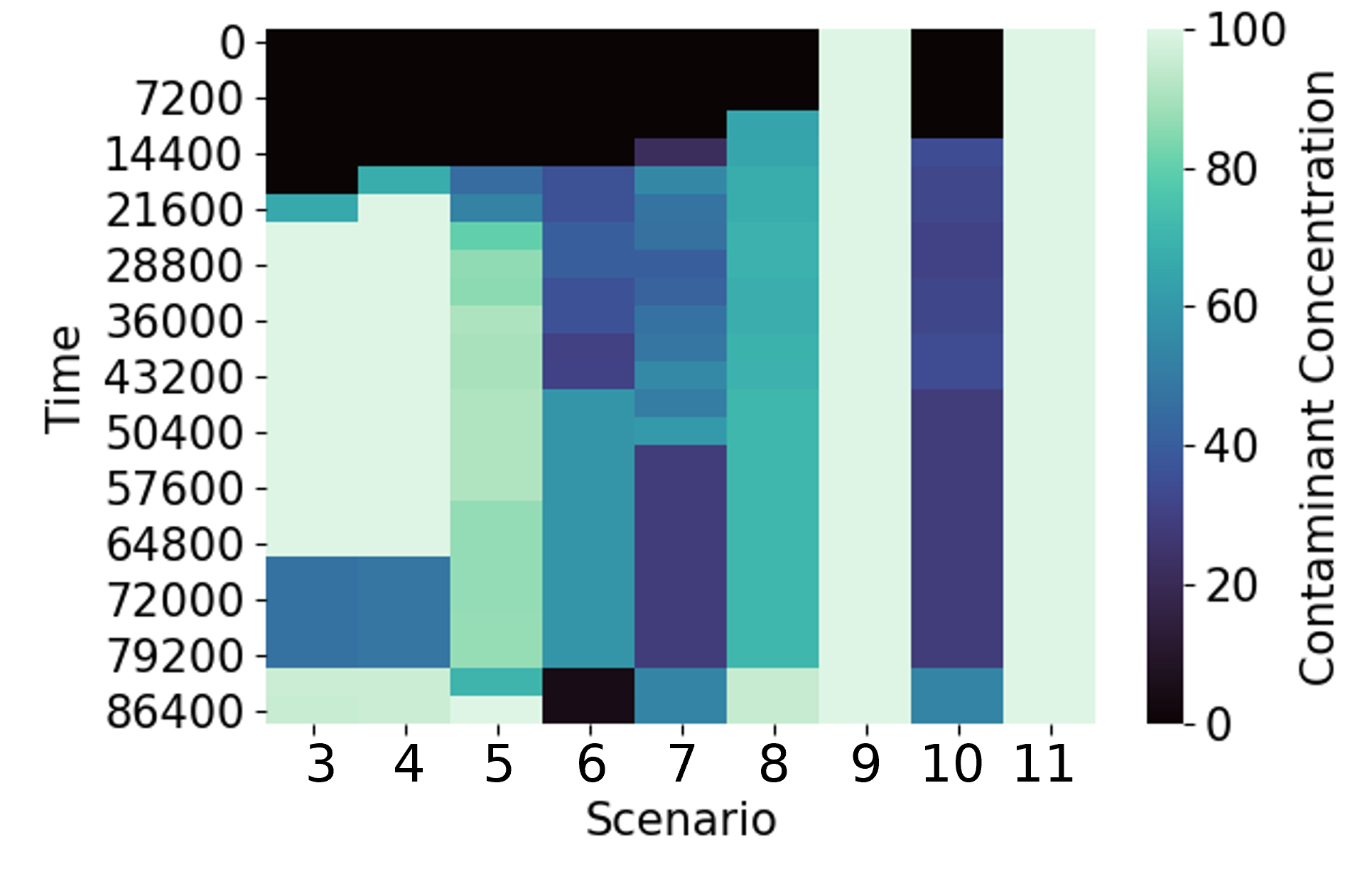}
    \caption{Example of a placement matrix. Two sensors placed at node $9$ and $11$ respectively.}
    \label{fig:ch07_placement_matrix}
\end{figure}

Denote with $S^\ell$ the so-called ``sensor matrix'', where $\ell = 1,\ldots,|L|$ is an index identifying the location where the sensor is deployed at. Each entry $s_{ta}^\ell$ represents the concentration of the contaminant for the event $a \in A$ at the simulation step $t = 0,\ldots,K$, with $T_{max} = K \Delta t$. Without loss of generality, assume that the contaminant is injected at the beginning of the simulation (i.e., $t = 0$). \\
Figure \ref{fig:ch07_sensor_matrix} shows the heatmaps of two sensor matrices of Net1, regarding respectively the nodes $9$ and $11$. In this example, all the nodes are considered as possible sensor locations and all the nodes but 1 (the tank) and 2 (the reservoir) are considered as events (i.e., location in which the contaminant can be injected). \\
 \\
Analogously, a ``sensor placement matrix'', $H^{(s)} \in \mathbb{R}^{(K+1)×|A|}$, is defined (Figure \ref{fig:ch07_placement_matrix}), where every entry $h_{ta}$ represents the maximum concentration over those detected by the sensors in $s$, for the event $a$ and at time step $t$. Suppose to have a sensor placement $s$ consisting of $m$ sensors with associated sensor matrices $S^1,\ldots,S^m$, then $h_{ta} = \max_{j=1,\ldots,m} s_{ta}^j \; \forall a \in A$. \\
There is a relation between $s$ and the associated $H^{(s)}$: more precisely, the columns of $H^{(s)}$ having maximum concentration at row $t = 0$ (i.e., injection time) are those associated to events with injection occurring at the deployment locations of the sensors in $s$. \\
Moreover, $H^{(s)}$ is the basic data structure on which MDT is computed. Unfortunately, the main issue is that it is not possible to work directly with sensor placement matrices. In searching for an optimal vector $s$, $H^{(s)}$ is just an additional observable information before computing the two objectives $f_1 (s)$ and $f_2 (s)$.

\subsubsection{Placement Histogram}
\begin{figure}[h]
    \centering
    \includegraphics[width=0.5\linewidth]{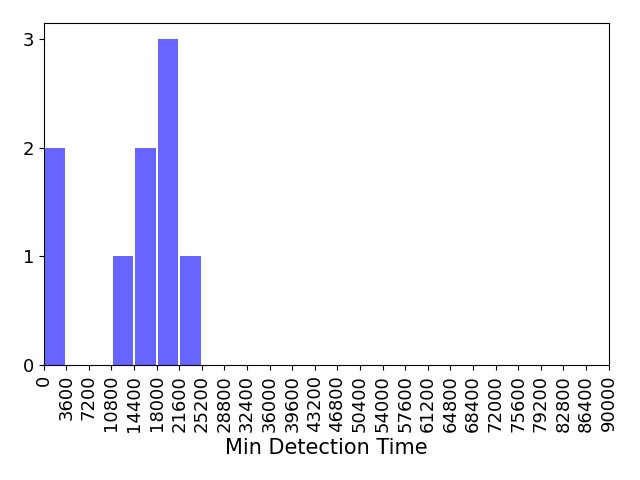}
    \caption{Example of a placement histogram of Net1 WDN. Two sensors placed at node $9$ and $11$ respectively. This histogram corresponds to the placement matrix displayed in Figure \ref{fig:ch07_placement_matrix}}
    \label{fig:ch07_placement_histo}
\end{figure}

The information in $H^{(s)}$ about a placement can be represented as an histogram $h^{(s)}$ (Figure \ref{fig:ch07_placement_histo}). Consider the time steps in the simulation $\Delta t_i = t_i - t_{i-1}$ where $i=1,...,k$ are equidistanced in the simulation time horizon $(0,T_{max})$ with $T_{max} = k \Delta t$, $\Delta t = 1$ and $k = 24$. Consider also the discrete random variable $|A_i|$ where $A_i = \{ a \in A: \hat{t}_a \in \Delta t_i \}$. To each sensor placement $s$ it is possible to associate not only the placement matrix $H^{(s)}$ but also the histogram $h^{(s)}$ whose bins are $\Delta t_i$ and weights are $|A_i|$. In other words, each bin of the histogram $h^{(s)}$ represents the number of events that are detected in a specific time range by $s$: these values can be extracted from the placement matrix $H^{(s)}$; indeed, each column of this matrix represents an event, and the detection time of this event is given by the row in which the contaminant concentration exceed a given threshold $\tau$. \\
 \\
An extra bin has been added ($86400$ to $90000$) whose weight $|A_{k+1}|$ represents for any sensor placement the number of contamination events which were undetected during the simulation (and hence the detection probability). The relation between SP and histograms is many to one: one histogram indeed can be associated to different SP.
Intuitively the larger the probability mass in lower $\Delta t_i$ the better is the sensor placement; the larger the probability mass in the higher $\Delta t$ the worse is sensor placement. The worst SPs are those for which no detection took place in the simulation horizon. An ``ideal'' placement can be defined as the histogram in which $|A^1| = |A|$.

\section{Search, Information and Objective space}
\label{ch07:info_space}
\begin{figure}[h]
    \centering
    \includegraphics[width=1\linewidth]{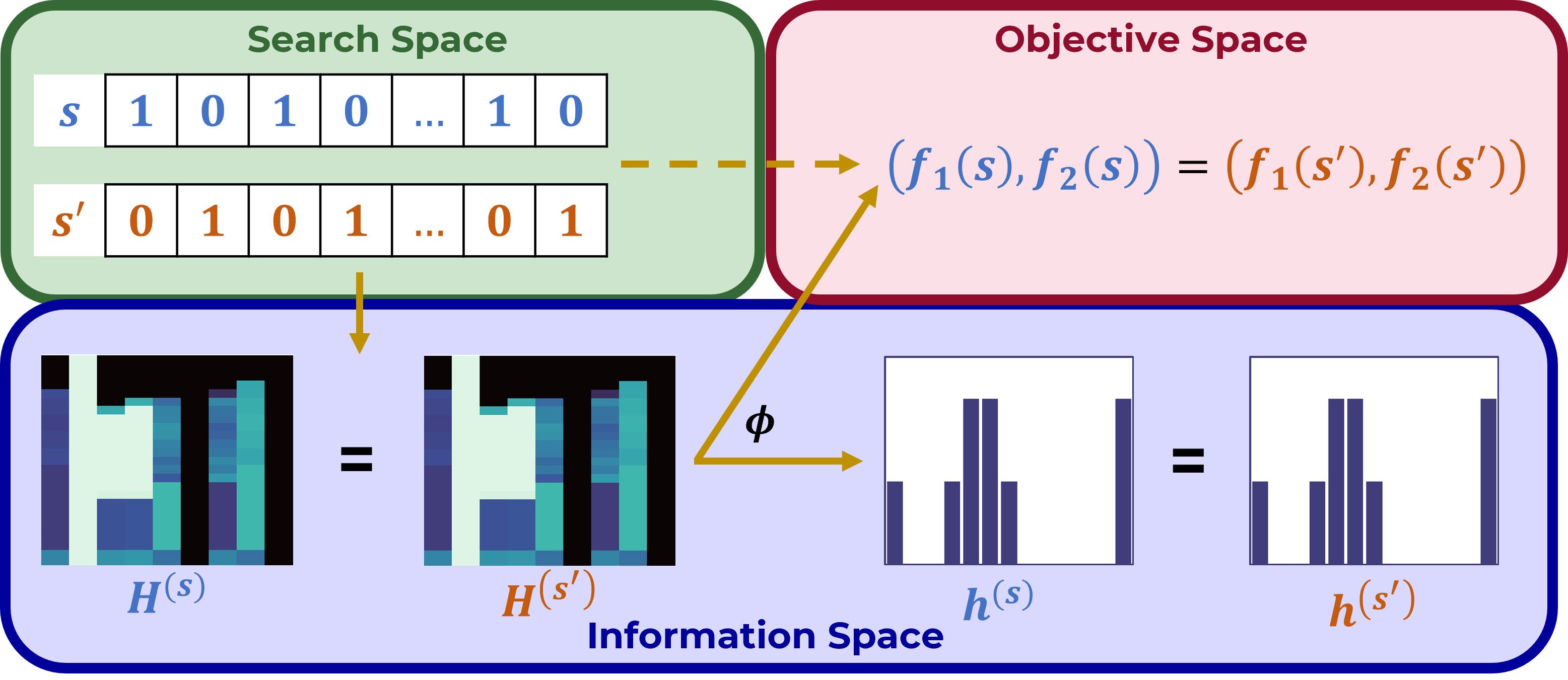}
    \caption{An example of how PI selects the new point.}
    \label{fig:ch07_info_space}
\end{figure}

The search space consists of all the possible SPs, given a set $L$ of possible locations for their deployment, and resulting feasible with respect to the constraints in (Equation \ref{eq:sp_formulation_generic}); formally, the feasible set is $\Omega = \left\{ s \in \{0,1\}^{|L|}: \sum_{i=1}^{|L|} s_i \leq p \right\}$. As already mentioned, the computation of the two objectives $f_1 (s)$ and $f_2 (s)$ requires the trace of the quality simulation. Beyond the computation of $f_1$ and $f_2$ matrix $H^{(s)}$ and histogram $h^{(s)}$ offer a much richer representation. This entire process can be schematized as in Figure \ref{fig:ch07_info_space}. \\
 \\
Then the Wasserstein distance, introduced in Chapter \ref{ch05:wst}, can be used to explore the information space. Any distance in $\Omega$ can be highly misleading, in that two SPs $s$ and $s'$ distant in $\Omega$ might correspond to close values of $H^{(s)}$ and $H^{(s')}$ sensor placement matrix, leading to close points in the objective space. This means that the landscape of the problem may have a huge number of global optima, also significantly distant among them in $\Omega$. Indeed, assume to have $s,s':d(s,s')=d_{max}$ (e.g., if $d(\cdot,\cdot)$ is the Hamming distance, $s=(0,1,0,1,\ldots)$ and $s'=(1,0,1,0,\ldots)$), then it is anyway possible to have $\delta(H^{(s)}, H^{(s')}) \backsimeq 0$, and to observe $(f_1(s),f_2(s)) \backsimeq (f_1(s'), f_2(s'))$ with $\delta(\cdot,\cdot)$ a suitable distance between matrices. In this thesis the landscape is explored through histograms in the information space and their Wasserstein distance.

\section{Computational settings}
Since the compared algorithms are non-deterministic algorithms, it is necessary to perform multiple run of them to obtain statistically robust results. In particular, 30 runs of each algorithm have been performed.
In the two evolutionary algorithms, NSGA-II and MOEA/WST as mutation operator, the Pymoo implementation of \textit{Bitflip Mutation} \cite{DBLP:conf/gecco/DebSO07} is used, with a probability of mutation equal to $0.5$. \\
In MOEA/WST a new problem specific crossover operator has been defined and used, as described in Chapter \ref{ch06:crossover}, while in NSGA-II the Pymoo implementation of the \textit{one-point crossover} \cite{DBLP:conf/gecco/DebSO07} has been used.
In these experiments, a population of $40$ individuals is considered and at each generation an offspring of $10$ new chromosome is generated for a total of $100$ generations in the case of Hanoi and $250$ generations in the case of Neptun. The initial population is sampled randomly. \\
To make the comparison fair with $q$ParEGO, in the experiments, $q=10$ is considered. In this way, in each iteration of BO, a batch of $10$ new points are observed.

\section{Computational results}

\subsubsection{Hanoi}
\begin{figure}[h!]
\centering
    \begin{subfigure}{.49\textwidth}
        \centering
        \includegraphics[width=1\linewidth]{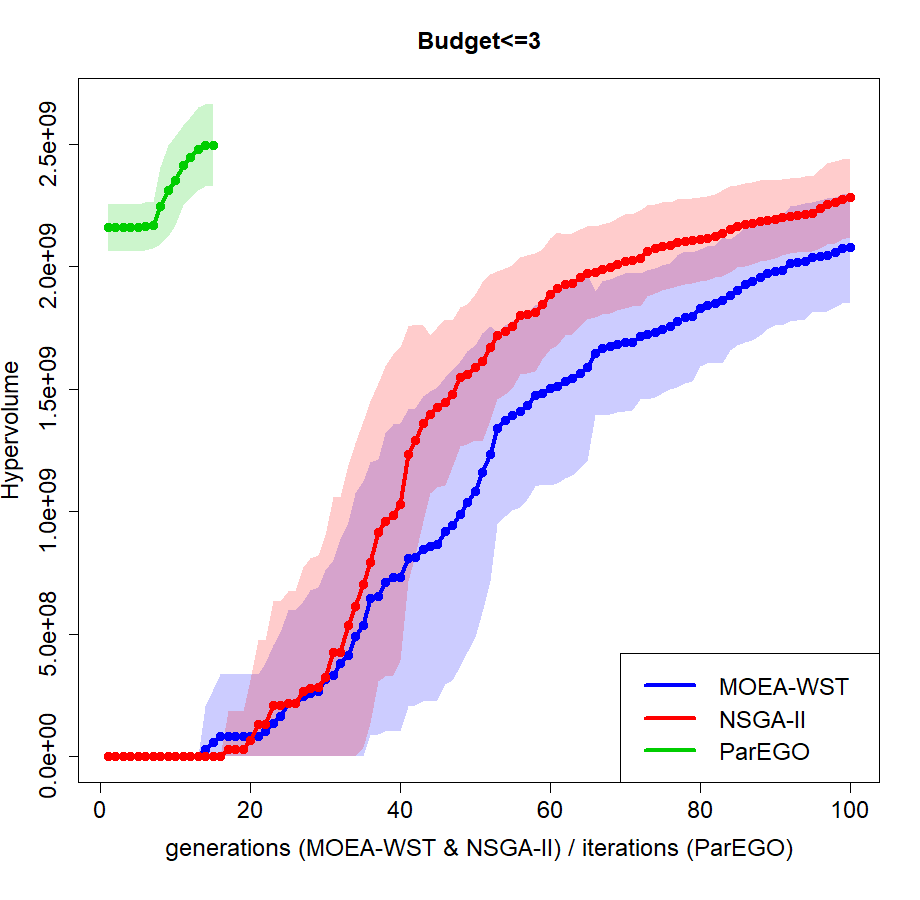}
        \caption{Hypervolume over generation.}
    \end{subfigure}
    \begin{subfigure}{.49\textwidth}
        \centering
        \includegraphics[width=1\linewidth]{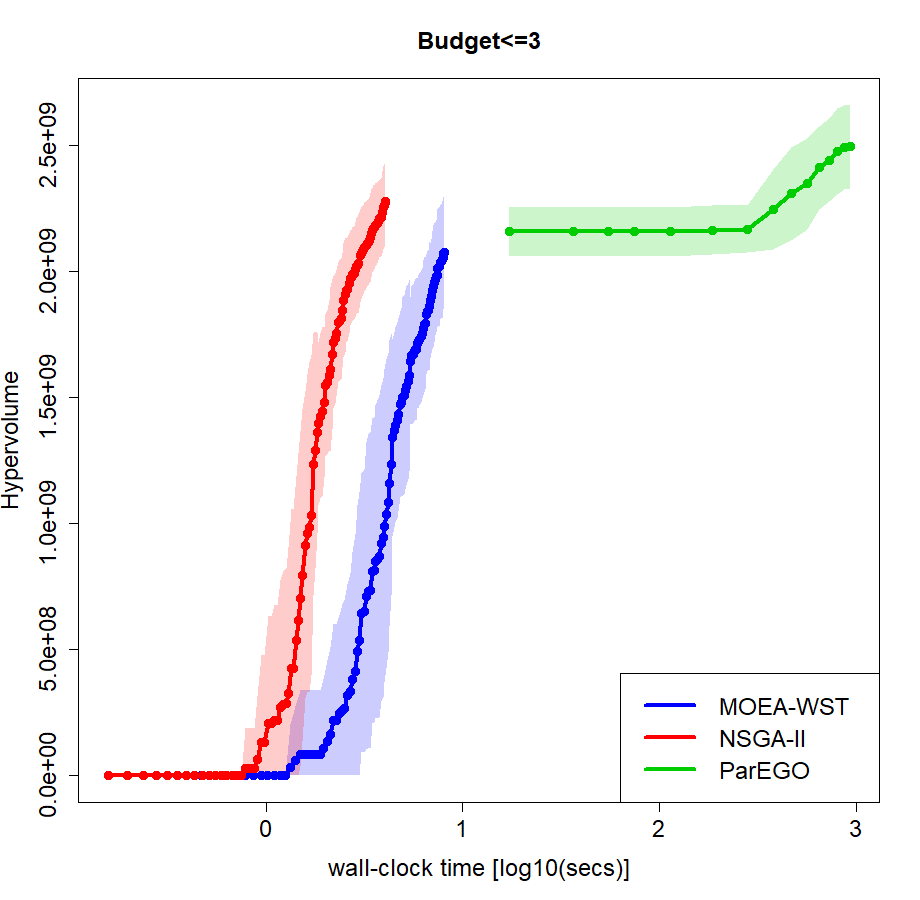}
        \caption{Hypervolume over wall-clock time.}
    \end{subfigure}
\caption{Hypervolume curves of the three algorithms in the case of budget $\leq 3$.}
\label{fig:ch07_hanoi_3}
\end{figure}

\begin{figure}[h!]
\centering
    \begin{subfigure}{.49\textwidth}
        \centering
        \includegraphics[width=1\linewidth]{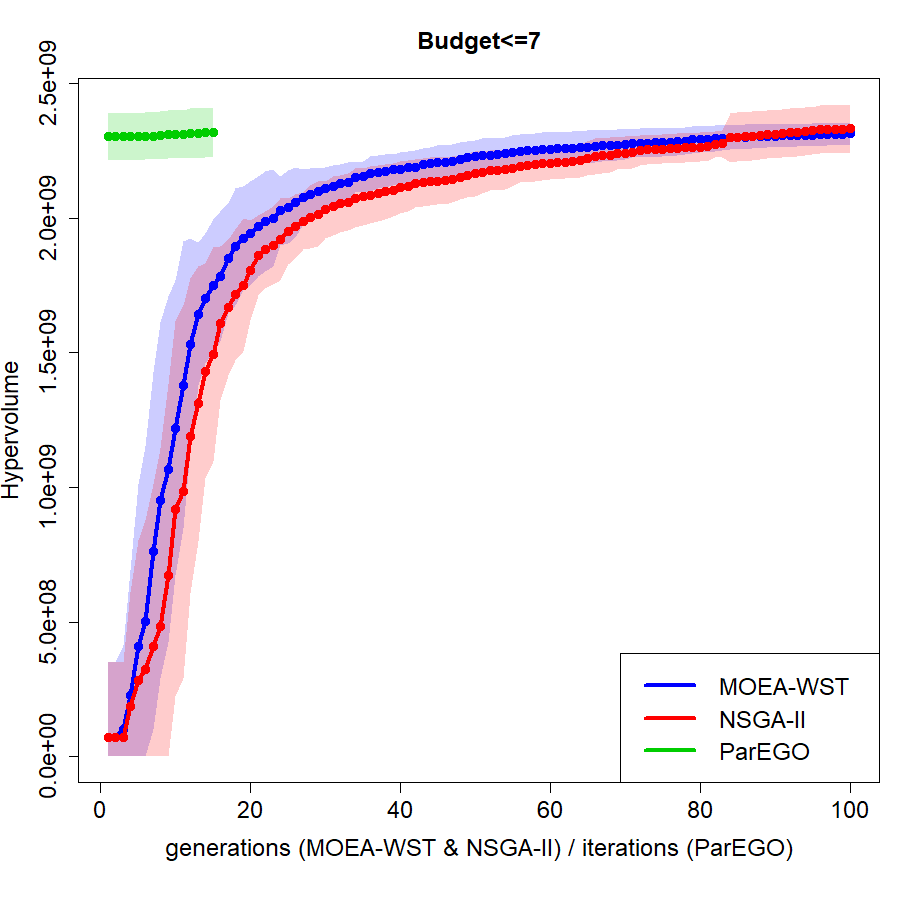}
        \caption{Hypervolume over generation.}
    \end{subfigure}
    \begin{subfigure}{.49\textwidth}
        \centering
        \includegraphics[width=1\linewidth]{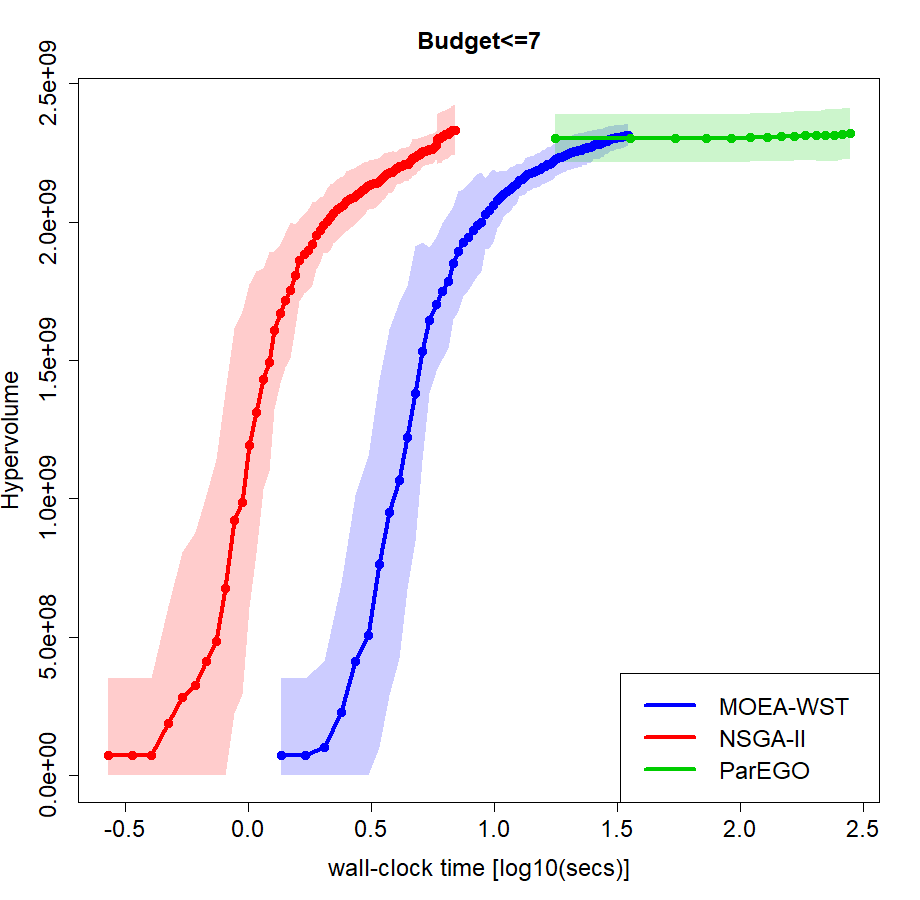}
        \caption{Hypervolume over wall-clock time.}
    \end{subfigure}
\caption{Hypervolume curves of the three algorithms in the case of budget $\leq 7$.}
\label{fig:ch07_hanoi_7}
\end{figure}

\begin{figure}[h]
\centering
    \begin{subfigure}{.49\textwidth}
        \centering
        \includegraphics[width=1\linewidth]{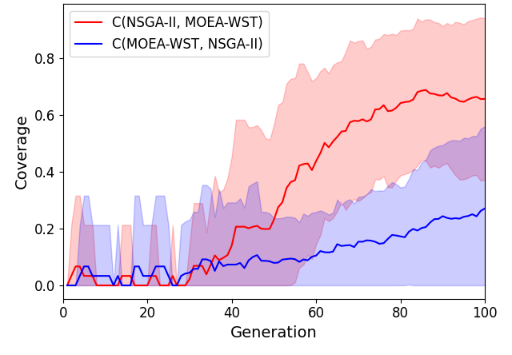}
        \caption{Budget $\leq 3$.}
    \end{subfigure}
    \begin{subfigure}{.49\textwidth}
        \centering
        \includegraphics[width=1\linewidth]{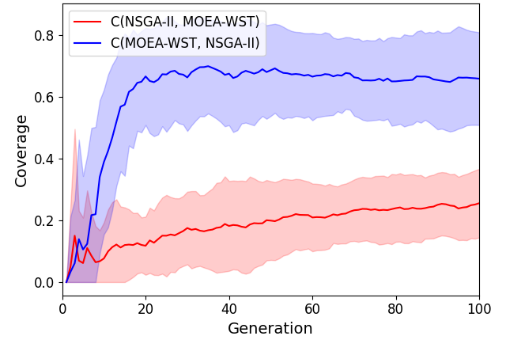}
        \caption{Budget $\leq 7$.}
    \end{subfigure}
\caption{Coverage over generations between the approximate Pareto fronts generated by NSGA-II and MOEA/WST.}
\label{fig:ch07_hanoi_coverage}
\end{figure}

Figures \ref{fig:ch07_hanoi_3} and \ref{fig:ch07_hanoi_7} display the average value and standard deviation of hypervolume ($y$-axis) obtained from experiments for different values of $p$. In each experiment $30$ replications have been performed to generate the estimation sample. Given the significant computation overhead inherent in ParEGO two different units in the $x$-axis have been used: number of generations for MOEA/WST and NSGA-II and iterations for ParEGO on the left, while on the rights the units are wall-clock time. \\
In terms of hypervolume MOEA/WST and NSGA-II offer a balanced performance. In terms of wall-clock time NSGA-II has an advantage due to the computations of the Wasserstein distances required by MOEA/WST. \\
ParEGO tells a different story: the sample efficiency of Bayesian optimization is transferred to ParEGO which reaches much earlier than other methods high values of hypervolume. It is fair to say that the advantage of ParEGO is offset by a significant computational overhead due to the updating of the mean and variance of the Gaussian process and in particular the inversion of the covariance matrix. Even if the BoTorch implementation of ParEGO is highly optimized, the impact of this overhead is clearly shown by the wall-clock time displayed in Figure \ref{fig:ch07_hanoi_3} and \ref{fig:ch07_hanoi_7}. \\
 \\
The difference in values of hypervolume between MOEA/WST, NSGA-II and $q$ParEGO has been tested for statistical significance for different values of $p$ and different generations/iterations counts. A Wilcoxon test for MOEA/WST and NSGA-II for the samples in generations 25/50/100 (each new generation requires 10 function evaluations) is used (Table \ref{tab:ch07_wilcoxon}). The null hypothesis (H0) is that the samples are from the same distribution. \\
\begin{table}[h!]
\caption{Comparing hypervolume of MOEA/WST against those of the other two approaches (values are $\times 10^9$) and with respect to different budgets $p$ and number of generations. Statistical significance has been investigated through a Wilcoxon test ($p$-value is reported).}
\begin{adjustbox}{max width=\textwidth}
\begin{tabular}{ccccccc}
\hline
$\pmb{p}$        & \textbf{Generations} & \textbf{MOEA/WST}                                          & \textbf{NSGA-II}                                           & \textbf{ParEGO}                                                             & \textbf{\begin{tabular}[c]{@{}c@{}}MOEA/WST \\ vs NSGA-II\\ p-value\end{tabular}} & \textbf{\begin{tabular}[c]{@{}c@{}}MOEA/WST \\ vs ParEGO\\ p-value\end{tabular}} \\ \hline \hline
\multirow{4}{*}{3}  & 25                   & \begin{tabular}[c]{@{}c@{}}0.2188 \\ (0.3790)\end{tabular} & \begin{tabular}[c]{@{}c@{}}0.2190 \\ (0.4538)\end{tabular} & \multirow{4}{*}{\begin{tabular}[c]{@{}c@{}}0.2496 \\ (0.1641)\end{tabular}} & 0.811                                                                             & \textless{}0.001                                                                 \\
                    & 50                   & \begin{tabular}[c]{@{}c@{}}1.0828 \\ (0.5942)\end{tabular} & \begin{tabular}[c]{@{}c@{}}\textbf{1.5883} \\ \textbf{(0.2969)}\end{tabular} &                                                                             & \textless{}0.001                                                                  & \textless{}0.001                                                                 \\
                    & 100                  & \begin{tabular}[c]{@{}c@{}}2.0769 \\ (0.2225)\end{tabular} & \begin{tabular}[c]{@{}c@{}}\textbf{2.2808} \\ \textbf{(0.1601)}\end{tabular} &                                                                             & \textless{}0.001                                                                  & \textless{}0.001                                                                 \\ \hline
\multirow{4}{*}{7}  & 25                   & \begin{tabular}[c]{@{}c@{}}\textbf{2.0421} \\ \textbf{(0.135)}\end{tabular}  & \begin{tabular}[c]{@{}c@{}}1.9517 \\ (0.1217)\end{tabular} & \multirow{4}{*}{\begin{tabular}[c]{@{}c@{}}2.3189 \\ (0.0911)\end{tabular}} & \textless{}0.001                                                                  & \textless{}0.001                                                                 \\
                    & 50                   & \begin{tabular}[c]{@{}c@{}}2.2290 \\ (0.0608)\end{tabular} & \begin{tabular}[c]{@{}c@{}}2.1649 \\ (0.0825)\end{tabular} &                                                                             & 0.001                                                                             & \textless{}0.001                                                                 \\
                    & 100                  & \begin{tabular}[c]{@{}c@{}}2.3145\\ (0.0414)\end{tabular}  & \begin{tabular}[c]{@{}c@{}}2.3328 \\ (0.0891)\end{tabular} &                                                                             & 0.686                                                                             & 0.820                                                                            \\ \hline
\end{tabular}
\end{adjustbox}
\label{tab:ch07_wilcoxon}
\end{table} \\ \\
Figure \ref{fig:ch07_hanoi_coverage} displays the curve of coverage as the function of number of generations. The figures show that MOEA/WST improves comparatively its coverage as the number $p$ increases. The advantage of MOEA/WST over NSGA-II is given more significant in terms of coverage. \\
 \\
The complete results considering budget 3, 7, 9, 15 and 20 are reported in Appendix \ref{apx:hanoi}. To add further elements of assessment of the comparative performance of the methods in more challenging conditions the algorithms have been tested on the real-life problem of Neptun.

\subsubsection{Neptun}
\begin{figure}[h]
\centering
    \begin{subfigure}{.49\textwidth}
        \centering
        \includegraphics[width=1\linewidth]{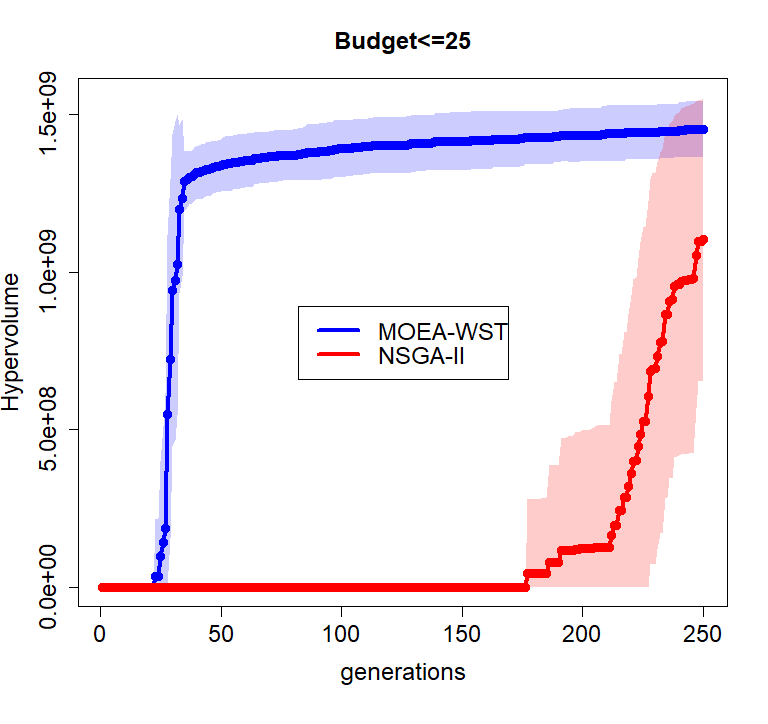}
        \caption{Hypervolume over generations.}
    \end{subfigure}
    \begin{subfigure}{.49\textwidth}
        \centering
        \includegraphics[width=1\linewidth]{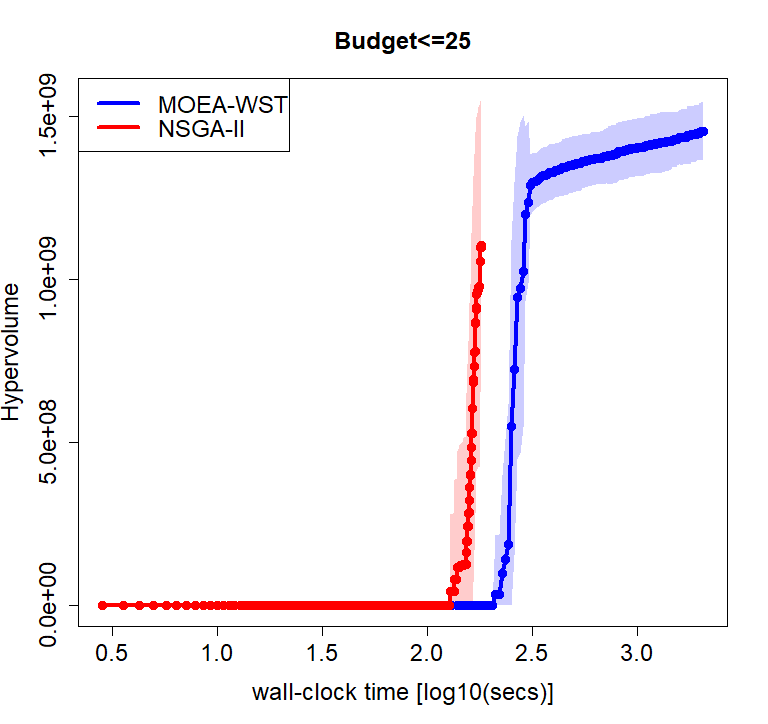}
        \caption{Hypervolume over wall-clock time.}
    \end{subfigure}
\caption{Hypervolume curves of the two evolutionary algorithms in the case of budget $\leq 25$.}
\label{fig:ch07_neptun_hv}
\end{figure}

In Figure \ref{fig:ch07_neptun_hv} are reported the results of the Neptun WDN only for the two evolutionary algorithms, NSGA-II and MOEA/WST. ParEGO has been excluded from this experiments due to its computational overhead. \\
 \\
The Wilcoxon test is used also in the case of Neptun, for MOEA/WST and NSGA-II for the samples in generations 50, 100, 150, 200 and 250 (Table \ref{tab:ch07_wilcoxon_neptun}). The null hypothesis (H0) is that the samples are from the same distribution.\\
\begin{table}[h]
\centering
\caption{Comparing hypervolume of MOEA/WST against the NSGA-II’s (values are $\times 10^9$). Statistical significance has been investigated through a Wilcoxon test ($p$-value is reported).}
\begin{tabular}{cccc}
\hline
\textbf{Generations} & \textbf{MOEA/WST} & \textbf{NSGA-II} & \textbf{\begin{tabular}[c]{@{}c@{}}MOEA/WST vs NSGA-II \\ p-value\end{tabular}} \\ \hline \hline
50                   & \textbf{1.3377 (0.0826)}   & 0.0000 (0.0000)  & \textless{}0.001                                                                \\
100                  & \textbf{1.3916 (0.0900)}   & 0.0000 (0.0000)  & \textless{}0.001                                                                \\
150                  & \textbf{1.4150 (0.0915)}   & 0.0000 (0.0000)  & \textless{}0.001                                                                \\
200                  & \textbf{1.4350 (0.0848)}   & 1.2232 (0.0374)  & \textless{}0.001                                                                \\
250                  & \textbf{1.4530 (0.0880)}   & 1.1042 (0.0448)  & \textless{}0.001                                                                \\ \hline
\end{tabular}
\label{tab:ch07_wilcoxon_neptun}
\end{table} \\
In Neptun the comparative performance of MOEA/WST is quite impressive in terms of hypervolume. In terms of wall clock NSGA-II has a small advantage and then it stops at a lower value after having performed its assigned 250 generations. \\
As for ParEGO, it is well known that the sample efficiency of Bayesian optimization methods degrades as the dimension of the search space increase. Indeed, the wall clock times of ParEGO are about four times the values of MOEA/WST. It is also worth remarking that ParEGO and NSGA-II come from consolidated software frameworks, while MOEA/WST is still highly experimental.

%% file: chapters/chapter08.tex
\chapter{Recommender Systems}
\label{ch08:recommender}
This chapter focus on the problem of recommend a given number of items to each user based on the rated items. A key element of this chapter is given by the mapping of each user and of the objectives function in a space of discrete distributions which is explored through the Wasserstein distance. This enables the usage of the evolutionary algorithm presented in Chapter \ref{ch06:moeawst}, namely MOEA/WST.

\section{The problem definition}
In the most general framework, a Collaborative Filtering (CF) problem is based on the definition of two sets \cite{DBLP:journals/jmlr/TakacsPNT09}:
\begin{itemize}
    \item The set of users $U=\{u_1,u_2,\ldots,u_M\}$, where $M$ is the number of users.
    \item The set of items $O=\{o_1,o_2,\ldots,o_N\}$, where $N$ is the number of items.
\end{itemize}
Each user expresses its judgement, or rating, $r \in X$, where typical rating values can be binary or integers from a given range. The set of all the ratings given by the users on the items can be represented as a partially specified matrix $R \in \mathbb{R}^{M×N}$, namely rating matrix (Table \ref{tab:ch08_rating_matrix}), where its entries $r_{uo}$ express the possible ratings of user $u$ for item $o$. Usually, each user rates only a small number of items, thus the matrix elements are known in a small number of positions. \\
\begin{table}[h]
    \centering
    \caption{An example of a rating matrix.}
    \begin{adjustbox}{width=.36\linewidth}
        \begin{tabular}{|c||cccc|}
            \hline
            \textbf{R} & $o_1$ & $o_2$ & $\ldots$ & $o_N$ \\
            \hline \hline
            $u_1$ & ? & 2 & $\ldots$ & 5 \\
            $u_2$ & 4 & 3 & $\ldots$ & ? \\
            $\ldots$ & $\ldots$ & $\ldots$ & $\ldots$ & $\ldots$ \\
            $u_M$ & 1 & ? & $\ldots$ & 4 \\
            \hline
        \end{tabular}
    \end{adjustbox}
    \label{tab:ch08_rating_matrix}
\end{table} \\
Once the rating matrix is filled, by any of the methods outlined in Chapter \ref{ch04:instances}, the objective targeted in this thesis is to recommend a number $L$ of items to each user (Table \ref{tab:ch07_top_recc}). A recommendation list $S$ of length $L$ for user $u_i$ is denoted as $S_L (u_i)$. \\
\begin{table}[h]
    \centering
    \caption{An example of a top-$L$ recommendation matrix.}
    \begin{adjustbox}{width=.5\linewidth}
        \begin{tabular}{|c||cccc|}
            \hline
            $S_L$ & Item 1 & Item 1 & $\ldots$ & Item $L$ \\
            \hline \hline
            $u_1$ & $o_{10}$ & $o_{36}$ & $\ldots$ & $o_2$ \\
            $u_2$ & $o_{15}$ & $o_{51}$ & $\ldots$ & $o_{28}$ \\
            $\ldots$ & $\ldots$ & $\ldots$ & $\ldots$ & $\ldots$ \\
            $u_M$ & $o_{39}$ & $o_1$ & $\ldots$ & $o_{15}$ \\
            \hline
        \end{tabular}
    \end{adjustbox}
    \label{tab:ch07_top_recc}
\end{table} \\
In the next sections the MovieLens dataset \cite{DBLP:journals/tiis/HarperK16} has been used. MovieLens is the best known repository of ratings for movies. It is also a benchmark in RS research. Datasets of different sizes are provided. In the following experiments the version containing 1000 users, 1700 items and 100k ratings has been used.

\section{Representing user and items as a graph}
\begin{figure}[h]
    \centering
    \includegraphics[width=0.6\linewidth]{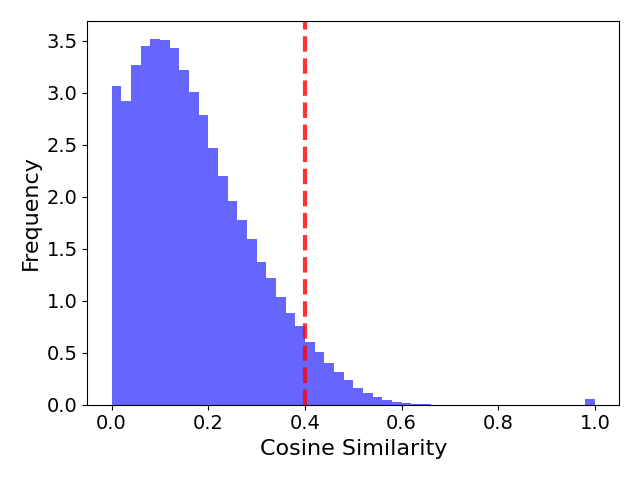}
    \caption{Distribution of cosine similarity between all pairs of users. The dotted red line is the threshold $\tau$.}
    \label{fig:ch08_cos_distr}
\end{figure}

\begin{figure}[h]
    \centering
    \includegraphics[width=0.6\linewidth]{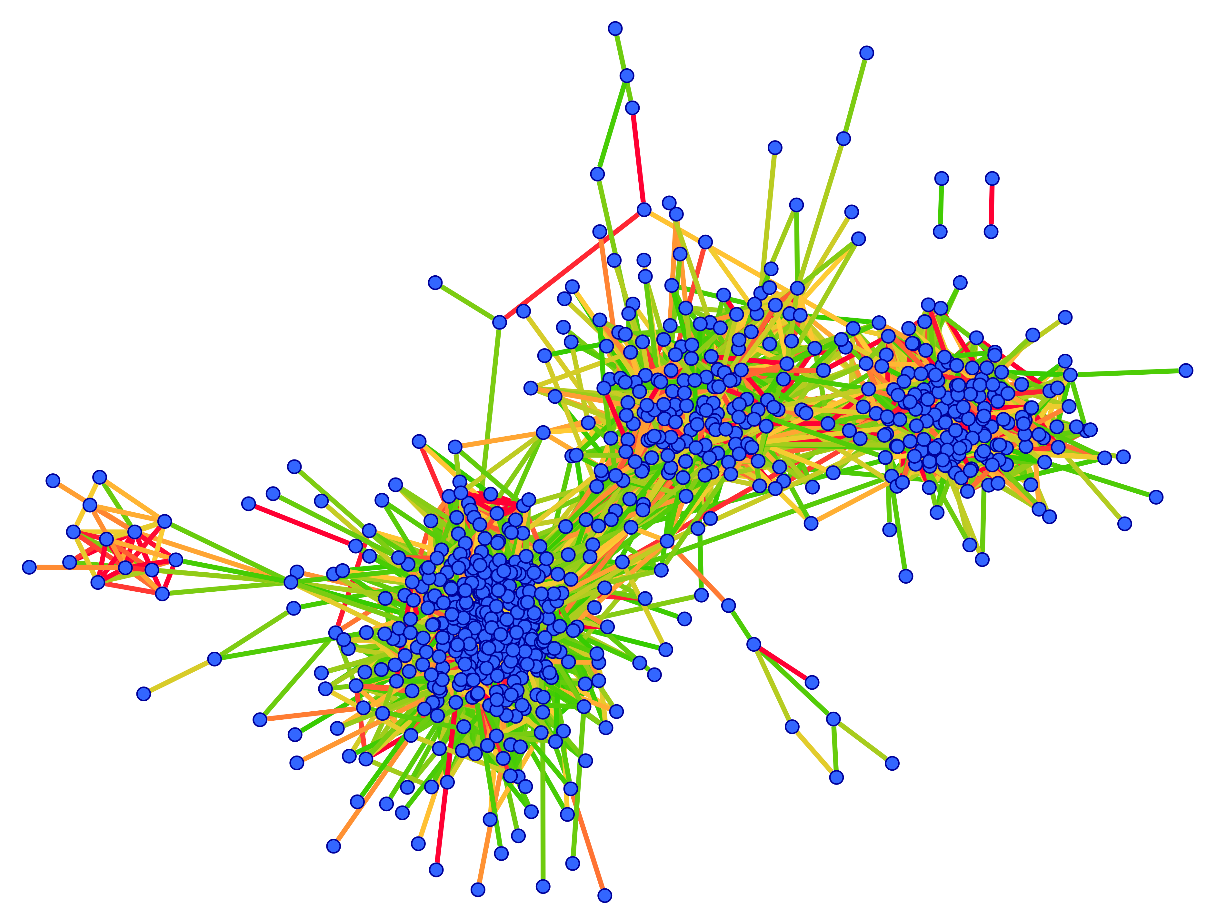}
    \caption{The cosine graph $G_c$.}
    \label{fig:ch08_cos_graph}
\end{figure}

\begin{figure}[h]
\centering
    \begin{subfigure}{.32\textwidth}
        \centering
        \includegraphics[width=1\linewidth]{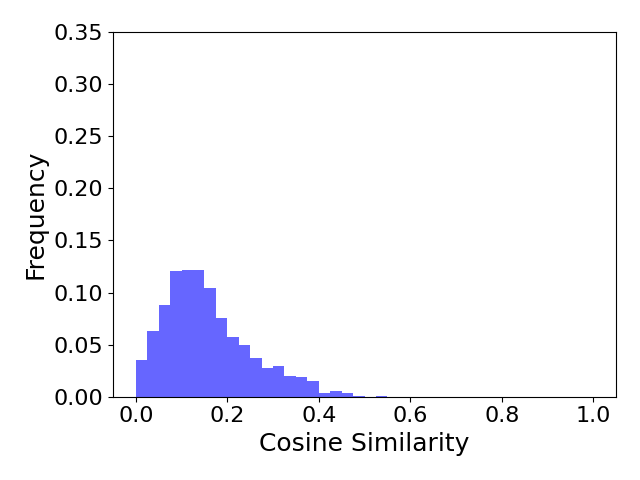}
        \caption{User $16$. Lilac cluster in the Wasserstein graph (Figure \ref{fig:ch08_cluster_wst_2}).}
    \end{subfigure}
    \begin{subfigure}{.32\textwidth}
        \centering
        \includegraphics[width=1\linewidth]{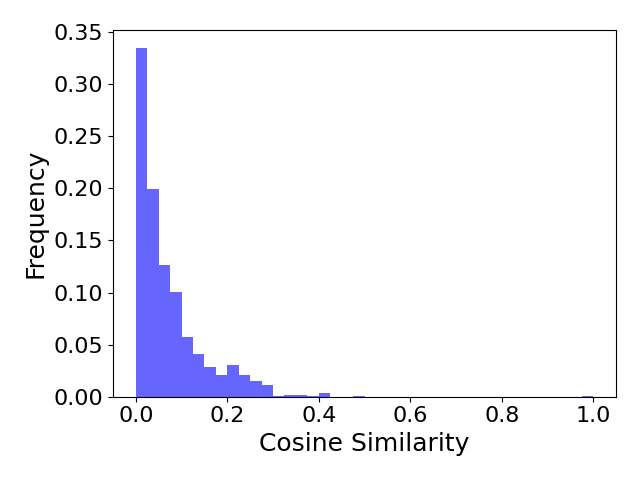}
        \caption{User $33$. Blue cluster in the Wasserstein graph (Figure \ref{fig:ch08_cluster_wst_2}).}
    \end{subfigure}
    \begin{subfigure}{.32\textwidth}
        \centering
        \includegraphics[width=1\linewidth]{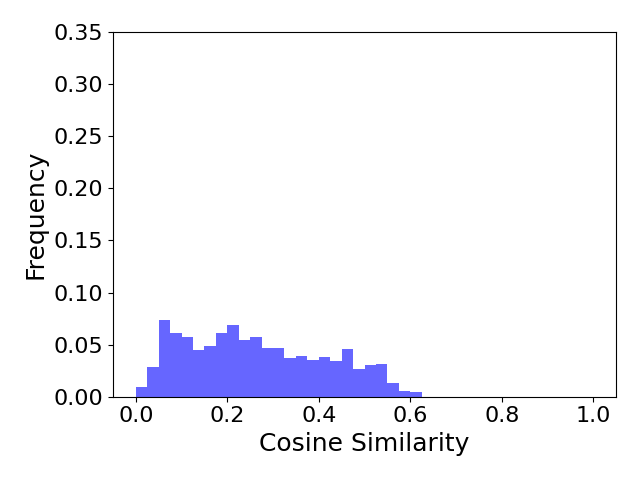}
        \caption{User $300$. Fuchsia cluster in the Wasserstein graph (Figure \ref{fig:ch08_cluster_wst_2}).}
    \end{subfigure}
\caption{Examples of users' cosine similarity distributions.}
\label{fig:ch08_users_distr}
\end{figure}

\begin{figure}[h]
    \centering
    \includegraphics[width=0.6\linewidth]{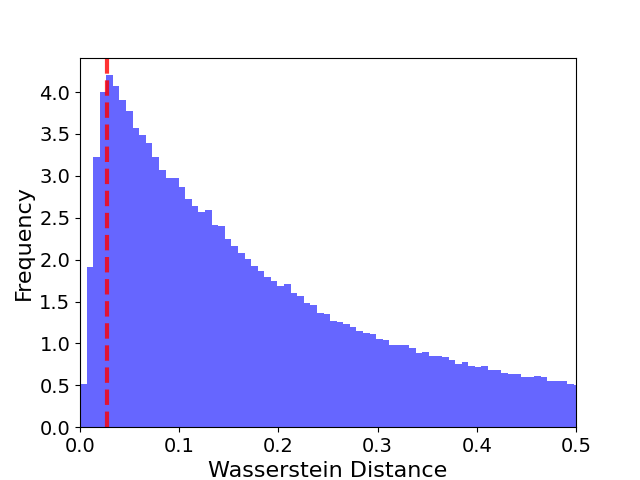}
    \caption{Distribution of Wasserstein distance between all pairs of users represented as histograms. The dotted red line is the threshold $\tau$.}
    \label{fig:ch08_wst_distr}
\end{figure}

\begin{figure}[h]
    \centering
    \includegraphics[width=1\linewidth]{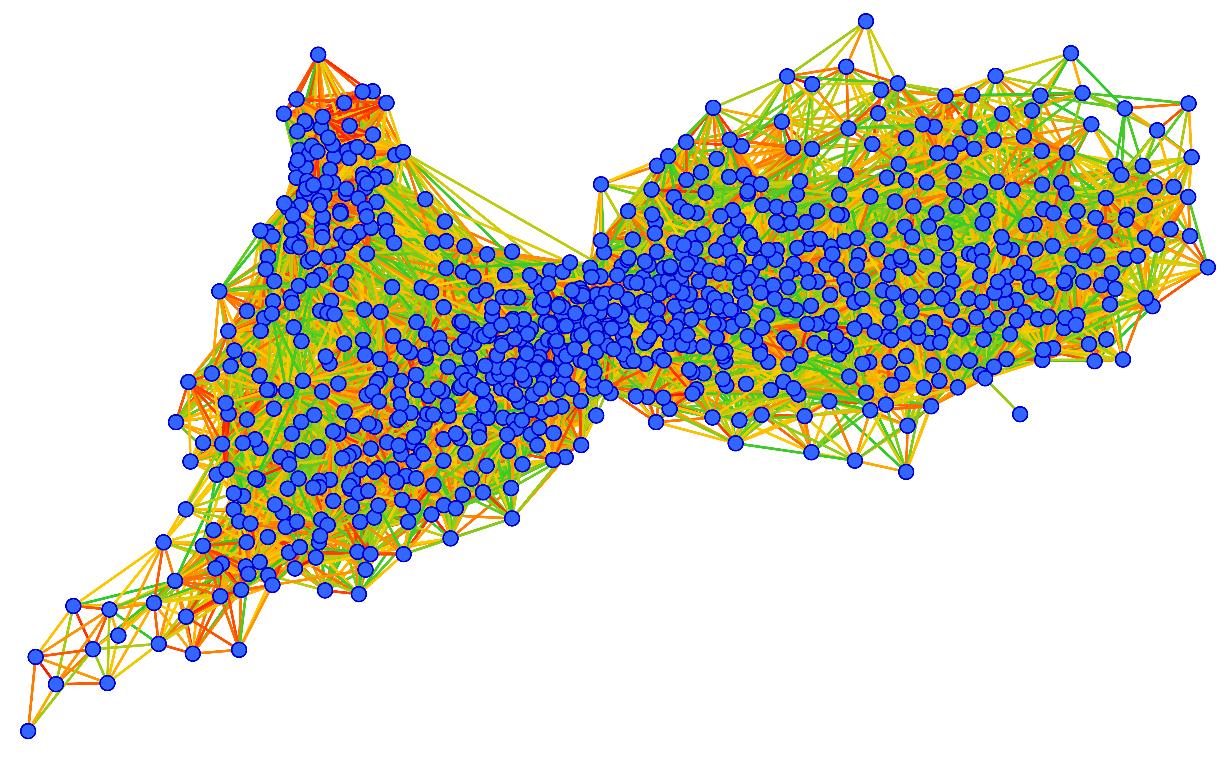}
    \caption{The Wasserstein graph $G_w$.}
    \label{fig:ch08_wst_graph}
\end{figure}

\begin{figure}[h]
\centering
    \begin{subfigure}{.43\textwidth}
        \centering
        \includegraphics[width=1\linewidth]{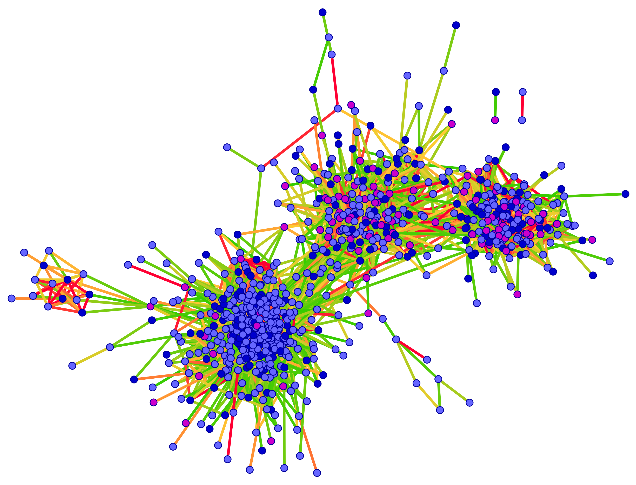}
        \caption{Clusters mapped on the cosine graph.}
    \end{subfigure}
    \begin{subfigure}{.55\textwidth}
        \centering
        \includegraphics[width=1\linewidth]{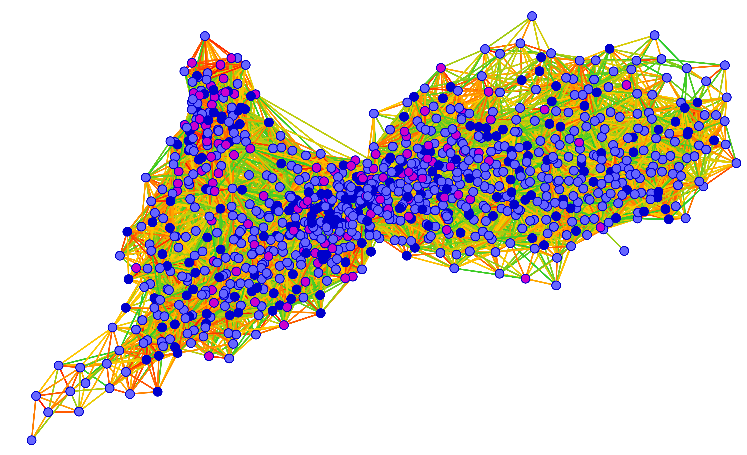}
        \caption{Clusters mapped on the Wasserstein graph.}
    \end{subfigure}
\caption{Results of the spectral clustering on the rating matrix. Three clusters have been identified (blue, lilac and fuchsia).}
\label{fig:ch08_cluster_cos}
\end{figure}

\begin{figure}[h]
\centering
    \begin{subfigure}{.43\textwidth}
        \centering
        \includegraphics[width=1\linewidth]{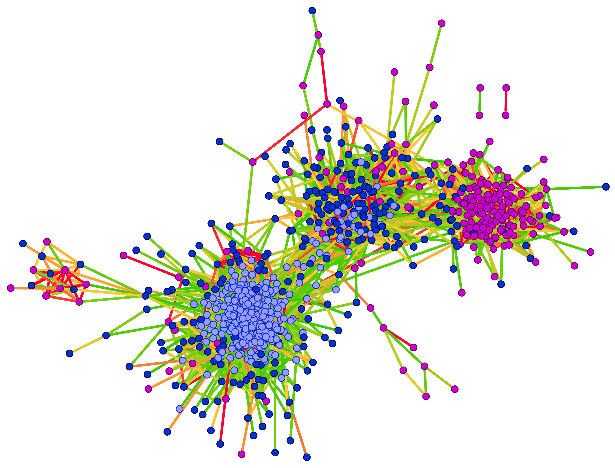}
        \caption{Clusters mapped on the cosine graph.}
        \label{fig:ch08_cluster_wst_1}
    \end{subfigure}
    \begin{subfigure}{.55\textwidth}
        \centering
        \includegraphics[width=1\linewidth]{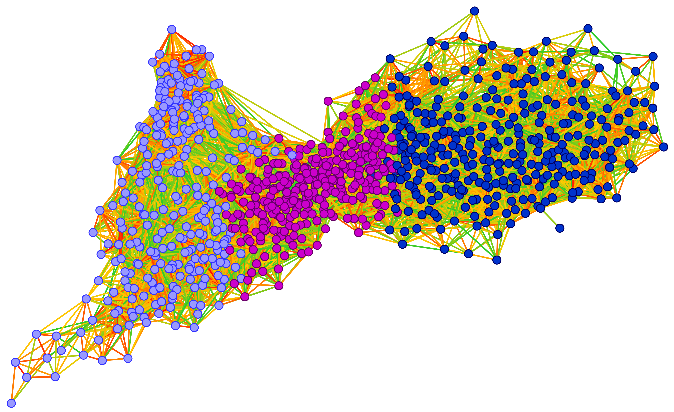}
        \caption{Clusters mapped on the Wasserstein graph.}
        \label{fig:ch08_cluster_wst_2}
    \end{subfigure}
\caption{Results of the spectral clustering on the Wasserstein graph. Three clusters have been identified (blue, lilac and fuchsia).}
\label{fig:ch08_cluster_wst}
\end{figure}

Consider now the rating matrix $R$ as previously defined. The basic idea of Collaborative Filtering methods is that the observed ratings are highly correlated between users and items. If two users have similar taste, it is very likely that the ratings given to the same object by these two users are similar. To compute the similarity between users or items different metrics exist, for instance the Cosine similarity ($\cos$) and the Pearson correlation ($\rho$). Let $R^{(1)}$ and $R^{(2)}$ be two rows (columns) of the rating matrix, then this two similarities measures are defined as follows (Equations \ref{eq:ch08_cos} and \ref{eq:ch08_pearson}):
\begin{gather}
    \cos \left( R^{(1)}, R^{(2)} \right) = \frac{\sum_{i=1}^{N} R^{(1)}_i R^{(2)}_i}{\sqrt{\sum_{i=1}^{N} \left(R^{(1)}_i\right)^2} \sqrt{\sum_{i=1}^{N} \left(R^{(2)}_i\right)^2}} \label{eq:ch08_cos}\\
    \rho \left( R^{(1)}, R^{(2)} \right) = \frac{\sum_{i=1}^{N} \left( R^{(1)}_i - \overline{R^{(1)}} \right)\left( R^{(2)}_i - \overline{R^{(2)}} \right)}{\sqrt{\sum_{i=1}^{N} \left( R^{(1)}_i - \overline{R^{(1)}} \right)^2}\sqrt{\sum_{i=1}^{N} \left( R^{(1)}_i - \overline{R^{(1)}} \right)^2}} \label{eq:ch08_pearson}
\end{gather}
where $\overline{R^{(1)}}$ and $\overline{R^{(2)}}$ are the sample mean respectively of $R^{(1)}$ and $R^{(2)}$. It is important to note that the cosine similarity, in the case of positive values (as the rating matrix) is in the interval $[0,1]$, while the Pearson correlation assumes values in $[-1,1]$. Figure \ref{fig:ch08_cos_distr} shows the distribution of cosine similarity between all pairs of users. \\
 \\
The cosine similarity between individual users can be used to build a graph $G_c=(V_c,E_c)$, in which two nodes (users) are linked together if their similarity is above a given threshold (the dotted red line in Figure \ref{fig:ch08_cos_distr}); then $V_c = \{u_i\}_{i=1,\ldots,M}$ is the set of users and $E_c = \left\{(i,j): \cos \left( u_i,u_j \right) > \tau \right\}$ is the set of edges. Each edge of this graph is then weighted based on the similarity $\cos(u_i, u_j)$. Figure \ref{fig:ch08_cos_graph} displays the resulting graph, in which the edge color represents its weight. The nodes connected by a red edge are the most similar, while the ones connected by a green edge are the most different. \\
 \\
Another representation of the rating matrix is through the association to each user $u_i$ of a one-dimensional histogram $h(u_i)$: the bins are the equi-subdivisions of the interval $[0,1]$ for cosine similarity (and $[-1,1]$ for Pearson correlation) and the weights are the fraction of users whose cosine similarity falls in each bin; the same representation can be item driven. In Figure \ref{fig:ch08_users_distr} cosine similarity is used with a bin length of $0.025$. \\
 \\
According to this representation each user is described by a signature, feature vector, given by the bins and the associated weights. In this feature space the elements are probabilistic distributions. Many models can be used to compute the distance between distributions, as analyzed in Chapter \ref{ch05:wst}. In this thesis, for the motivations expressed in the same chapter, the focus is on the Wasserstein distance. Figure \ref{fig:ch08_wst_distr} shows the distribution of the Wasserstein distance over all the pairs of nodes.
 \\
This distributional representation of the users enables the embedding of the rating matrix into another graph (Figure \ref{fig:ch08_wst_graph}) in which the nodes are the users that are linked if their distribution of similarity are close enough, i.e., their Wasserstein distance is below a given threshold. \\
Let $G_w=(V_w,E_w)$ be the Wasserstein graph, then $V_w = \left\{ h(u_i) \right\}_{i=1,\ldots,M}$ is the set of users represented as histograms and $E_w = \left\{ (i,j) : \mathcal{W}(h(u_i ),h(u_j)) < \tau \right\}$ is the set of edges. The edge $(i,j)$ is then weighted based on the Wasserstein distance $\mathcal{W}(h(u_i),h(u_j))$ between the similarity distributions of nodes $i$ and $j$.
 \\
In the same way each item can be represented as the distribution of the similarity with all the other items and analogously, also the distributional representation of the items enables the embedding of $R$ in a graph, in which the nodes are the items. \\
To simplify the optimization process to find the top-$L$ recommendation lists the users have been clustered. \\
Firstly, the spectral clustering has been used on the original rating matrix. The resulting clusters have been mapped in the two previously defined graphs as shown in Figure \ref{fig:ch08_cluster_cos}. \\
As these figures highlight, the clusters obtained by the rating matrix are difficult to explain. For this reason, the spectral clustering have been then applied to the Wasserstein graph $G_w$. \\
 \\ \\
Figure \ref{fig:ch08_cluster_wst_2} shows the three resulting clusters. This results have been mapped back in the initial graph $G_c$ (Figure \ref{fig:ch08_cluster_wst_1}). Analyzing Figure \ref{fig:ch08_cluster_wst} it is clear that the blue cluster contains the more central users, therefore this cluster is highly connected to the other two. Instead, the lilac cluster is composed by the most different users.
 \\
\section{Objective functions and its distributional representation}
\begin{figure}[h]
    \centering
    \includegraphics[width=1\linewidth]{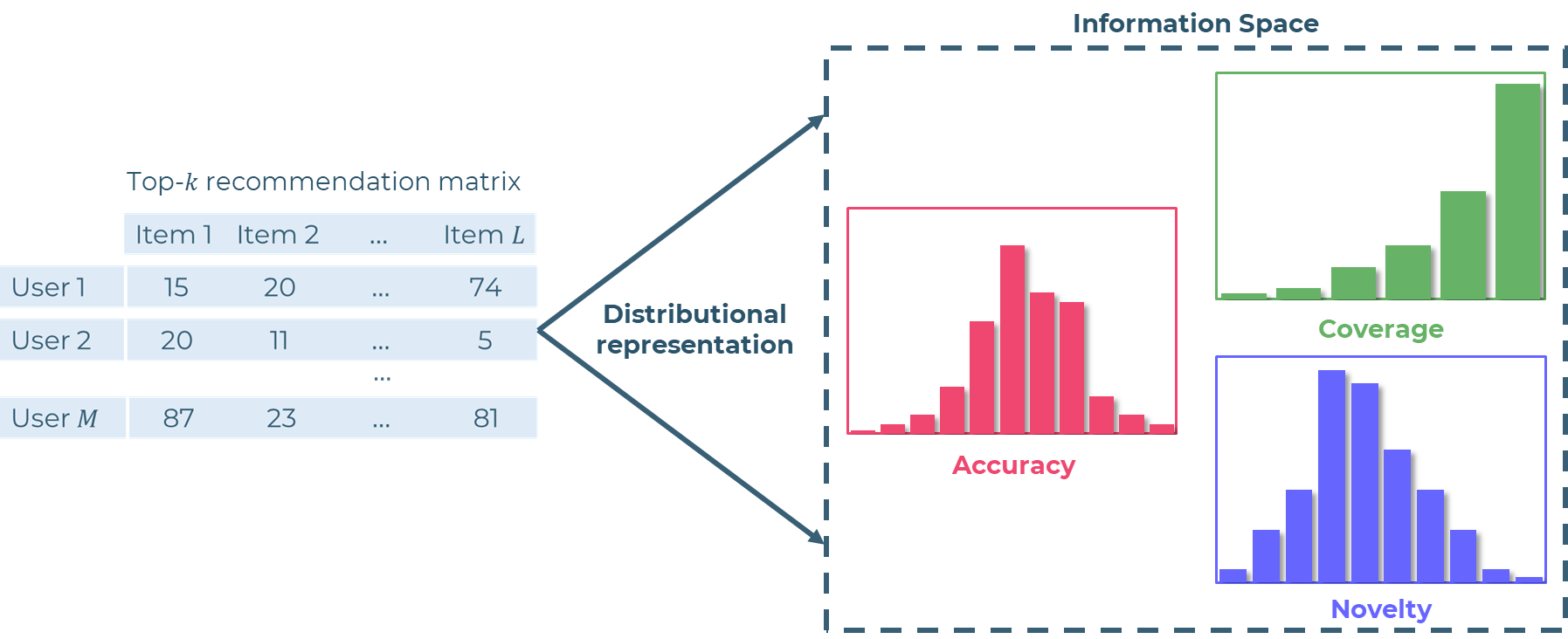}
    \caption{Distributional representation of a top-$L$ recommendation matrix and the information space.}
    \label{fig:ch08_info_space}
\end{figure}

For the problem of finding the optimal top-$L$ recommendation list, three conflicting objectives are considered, i.e., accuracy, coverage, and novelty. The distributional representation of these three metrics enables the definition of an information space (Figure \ref{fig:ch08_info_space}) that allows the use of MOEA/WST analyzed in Chapter \ref{ch06:moeawst}. In particular, each recommendation matrix can be represented by a three dimensional histogram that enables the selection process of MOEA/WST explained in Chapter \ref{ch06:selection}.
 \\
\subsubsection{Accuracy}
\begin{figure}[h]
    \centering
    \includegraphics[width=0.6\linewidth]{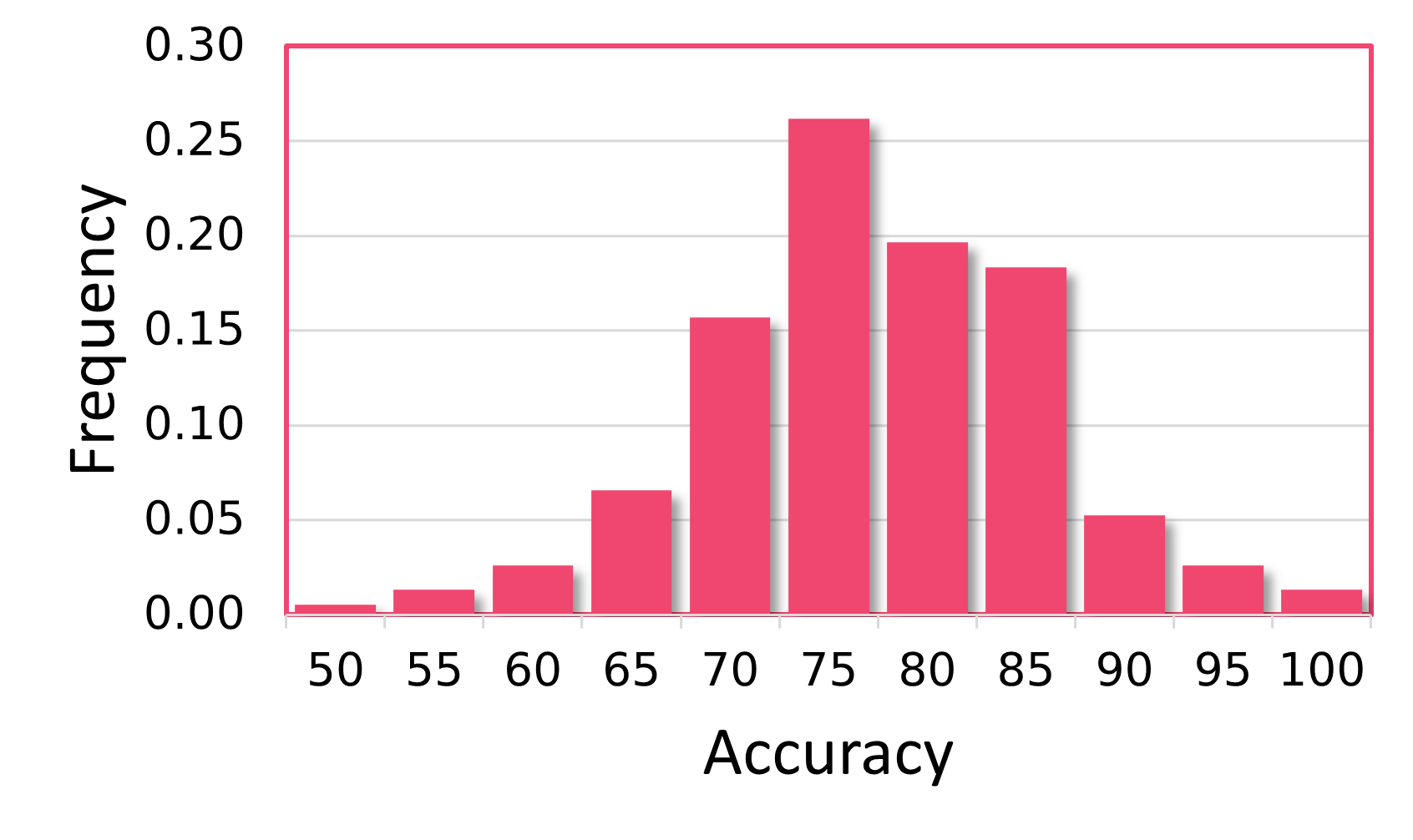}
    \caption{Example of accuracy distributions over the users.}
    \label{fig:ch08_acc_distr}
\end{figure}

The accuracy measures the similarity between the predicted rating and the true ratings. To each recommendation list, it is possible to assign a score that represents how ``good'' the items recommended to users are. This score is based on the sum of the ratings given by the users to the recommended items and is given by Equation \ref{eq:ch08_accuracy}:
\begin{equation}
    accuracy = \frac{1}{M \cdot L} \sum_{u_i \in U} \sum_{o_j \in S_L (u_i)} r(u_i, o_j)
    \label{eq:ch08_accuracy}
\end{equation}
where $r(u_i,o_j)$ is the rating given by user $u_i$ to item $o_j$. Maximize this score ensures that the recommendation list of each user contains only items that the user has given a high rating. \\
This particular definition of accuracy admits a distributional representation. The distribution is given by the values of accuracy of each user. This distribution can be represented by a histogram (Figure \ref{fig:ch08_acc_distr}) in which the support points $k_1,\ldots,k_{N_a}$ correspond to accuracy values, and the weights $w_{k_i}$ with $i=1,\ldots,N_a$ represent the fraction of users with a certain value of accuracy (Equation \ref{eq:ch08_acc_distr}). \\
\begin{equation}
    w_{k_i} = \frac{1}{M} \left| \left\{ u_i: \sum_{o_j \in S_L (u_i)} r(u_i, o_j) \in [k_i, k_{i+1}) \right\} \right|
    \label{eq:ch08_acc_distr}
\end{equation}
 \\
One problem with Collaborative Filtering recommendation is the ``popularity bias'': popular items are being recommended too frequently while most of the items do not get attention. For this reason, in this thesis the accuracy is considered together with other two objectives, coverage and novelty.

\subsubsection{Coverage}
\begin{figure}[h]
    \centering
    \includegraphics[width=0.6\linewidth]{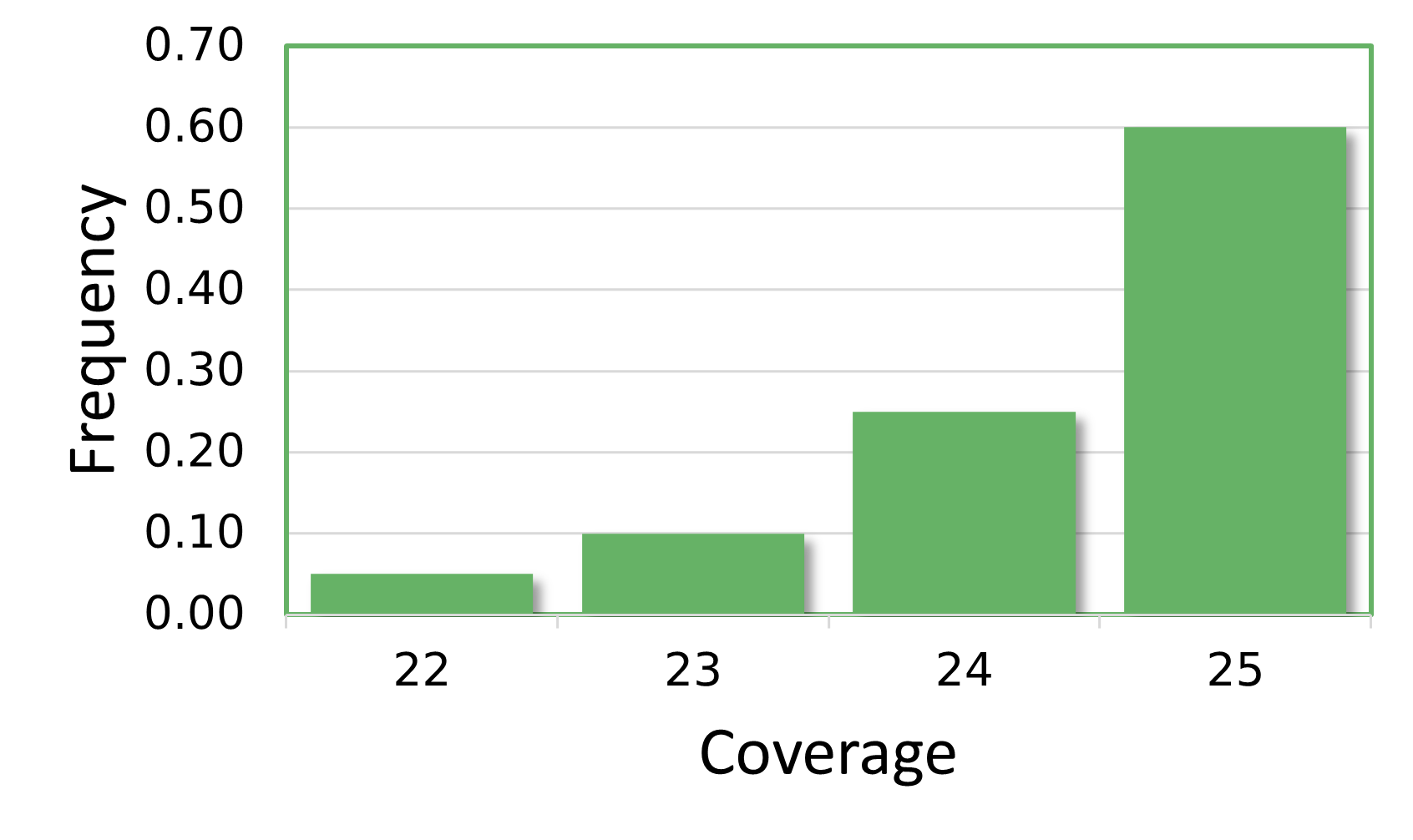}
    \caption{Example of coverage distributions over the users.}
    \label{fig:ch08_cov_distr}
\end{figure}

A recommender system is expected to provide $M$ recommendation lists. Each list corresponds to a user and consists of $L$ items. The coverage is defined as the number of different items in all users’ top-$L$ lists (Equation \ref{eq:ch08_coverage}).
\begin{equation}
    coverage = \frac{1}{N} \left| \bigcup_{u_i \in U} S_L(u_i) \right|
    \label{eq:ch08_coverage}
\end{equation}
The objective function coverage is averaged over the total number of items $N$. Coverage reflects the diversity of recommendation. A larger value of coverage is better because more choices are provided to the users. \\
It's important to note that also the coverage admits a distributional representation. The distribution is given by the ratio between the non-duplicated items in the recommendation list and the total number of items for each user, i.e., the coverage of the user recommendation list $S_L(u)$. This distribution can be represented by a histogram (Figure \ref{fig:ch08_cov_distr}) in which the support points are the values of coverage $k_1,\ldots,k_{N_c}$, and the weights  $w_{k_i}$ with $i = 1,\ldots,N_c$ represent the fraction of users with a certain value of coverage (Equation \ref{eq:ch08_cov_distr}).
\begin{equation}
    w_{k_i} = \frac{1}{M} \left| \left\{ u_i: \left| S_L(u_i) \right| \in [k_i, k_{i+1}) \right\} \right|
    \label{eq:ch08_cov_distr}
\end{equation}
 \\
\subsubsection{Novelty}
\begin{figure}[h]
    \centering
    \includegraphics[width=0.6\linewidth]{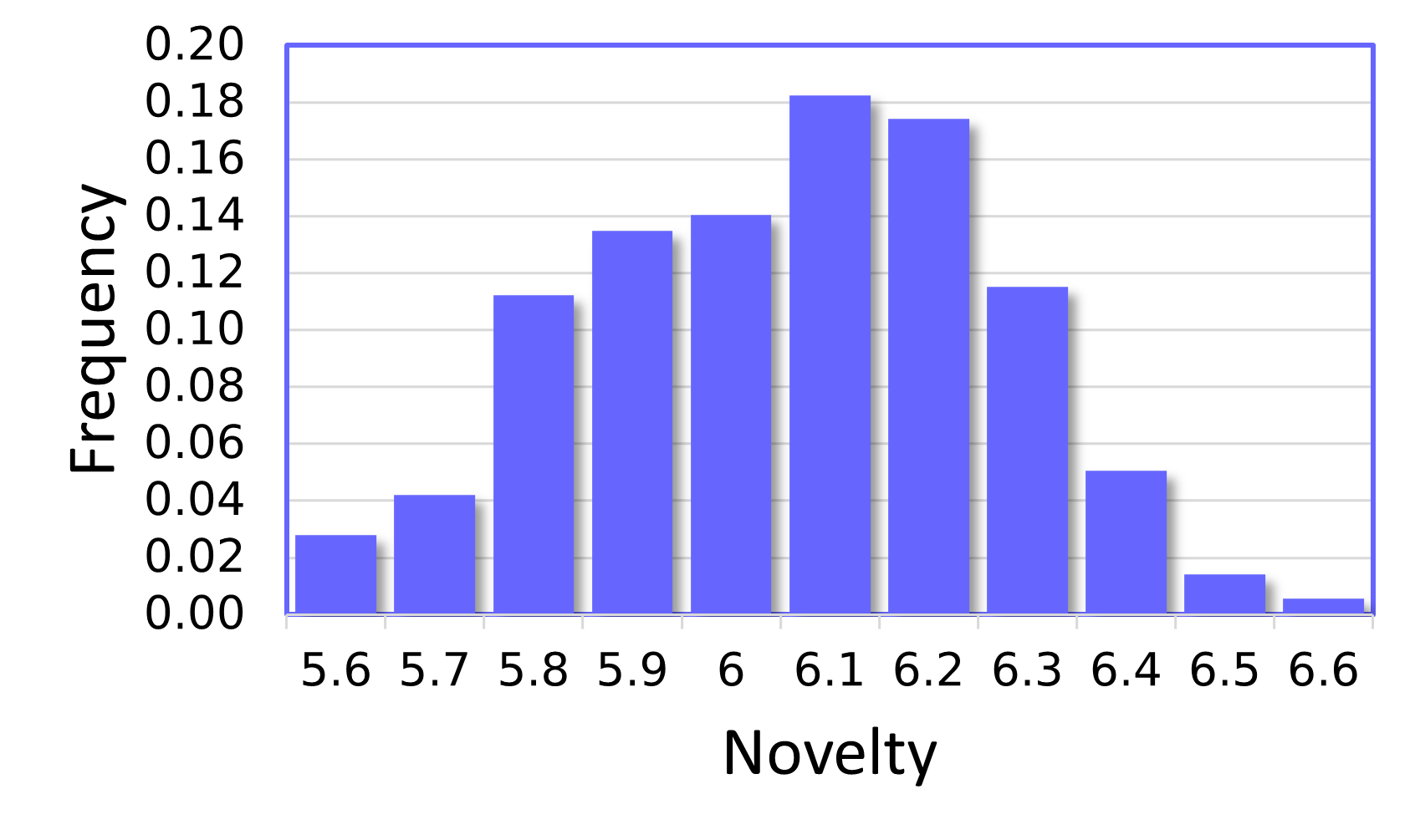}
    \caption{Example of novelty distributions over the users.}
    \label{fig:ch08_nov_distr}
\end{figure}

The novelty reflects the number of unknown items (i.e., items that are still unrated) that are recommended to users. This objective is based on the degree $d_j$ of an item $o_j$ that is the number of times it has been rated by a user. Then, the self-information \cite{DBLP:journals/cim/ZuoGZMJ15} of the item $o_j$ is given by Equation \ref{eq:ch08_self_info}.
\begin{equation}
    N_j = \log_2 \frac{M}{d_j}
    \label{eq:ch08_self_info}
\end{equation}
The novelty is then defined as the average self-information of all the items in the recommendation lists of each users (Equation \ref{eq:ch08_novelty}).
\begin{equation}
    Novelty = \frac{1}{M} \sum_{u_i \in U} \sum_{j \in S_L(u_i)} \frac{N_j}{L}
    \label{eq:ch08_novelty}
\end{equation}
It's important to note that novelty also admits a distributional representation. The distribution is given by the values of novelty $\sum_{i\in S_L (u)} \frac{N_i}{L}$ for each user $u$. This distribution can be represented by a histogram (Figure \ref{fig:ch08_nov_distr}) in which the support points $k_1,\ldots,k_{N_n}$ are the values of novelty, and the weights $w_{k_i}$ with $i = 1,\ldots,N_n$ represent the number of users with a certain value of novelty (Equation \ref{eq:ch08_nov_distr}).
\begin{equation}
    w_{k_i} = \frac{1}{M} \left| \left\{ u_i: \sum_{i\in S_L (u_i)} \frac{N_i}{L} \in [k_i, k_{i+1}) \right\} \right|
    \label{eq:ch08_nov_distr}
\end{equation}
 \\

\subsubsection{Information space}
The distributional representation of the three previously defined objective can be viewed as a three dimensional histogram. For each recommendation list $S_L$ the support points of this histogram are the values of accuracy along the $x$-axis, the values of coverage along the $y$-axis and the values of novelty along the $z$-axis; the weights represents the fraction of users whose values of accuracy, coverage and novelty fall in a specific range.\\
These distributions compose the so-called \textit{Information Space} on which the MOEA/WST algorithm is based. Therefore, it uses the Wasserstein distance to compare the histograms associated to different top-$L$ recommendation lists, in the selection operator, to speed up the entire optimization process.

\section{Computational settings}
In MOEA/WST the top-$L$ rating lists are encoded as $L \times M$ integer matrices. The entries of these matrices are the integer id of recommended items. \\
To recombine chromosome, the Pymoo implementation of \textit{Simulated Binary Crossover} \cite{DBLP:conf/gecco/DebSO07} is used. It simulates the working principle of the single-point crossover operator on binary data, by using a probability distribution. Pymoo allows the definition of a parameter $\eta$ to fine-tune the exponential distribution, in these experiments $\eta = 3.0$. As mutation operator, the Pymoo implementation of \textit{Inverse Mutation} \cite{DBLP:conf/gecco/DebSO07} is used. This mutation is applied to permutations, and it randomly select a segment of a chromosome and reverse its order. For instance, for the permutation $[1,2,3,4,5]$ the segment can be $[2,3,4]$ which results in $[1,4,3,2,5]$.\\
In these experiments, a population of 40 individuals is considered and at each generation an offspring of 10 new chromosomes is generated for a total of 50 generations. The initial population is sampled randomly. For a fair comparison between MOEA/WST and NSGA-II the same settings are used in both algorithms. 

\section{Computational results}
\begin{figure}[h]
\centering
    \begin{subfigure}{.32\textwidth}
        \centering
        \includegraphics[width=1\linewidth]{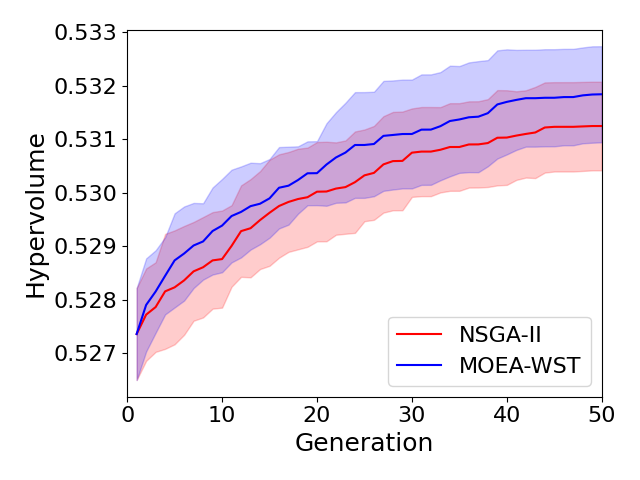}
        \caption{Cluster 1.}
    \end{subfigure}
    \begin{subfigure}{.32\textwidth}
        \centering
        \includegraphics[width=1\linewidth]{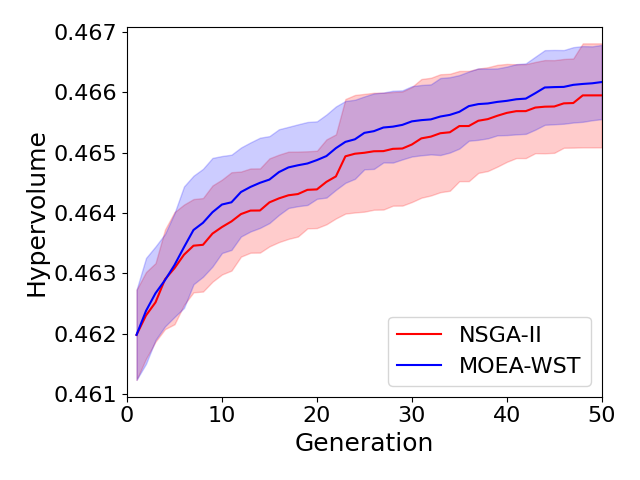}
        \caption{Cluster 2.}
    \end{subfigure}
    \begin{subfigure}{.32\textwidth}
        \centering
        \includegraphics[width=1\linewidth]{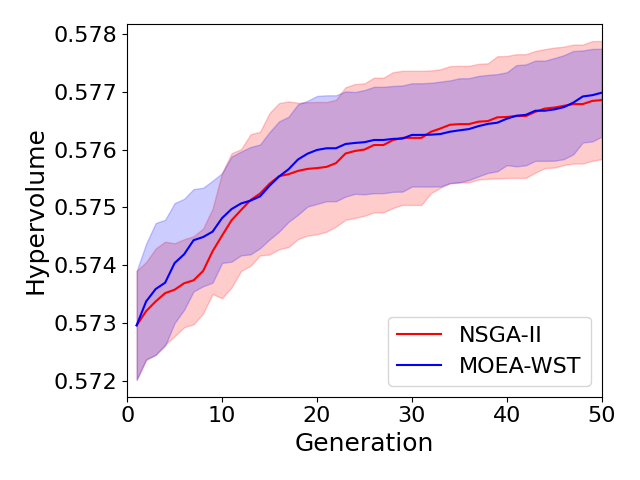}
        \caption{Cluster 3.}
    \end{subfigure}
\caption{Mean and standard deviation of the hypervolume over the generations. Clusters are the ones obtained using the spectral clustering on the rating matrix.}
\label{fig:ch08_hv_cluster_rating}
\end{figure}

\begin{figure}[h]
\centering
    \begin{subfigure}{.32\textwidth}
        \centering
        \includegraphics[width=1\linewidth]{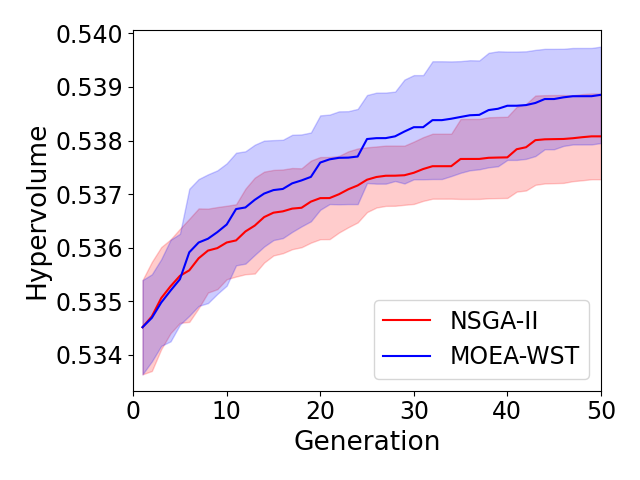}
        \caption{Cluster 1.}
    \end{subfigure}
    \begin{subfigure}{.32\textwidth}
        \centering
        \includegraphics[width=1\linewidth]{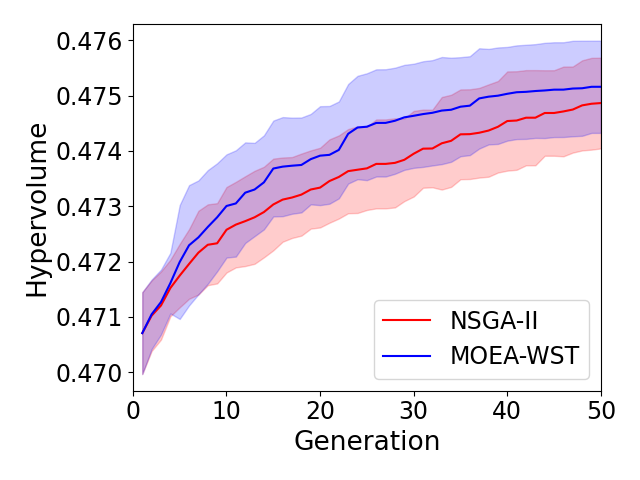}
        \caption{Cluster 2.}
    \end{subfigure}
    \begin{subfigure}{.32\textwidth}
        \centering
        \includegraphics[width=1\linewidth]{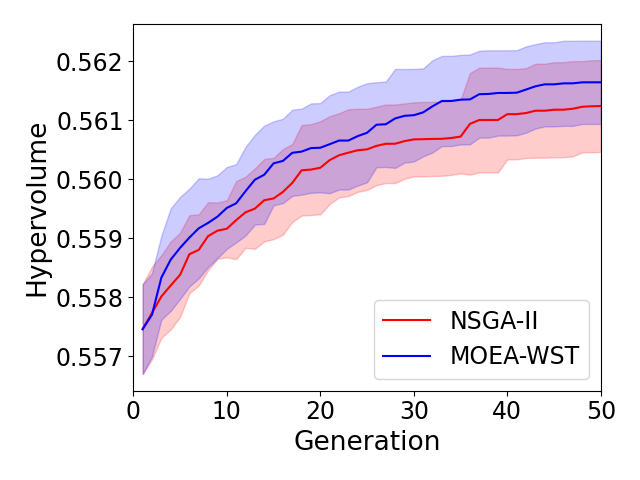}
        \caption{Cluster 3.}
    \end{subfigure}
\caption{Mean and standard deviation of the hypervolume over the generations. Clusters are the ones obtained using the spectral clustering on the Wasserstein graph.}
\label{fig:ch08_hv_cluster_wst}
\end{figure}

\begin{figure}[h]
\centering
    \begin{subfigure}{.32\textwidth}
        \centering
        \includegraphics[width=1\linewidth]{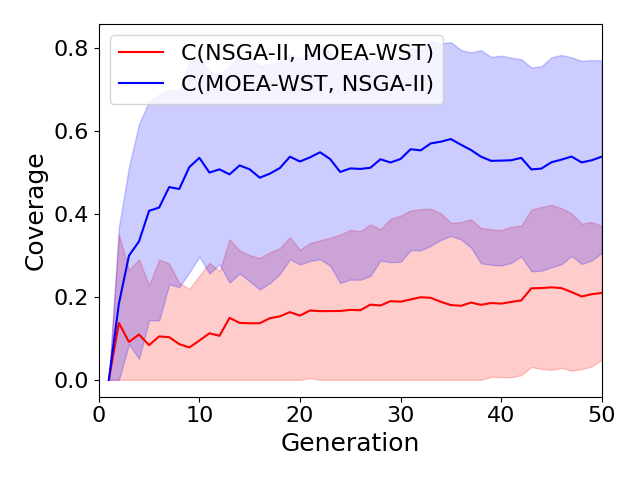}
        \caption{Cluster 1.}
    \end{subfigure}
    \begin{subfigure}{.32\textwidth}
        \centering
        \includegraphics[width=1\linewidth]{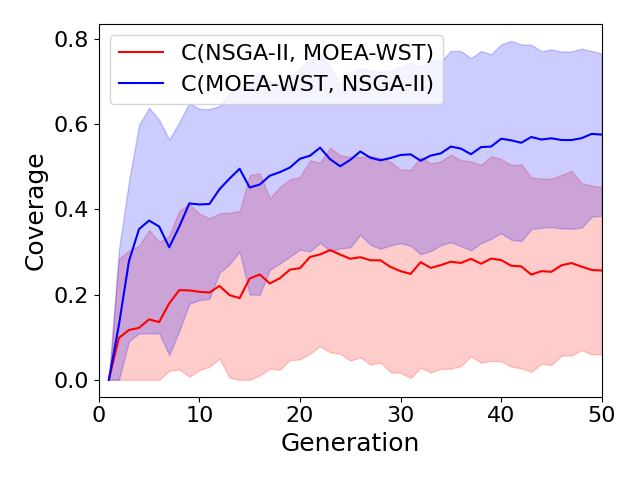}
        \caption{Cluster 2.}
    \end{subfigure}
    \begin{subfigure}{.32\textwidth}
        \centering
        \includegraphics[width=1\linewidth]{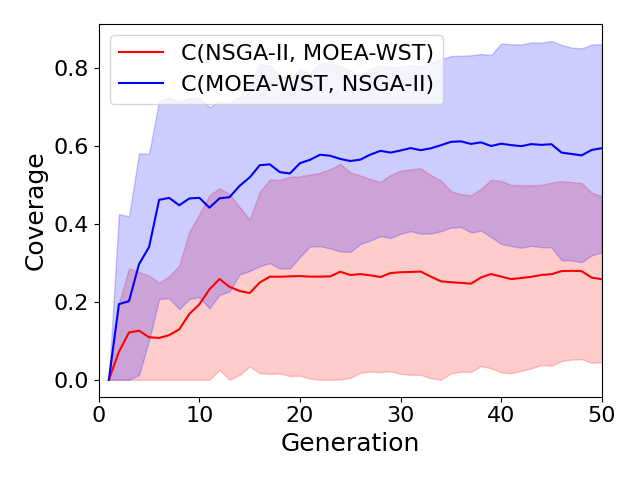}
        \caption{Cluster 3.}
    \end{subfigure}
\caption{Mean and standard deviation of the coverage over the generations. Clusters are the ones obtained using the spectral clustering on the rating matrix.}
\label{fig:ch08_cov_cluster_rating}
\end{figure}

\begin{figure}[h]
\centering
    \begin{subfigure}{.32\textwidth}
        \centering
        \includegraphics[width=1\linewidth]{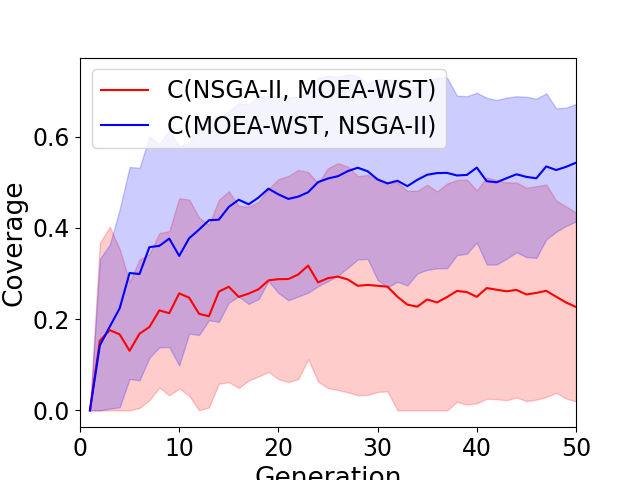}
        \caption{Cluster 1.}
    \end{subfigure}
    \begin{subfigure}{.32\textwidth}
        \centering
        \includegraphics[width=1\linewidth]{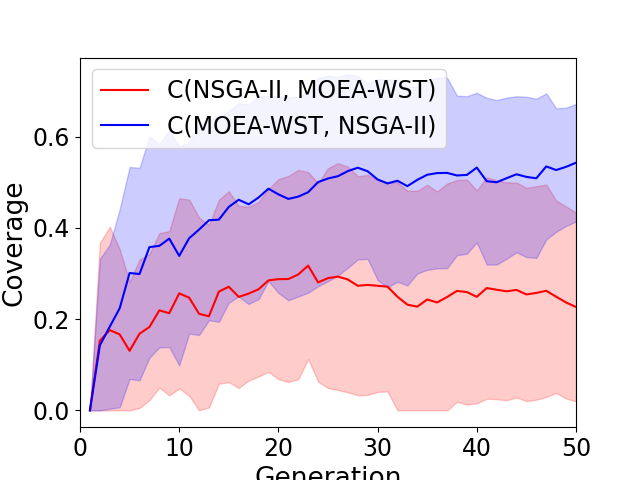}
        \caption{Cluster 2.}
    \end{subfigure}
    \begin{subfigure}{.32\textwidth}
        \centering
        \includegraphics[width=1\linewidth]{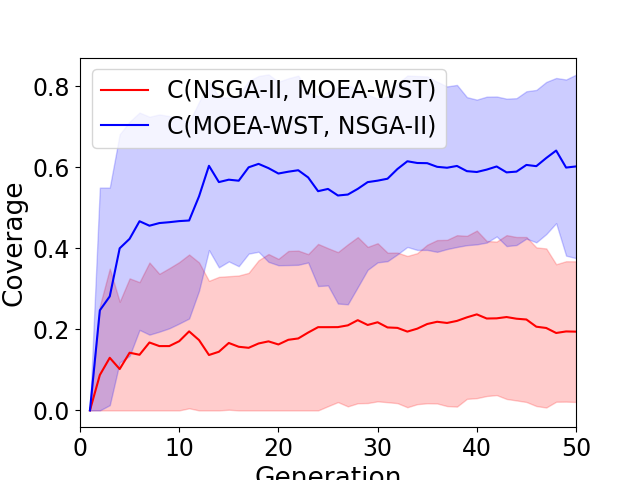}
        \caption{Cluster 3.}
    \end{subfigure}
\caption{Mean and standard deviation of the coverage over the generations. Clusters are the ones obtained using the spectral clustering on the Wasserstein graph.}
\label{fig:ch08_cov_cluster_wst}
\end{figure}

In this section, the computational results over the MovieLens dataset are reported. \\
First, the two algorithms, NSGA-II and MOEA/WST, have been used on the clusters resulting from the rating matrix. Figure \ref{fig:ch08_hv_cluster_rating} shows the Hypervolume over generation of both, NSGA-II (red) and MOEA/WST (blue) on the clusters computed on the original dataset (rating matrix). Since multiple runs of the algorithms are performed, the charts display mean and standard deviation of the metric. \\
 \\
Then the two algorithms have been used, with the same settings, on the clusters obtained from the Wasserstein graph. Figure \ref{fig:ch08_hv_cluster_wst} shows the same comparative results where the clusters are computed on the WST graph.
 \\ \\
In both cases the Hypervolume curve of MOEA/WST converge faster than in the case of NSGA-II. It is also important to note that, using the clusters over the Wasserstein graph, both algorithms perform well. \\
Figures \ref{fig:ch08_cov_cluster_rating}, \ref{fig:ch08_cov_cluster_wst} display the curve of coverage as the function of number of generations. As in the case of sensor placements, the advantage of MOEA/WST over NSGA-II is given more significant in terms of coverage.
 \\
 \\

%% file: chapters/chapter09.tex
\chapter{Software Resources}
\label{ch09:sw}
In this chapter the library and tools used in the experiments are presented. Python is chosen as the programming language thanks to its flexibility and to the high number of libraries related to machine learning and optimization problems.

\section{Pymoo}
The main library considered to model and to solve the multi-objective problems is \textit{Pymoo} \cite{DBLP:journals/access/BlankD20} (Python Multi-objective Optimization). This framework offers state of the art single and multi-objective optimization algorithms and many more features related to multi-objective optimization, such as visualization and decision making. Among all the genetic algorithms offered by \textit{Pymoo}, NSGA-II is considered in the experiments presented in the previous chapters.
The strength of \textit{Pymoo} is its flexibility; therefore, it allows the definition of customized problems as well as new algorithms and genetic operators based on consolidated structures. These functionalities of \textit{Pymoo} are used to develop the MOEA/WST algorithm defined in Chapter \ref{ch06:moeawst}. \\
\textit{Pymoo} also offers a rich set of performance indicator for multi-objective problems, among which the hypervolume is used.

\section{BoTorch}
\label{ch09:botorch}
In the experiments in Chapter \ref{ch07:water} the \textit{q}ParEGO \cite{DBLP:conf/nips/DaultonBB20} implementation of \textit{BoTorch} is used. \textit{BoTorch} \cite{DBLP:conf/nips/BalandatKJDLWB20} is a library for Bayesian Optimization built on the well known open source machine learning framework \textit{PyTorch}. This choice is driven by the fact that \textit{BoTorch} is compatible with the problems defined using \textit{Pymoo}. In this way, the same problem can be solved using algorithms of both, Pymoo and BoTorch frameworks.

\section{Python Optimal Transport}
Different libraries related to the Wasserstein distance exist in Python, for instance \textit{SciPy} \cite{2020SciPy-NMeth} or \textit{GUDHI} \cite{DBLP:conf/icms/MariaBGY14}. Python Optimal Transport (\textit{POT}) \cite{flamary2021pot} is one of the most complete frameworks. It provides several solvers for optimization problems related to Optimal Transport. Among all the functionalities offered by this library the one explored and used in this thesis is the linear programming solver for the Earth Mover’s Distance in the case of single and multi-dimensional discrete distributions. These functions to compute the EMD are the core of the MOEA/WST algorithm implemented in \textit{Pymoo}. \\
\textit{POT} also contains some approximations of the Wasserstein distance. Examples are the entropic regularization OT solver based on Sinkhorn \cite{DBLP:conf/nips/Cuturi13} or the implementation of the sliced Wasserstein distance \cite{DBLP:journals/jmiv/BonneelRPP15}. These implementations are significantly more efficient in terms of computation time than the one based on linear programming but presents problem of numerical instability. Another useful functionality of POT is the computation of the Wasserstein barycenters \cite{DBLP:journals/siamma/AguehC11} even in the case of distributions with different supports \cite{DBLP:conf/icml/CuturiD14}.

\section{Water Networks Tool for Resilience}
The Water Network Tool for Resilience (\textit{WNTR}, pronounced winter) \cite{klise2018overview} is a Python package designed to simulate and analyse resilience of water distribution networks. \textit{WNTR} has an application programming interface (API) that is flexible and allows for changes to the network structure and operations, along with simulation of disruptive incidents and recovery actions. It is based upon EPANET, which is a tool to simulate the movement and fate of drinking water constituents within distribution systems. \\
Over all the functionalities of \textit{WNTR}, the hydraulic and water quality simulation are used to simulate the flow of a contaminant in different water distribution networks. Therefore, a water quality simulation can be used to compute the percentage of contaminant flow originating from a specific location. In the experiments reported in Chapter \ref{ch07:water} multiple simulations are executed for each network changing the location in which the contaminant is injected.

\section{Cytoscape and ClusterMaker}
\textit{Cytoscape} \cite{shannon2003cytoscape} is an open source software platform for visualizing complex networks and integrating these with any type of attribute data. It is mainly used in field like biology to analyse molecular and genomics systems, and social science, but \textit{Cytoscape} is domain-independent and therefore it is a powerful tool for complex network analysis in general. The key feature of \textit{Cytoscape} is that it is expandable and extensible; therefore, there exist hundreds of third parties’ plugins that extend its functionalities. \\
\textit{ClusterMaker2} \cite{DBLP:journals/bmcbi/MorrisANBWSBF11} is one of these plugins and it unifies different clustering techniques. In this thesis, the spectral clustering implementation of \textit{ClusterMaker2} is used in the networks related to the recommender systems in Chapter \ref{ch08:recommender}. As any spectral methods it uses the eigenvalues in an input similarity matrix to perform dimensionality reduction for clustering in fewer dimensions using a standard \textit{k}-means algorithm. In the case of graph clustering, the normalized Laplacian matrix is used as similarity matrix. Consider a graph $G$ the normalized Laplacian is defined as $L_N(G) = D^{-\frac{1}{2}} S(G) D^{-\frac{1}{2}}$ where $D$ is the degree matrix and $S(G) = I + A(G)$ with $I$ the identity matrix and $A(G)$ the adjacency matrix.

%% file: chapters/chapter10.tex
\chapter{Conclusions}
\label{ch10:conclusion}
The key objective of this work is to show that the distributional framework for data analysis and in particular the Wasserstein distance and the associated analytical methods offer significant advantages over comparing distributions using a set of parametric values such as mean, variance and higher order moments. Indeed, the analysis of these parameters does not take the whole distribution into account. \\
Even if the roots of WST are in abstract spaces of probability distributions, WST and the associated optimal transport map offer a visually intuitive representation of the similarity between distributions. These advantages come at a considerable computational cost in particular in the multi-dimensional case. To overcome this drawback substantial research work has been conducted in the last years to provide feasible algorithm to compute barycenters and, as a consequence, to enable clustering in WST spaces. These results have allowed to widen the application of WST beyond the domain of imaging science, which was an early adopter of WST, to many applications in data analysis and machine learning, for instance in the generation of adversarial deep networks. \\
In this thesis the representation power of WST has been applied to multi-task learning on networks modelled as a multi-objective optimization problem. \\
This new approach has been instantiated in different domains as networked infrastructure, specifically water distribution networks and collaborative filtering in recommender systems.
WST has also enabled new operators in multi-task evolutionary learning for optimal sensor placement, for intrusion detection, and for the design of multi-objective recommender systems. \\
Beside the specific issues analyzed in this thesis, distributional inputs can occur in a number of practical situations. As already said, compare parametric features associated to distributions is not always the right way. Commonly used kernels depend on the Euclidean distance between points. In the case of distributional inputs, it is important to construct positive definite kernels on sets of probability measures. The research on this topic has branched into two directions. \\
A first solution is to build a Wasserstein induced kernel, that has been proved to be effective for instance in the problem of neural architectural search \cite{DBLP:conf/nips/KandasamyNSPX18}. Using kernels limits the choice of distribution distances as the resulting kernel has to be definite positive: a widely used distance as Kullback-Leibler does not qualify. The key difficulty of a Wasserstein kernel is that the kernel obtained computing the exponential of the squared Wasserstein distance between distributional inputs does not lead to a positive definite kernel. Therefore, it has been shown that many eigenvalues of the Wasserstein induced covariance matrix are negative. \\
A very interesting perspective consists in embed directly the probabilistic distances in the learning process, without using a kernel. In this thesis a distributional distance based learning paradigm has been shown to be very effective in simulation-optimization problem like optimal sensor placement in Chapter \ref{ch07:water} or in the context of recommender systems in Chapter \ref{ch08:recommender}. \\
These results agree with very recent results as to embed non-Euclidean distance in the learning process and offer new insights into the design of algorithms of machine learning and optimization. 

%% file: chapters/appendix.tex
\begin{appendices}
\chapter{Metrics on graphs}
In the following the centrality measures and some vulnerability metrics on graphs are defined.
\section{Centrality measures}
\label{apx:centrality}
Consider a graph $G = (V, E)$ with $|V| = n$ and $|E| = m$. \\
The density $q$ of the network is the fraction of edges which are present in the network (Equation \ref{eq:apx_density}).
\begin{equation}
    q = \frac{2m}{n(n-1)}
    \label{eq:apx_density}
\end{equation}
The link-per-node ration $e$ of a graph is defined as the number of edges with respect to the number of its nodes (Equation \ref{eq:apx_ratio}).
\begin{equation}
    e = \frac{m}{n}
    \label{eq:apx_ratio}
\end{equation}
The diameter $D(G)$ of a network is defined as the largest distance (the largest shortest path) among each possible pair of nodes in the graph. Instead, the characteristic path length $L(G)$ is the average distance for every possible pair of nodes as in Equation \ref{eq:apx_path_lenght}:
\begin{equation}
    L(G) = \frac{1}{n(n - 1)} \sum_{i=1}^n \sum_{j=1 : j \neq i}^n d(i, j)
    \label{eq:apx_path_lenght}
\end{equation}
where $d(i, j)$ is the distance between node $i$ and node $j$. \\
The betweenness centrality $b_i$ of a node $i$ is defined as (Equation \ref{eq:apx_betwe}):
\begin{equation}
    b_i = \frac{1}{n^2} \sum_{s=1}^n\sum_{t=1}^n \eta_{st}^i
    \label{eq:apx_betwe}
\end{equation}
where $\eta_{st}^i = 1$ if node $i$ lies on the shortest path from $s$ to $t$ and $0$ otherwise. \\
The central point dominance $c_b'$, based on betweenness centrality, is a measure for characterizing the organization of a network according to its path-related connectivity and is defined as (Equation \ref{eq:apx_central}):
\begin{equation}
    c_b' = \frac{1}{n - 1} \sum_{i = 1}^n (b_{max} - b_i)
    \label{eq:apx_central}
\end{equation}
where $b_i$ is the betweenness centrality of the node $i$ and $b_{max}$ is the maximum value of betweenness centrality over all the network's nodes. \\
The clustering coefficient $cc$ is the number of triangles $N_{tg}$ with respect to the overall number of possible connected triples $N_{tp}$, where a triple consists of three nodes connected at least by two edges while a triangle consists of three nodes connected exactly by three edges (Equation \ref{eq:apx_cluster}).
\begin{equation}
    cc = \frac{N_{tg}}{N_{tp}}
    \label{eq:apx_cluster}
\end{equation}

\section{Vulnerability measures}
\label{apx:vulnerability}
Consider a graph $G = (V, E)$ with $|V| = n$. The performance of the network after the removal of nodes/edges is often evaluated as the change of the efficiency, as in Equation \ref{eq:apx_efficiency}.
\begin{equation}
    E = \frac{1}{n(n - 1)} \sum_{ij \in V : i \neq j} \frac{1}{d_{ij}}
    \label{eq:apx_efficiency}
\end{equation}
where $d_{ij}$ represents the distance between $i$ and $j$. Normalization by $n(n-1)$ ensures that $E \leq 1$, in case of unweighted graph. The maximum value $E=1$, is assumed if and only if the graph is complete. \\
A way to measure the vulnerability of the network is using the loss of efficiency observed when we remove some nodes/edges. The relative drop in the network efficiency caused by the removal of a node i from the graph is defined as (Equation \ref{eq:apx_efficiency_loss}):
\begin{equation}
    C_{\Delta}^E(i) = \frac{E(G)}{E(G\setminus\{v\})}
    \label{eq:apx_efficiency_loss}
\end{equation}
where $G\setminus\{v\}$ denotes the network $G$ without the node $i$. The vulnerability of $G$ is defined as (Equations \ref{eq:apx_vmax} and \ref{eq:apx_vmean}):
\begin{gather}
    V_{MAX}(G) = \max_{i \in V} C_{\Delta}^E(i) \label{eq:apx_vmax}\\
    V_{MEAN}(G) = \frac{1}{n}\sum_{i \in V} C_{\Delta}^E(i) \label{eq:apx_vmean}
\end{gather}
Analogous formulas can be written removing the edges.

\chapter{Computational Results}
\label{apx:result}
In the following the complete computational results are reported.
\section{Sensor placement in Hanoi WDN}
\label{apx:hanoi}
\begin{figure}[H]
\centering
    \begin{subfigure}{.49\textwidth}
        \centering
        \includegraphics[width=1\linewidth]{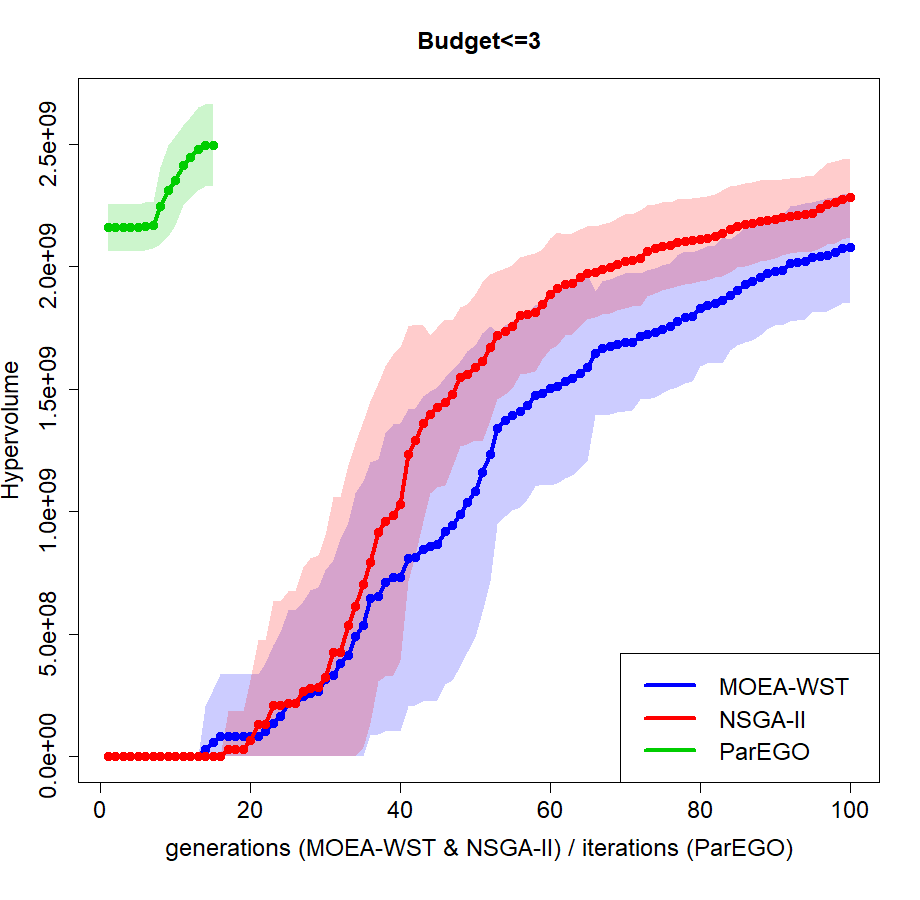}
        \caption{Hypervolume over generation.}
    \end{subfigure}
    \begin{subfigure}{.49\textwidth}
        \centering
        \includegraphics[width=1\linewidth]{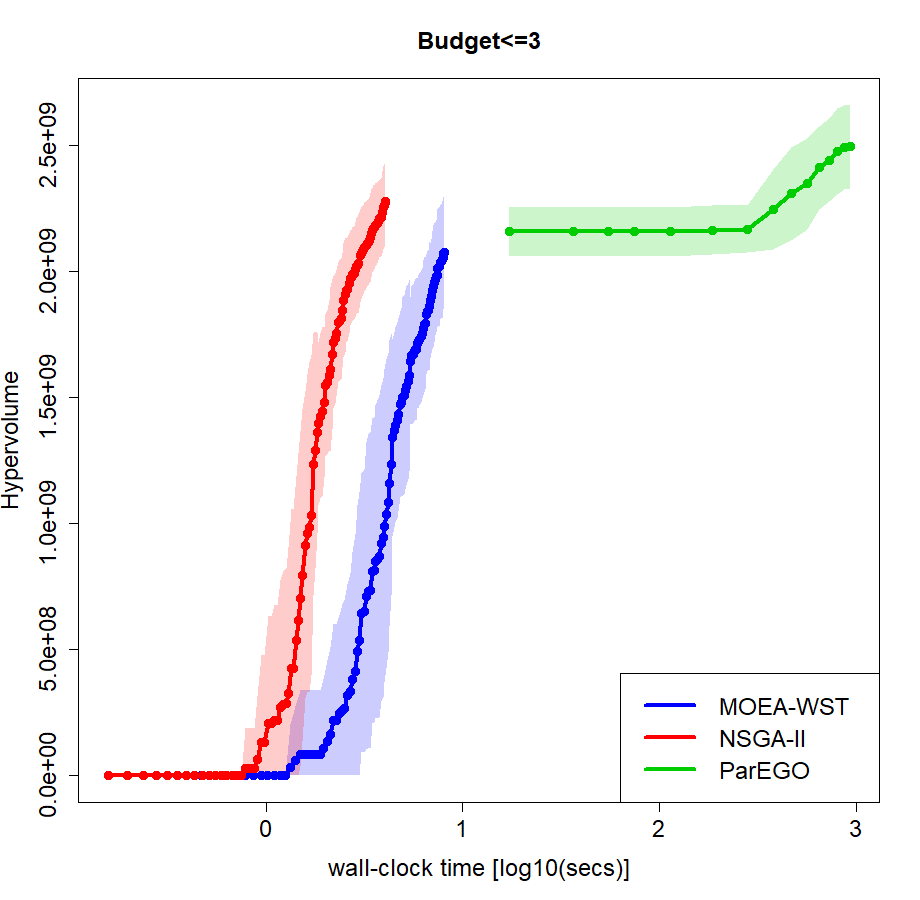}
        \caption{Hypervolume over wall-clock time.}
    \end{subfigure}
\caption{Hypervolume curves of the three algorithms in the case of budget $\leq 3$.}
\label{fig:apx_hanoi_3}
\end{figure}

\begin{figure}[H]
\centering
    \begin{subfigure}{.49\textwidth}
        \centering
        \includegraphics[width=1\linewidth]{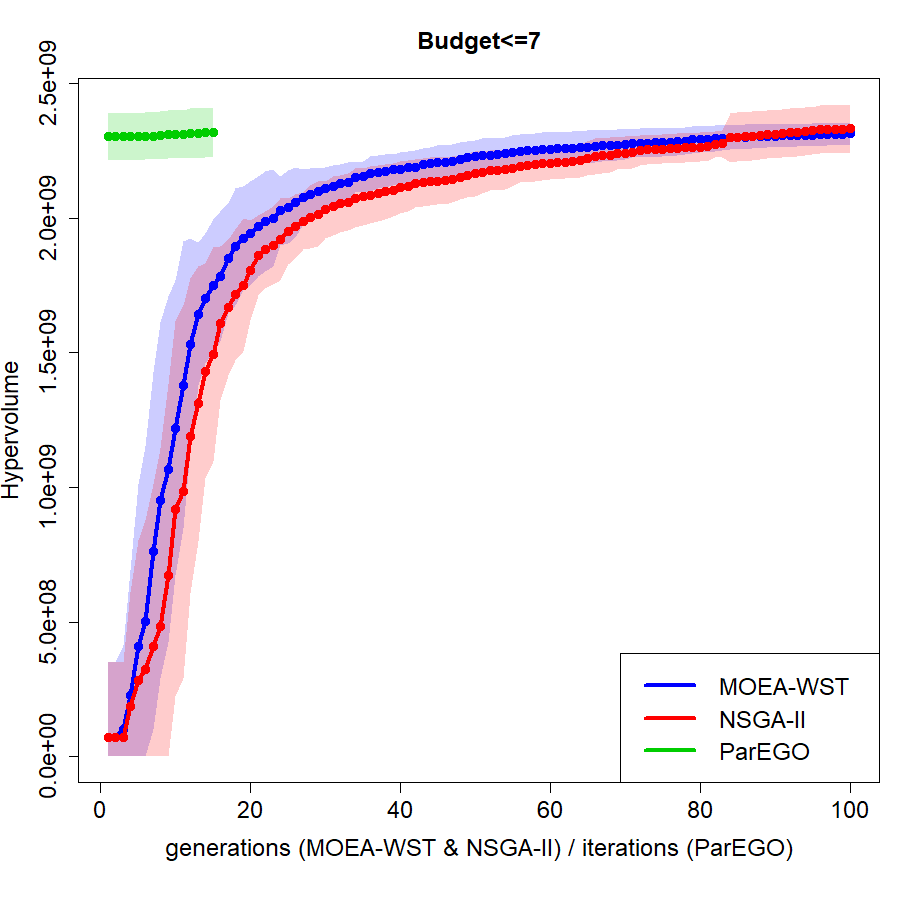}
        \caption{Hypervolume over generation.}
    \end{subfigure}
    \begin{subfigure}{.49\textwidth}
        \centering
        \includegraphics[width=1\linewidth]{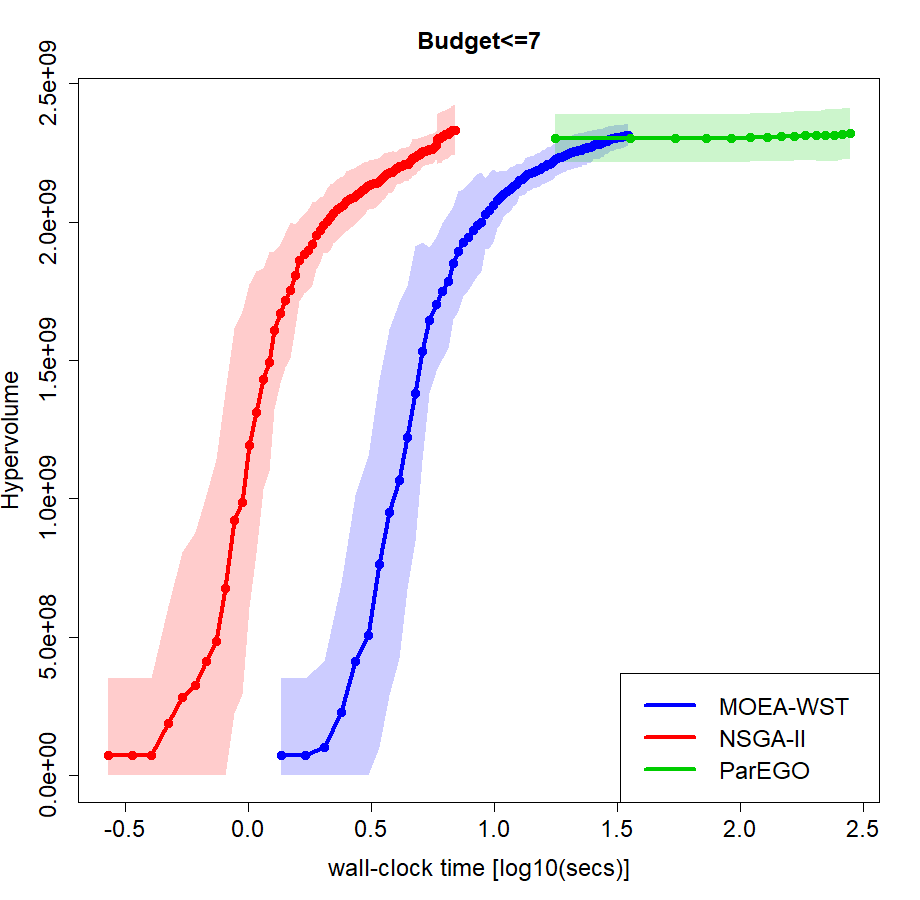}
        \caption{Hypervolume over wall-clock time.}
    \end{subfigure}
\caption{Hypervolume curves of the three algorithms in the case of budget $\leq 7$.}
\label{fig:apx_hanoi_7}
\end{figure}

\begin{figure}[H]
\centering
    \begin{subfigure}{.49\textwidth}
        \centering
        \includegraphics[width=1\linewidth]{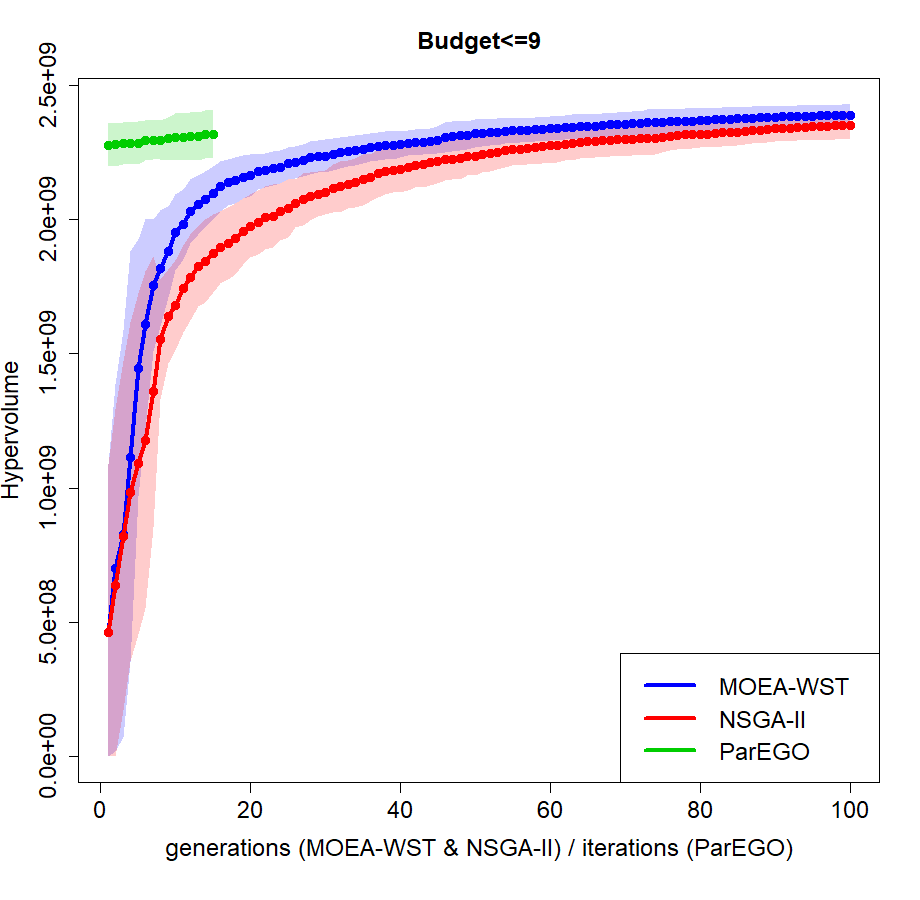}
        \caption{Hypervolume over generation.}
    \end{subfigure}
    \begin{subfigure}{.49\textwidth}
        \centering
        \includegraphics[width=1\linewidth]{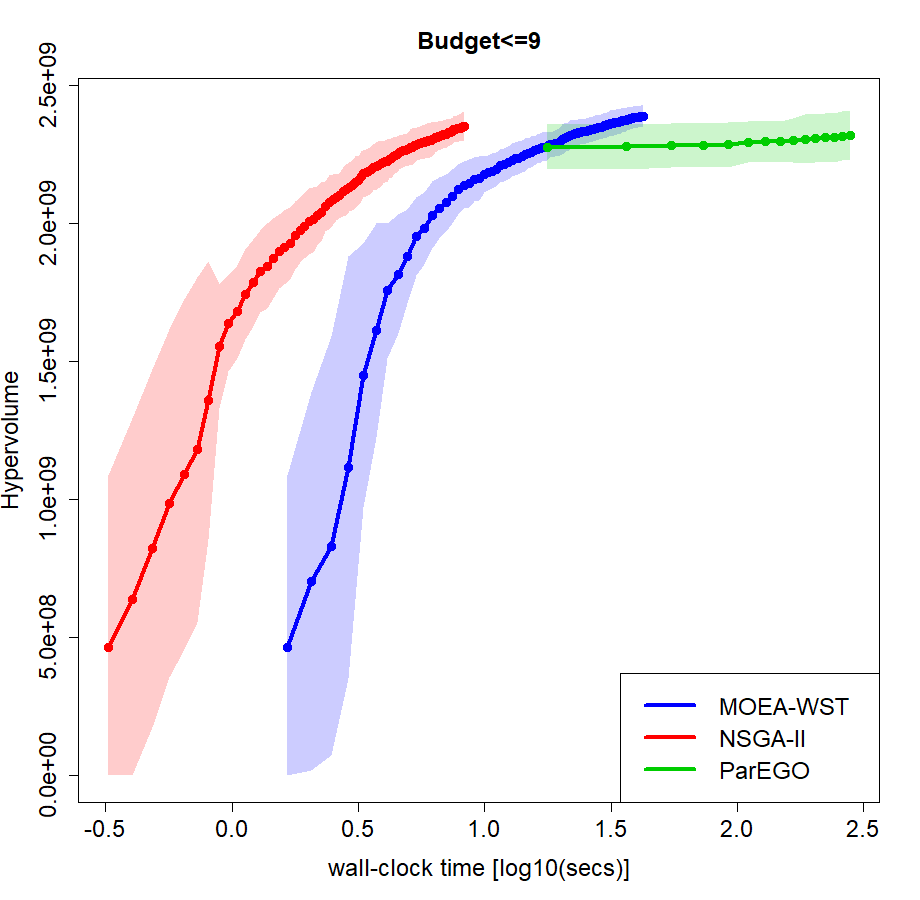}
        \caption{Hypervolume over wall-clock time.}
    \end{subfigure}
\caption{Hypervolume curves of the three algorithms in the case of budget $\leq 9$.}
\label{fig:apx_hanoi_9}
\end{figure}

\begin{figure}[H]
\centering
    \begin{subfigure}{.49\textwidth}
        \centering
        \includegraphics[width=1\linewidth]{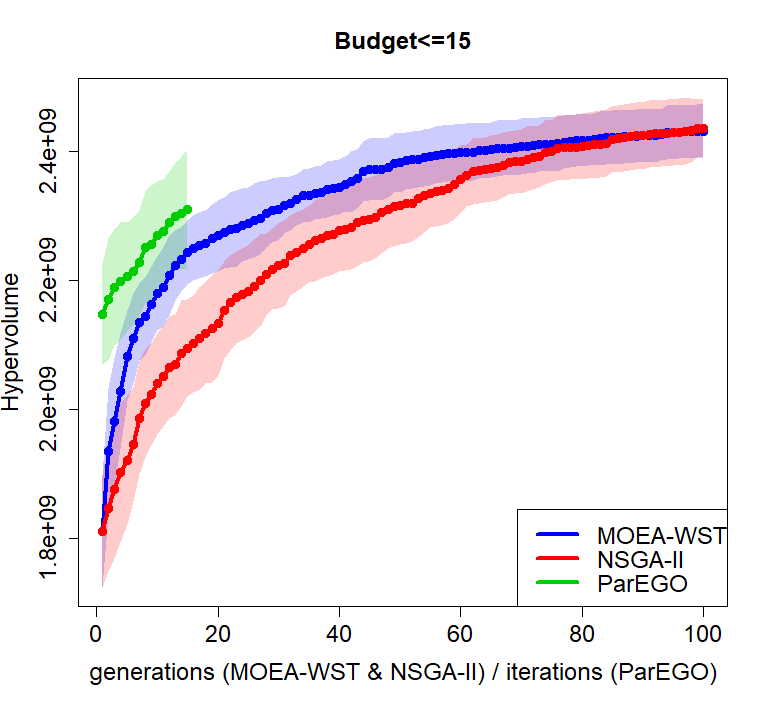}
        \caption{Hypervolume over generation.}
    \end{subfigure}
    \begin{subfigure}{.49\textwidth}
        \centering
        \includegraphics[width=1\linewidth]{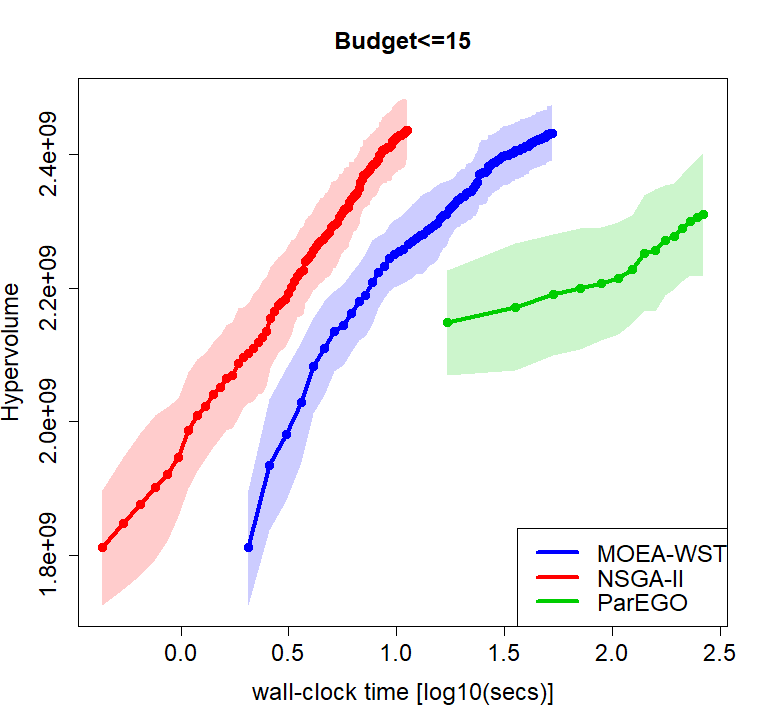}
        \caption{Hypervolume over wall-clock time.}
    \end{subfigure}
\caption{Hypervolume curves of the three algorithms in the case of budget $\leq 15$.}
\label{fig:apx_hanoi_15}
\end{figure}

\begin{figure}[H]
\centering
    \begin{subfigure}{.49\textwidth}
        \centering
        \includegraphics[width=1\linewidth]{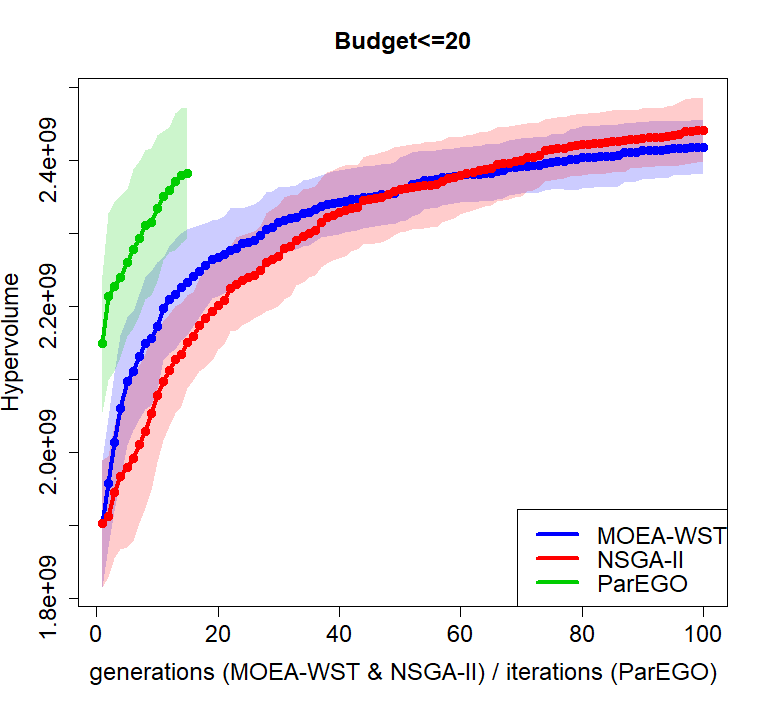}
        \caption{Hypervolume over generation.}
    \end{subfigure}
    \begin{subfigure}{.49\textwidth}
        \centering
        \includegraphics[width=1\linewidth]{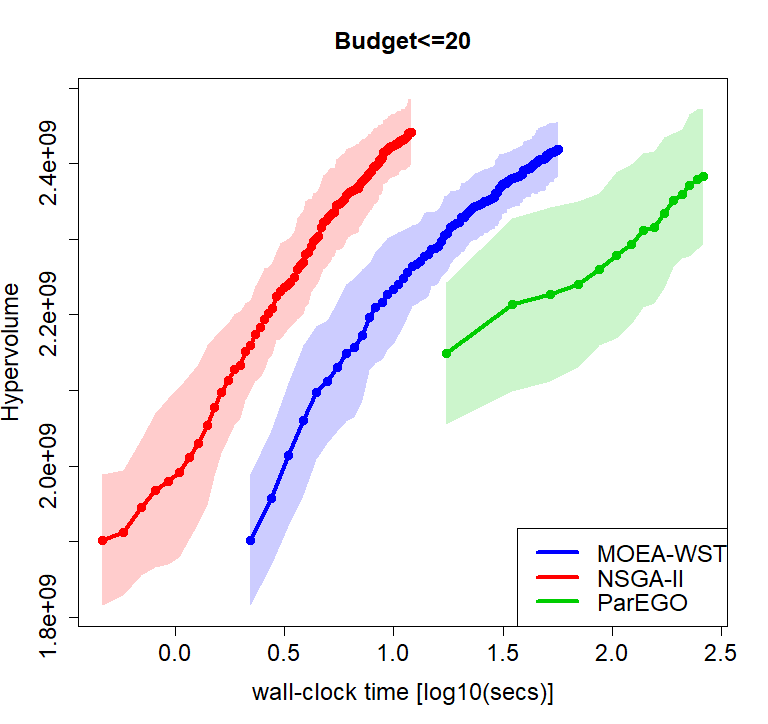}
        \caption{Hypervolume over wall-clock time.}
    \end{subfigure}
\caption{Hypervolume curves of the three algorithms in the case of budget $\leq 20$.}
\label{fig:apx_hanoi_20}
\end{figure}

Figures \ref{fig:apx_hanoi_3} - \ref{fig:apx_hanoi_20} show the comparison between the three algorithms, MOEA/WST, NSGA-II and ParEGO, in terms of hypervolume over generations and wall-clock time for different sensor budgets $p$.\\
\noindent
Table \ref{tab:apx_wilcoxon} reports the results of Wilcoxon test between the hypervolumes of the three algorithms, MOEA/WST, NSGA-II and ParEGO, for different generations and sensor budgets $p$.
\begin{table}[H]
\caption{Comparing hypervolume of MOEA/WST against those of the other two approaches (values are $\times 10^9$) and with respect to different budgets $p$ and number of generations. Statistical significance has been investigated through a Wilcoxon test ($p$-value is reported).}
\begin{adjustbox}{max width=\textwidth}
\begin{tabular}{ccccccc}
\hline
$\pmb{p}$        & \textbf{Generations} & \textbf{MOEA/WST}                                          & \textbf{NSGA-II}                                           & \textbf{ParEGO}                                                             & \textbf{\begin{tabular}[c]{@{}c@{}}MOEA/WST \\ vs NSGA-II\\ p-value\end{tabular}} & \textbf{\begin{tabular}[c]{@{}c@{}}MOEA/WST \\ vs ParEGO\\ p-value\end{tabular}} \\ \hline \hline
\multirow{4}{*}{3}  & 25                   & \begin{tabular}[c]{@{}c@{}}0.2188 \\ (0.3790)\end{tabular} & \begin{tabular}[c]{@{}c@{}}0.2190 \\ (0.4538)\end{tabular} & \multirow{4}{*}{\begin{tabular}[c]{@{}c@{}}0.2496 \\ (0.1641)\end{tabular}} & 0.811                                                                             & \textless{}0.001                                                                 \\
                    & 50                   & \begin{tabular}[c]{@{}c@{}}1.0828 \\ (0.5942)\end{tabular} & \begin{tabular}[c]{@{}c@{}}\textbf{1.5883} \\ \textbf{(0.2969)}\end{tabular} &                                                                             & \textless{}0.001                                                                  & \textless{}0.001                                                                 \\
                    & 100                  & \begin{tabular}[c]{@{}c@{}}2.0769 \\ (0.2225)\end{tabular} & \begin{tabular}[c]{@{}c@{}}\textbf{2.2808} \\ \textbf{(0.1601)}\end{tabular} &                                                                             & \textless{}0.001                                                                  & \textless{}0.001                                                                 \\ \hline
\multirow{4}{*}{7}  & 25                   & \begin{tabular}[c]{@{}c@{}}\textbf{2.0421} \\ \textbf{(0.135)}\end{tabular}  & \begin{tabular}[c]{@{}c@{}}1.9517 \\ (0.1217)\end{tabular} & \multirow{4}{*}{\begin{tabular}[c]{@{}c@{}}2.3189 \\ (0.0911)\end{tabular}} & \textless{}0.001                                                                  & \textless{}0.001                                                                 \\
                    & 50                   & \begin{tabular}[c]{@{}c@{}}2.2290 \\ (0.0608)\end{tabular} & \begin{tabular}[c]{@{}c@{}}2.1649 \\ (0.0825)\end{tabular} &                                                                             & 0.001                                                                             & \textless{}0.001                                                                 \\
                    & 100                  & \begin{tabular}[c]{@{}c@{}}2.3145\\ (0.0414)\end{tabular}  & \begin{tabular}[c]{@{}c@{}}2.3328 \\ (0.0891)\end{tabular} &                                                                             & 0.686                                                                             & 0.820                                                                            \\ \hline
\multirow{4}{*}{9}  & 25                   & \begin{tabular}[c]{@{}c@{}}\textbf{2.2084}\\ \textbf{(0.0608)}\end{tabular}  & \begin{tabular}[c]{@{}c@{}}2.0404 \\ (0.1079)\end{tabular} & \multirow{4}{*}{\begin{tabular}[c]{@{}c@{}}2.3171 \\ (0.0880)\end{tabular}} & \textless{}0.001                                                                  & \textless{}0.001                                                                 \\
                    & 50                   & \begin{tabular}[c]{@{}c@{}}\textbf{2.3189} \\ \textbf{(0.0521)}\end{tabular} & \begin{tabular}[c]{@{}c@{}}2.2350 \\ (0.0721)\end{tabular} &                                                                             & \textless{}0.001                                                                  & 0.959                                                                            \\
                    & 100                  & \begin{tabular}[c]{@{}c@{}}2.3880 \\ (0.0389)\end{tabular} & \begin{tabular}[c]{@{}c@{}}2.3517 \\ (0.0529)\end{tabular} &                                                                             & 0.006                                                                             & \textless{}0.001                                                                 \\ \hline
\multirow{4}{*}{15} & 25                   & \begin{tabular}[c]{@{}c@{}}\textbf{2.2891} \\ \textbf{(0.0560)}\end{tabular} & \begin{tabular}[c]{@{}c@{}}2.1830 \\ (0.0715)\end{tabular} & \multirow{4}{*}{\begin{tabular}[c]{@{}c@{}}2.3101 \\ (0.0920)\end{tabular}} & \textless{}0.001                                                                  & 0.398                                                                            \\
                    & 50                   & \begin{tabular}[c]{@{}c@{}}\textbf{2.3827} \\ \textbf{(0.0456)}\end{tabular} & \begin{tabular}[c]{@{}c@{}}2.3170\\ (0.0612)\end{tabular}  &                                                                             & \textless{}0.001                                                                  & \textless{}0.001                                                                 \\
                    & 100                  & \begin{tabular}[c]{@{}c@{}}2.4317 \\ (0.0415)\end{tabular} & \begin{tabular}[c]{@{}c@{}}2.4364 \\ (0.0444)\end{tabular} &                                                                             & 0.552                                                                             & \textless{}0.001                                                                 \\ \hline
\multirow{4}{*}{20} & 25                   & \begin{tabular}[c]{@{}c@{}}2.2880 \\ (0.0489)\end{tabular} & \begin{tabular}[c]{@{}c@{}}2.2393 \\ (0.0605)\end{tabular} & \multirow{4}{*}{\begin{tabular}[c]{@{}c@{}}2.3827 \\ (0.0892)\end{tabular}} & 0.001                                                                             & \textless{}0.001                                                                 \\
                    & 50                   & \begin{tabular}[c]{@{}c@{}}2.3605 \\ (0.0433)\end{tabular} & \begin{tabular}[c]{@{}c@{}}2.3610 \\ (0.0590)\end{tabular} &                                                                             & 0.809                                                                             & 0.248                                                                            \\
                    & 100                  & \begin{tabular}[c]{@{}c@{}}2.4186 \\ (0.0366)\end{tabular} & \begin{tabular}[c]{@{}c@{}}2.4415 \\ (0.0441)\end{tabular} &                                                                             & 0.031                                                                             & 0.001                                                                            \\ \hline
\end{tabular}
\end{adjustbox}
\label{tab:apx_wilcoxon}
\end{table}

\end{appendices}

%% file: main.bbl
\begin{thebibliography}{91}
\providecommand{\natexlab}[1]{#1}
\providecommand{\url}[1]{\texttt{#1}}
\expandafter\ifx\csname urlstyle\endcsname\relax
  \providecommand{\doi}[1]{doi: #1}\else
  \providecommand{\doi}{doi: \begingroup \urlstyle{rm}\Url}\fi

\bibitem[Caruana(1997)]{DBLP:journals/ml/Caruana97}
Rich Caruana.
\newblock Multitask learning.
\newblock \emph{Mach. Learn.}, 28\penalty0 (1):\penalty0 41--75, 1997.
\newblock \doi{10.1023/A:1007379606734}.
\newblock URL \url{https://doi.org/10.1023/A:1007379606734}.

\bibitem[Agarwal et~al.(2019)Agarwal, Dud{\'{\i}}k, and
  Wu]{DBLP:conf/icml/AgarwalDW19}
Alekh Agarwal, Miroslav Dud{\'{\i}}k, and Zhiwei~Steven Wu.
\newblock Fair regression: Quantitative definitions and reduction-based
  algorithms.
\newblock In Kamalika Chaudhuri and Ruslan Salakhutdinov, editors,
  \emph{Proceedings of the 36th International Conference on Machine Learning,
  {ICML} 2019, 9-15 June 2019, Long Beach, California, {USA}}, volume~97 of
  \emph{Proceedings of Machine Learning Research}, pages 120--129. {PMLR},
  2019.
\newblock URL \url{http://proceedings.mlr.press/v97/agarwal19d.html}.

\bibitem[Strubell et~al.(2019)Strubell, Ganesh, and
  McCallum]{DBLP:conf/acl/StrubellGM19}
Emma Strubell, Ananya Ganesh, and Andrew McCallum.
\newblock Energy and policy considerations for deep learning in {NLP}.
\newblock In Anna Korhonen, David~R. Traum, and Llu{\'{\i}}s M{\`{a}}rquez,
  editors, \emph{Proceedings of the 57th Conference of the Association for
  Computational Linguistics, {ACL} 2019, Florence, Italy, July 28- August 2,
  2019, Volume 1: Long Papers}, pages 3645--3650. Association for Computational
  Linguistics, 2019.
\newblock \doi{10.18653/v1/p19-1355}.
\newblock URL \url{https://doi.org/10.18653/v1/p19-1355}.

\bibitem[Candelieri et~al.(2021{\natexlab{a}})Candelieri, Perego, and
  Archetti]{DBLP:journals/soco/CandelieriPA21}
Antonio Candelieri, Riccardo Perego, and Francesco Archetti.
\newblock Green machine learning via augmented gaussian processes and
  multi-information source optimization.
\newblock \emph{Soft Comput.}, 25\penalty0 (19):\penalty0 12591--12603,
  2021{\natexlab{a}}.
\newblock \doi{10.1007/s00500-021-05684-7}.
\newblock URL \url{https://doi.org/10.1007/s00500-021-05684-7}.

\bibitem[Candelieri et~al.(2021{\natexlab{b}})Candelieri, Ponti, and
  Archetti]{DBLP:conf/gecco/CandelieriPA21}
Antonio Candelieri, Andrea Ponti, and Francesco Archetti.
\newblock Risk aware optimization of water sensor placement.
\newblock In Krzysztof Krawiec, editor, \emph{{GECCO} '21: Genetic and
  Evolutionary Computation Conference, Companion Volume, Lille, France, July
  10-14, 2021}, pages 295--296. {ACM}, 2021{\natexlab{b}}.
\newblock \doi{10.1145/3449726.3459477}.
\newblock URL \url{https://doi.org/10.1145/3449726.3459477}.

\bibitem[Perego et~al.(2020)Perego, Candelieri, Archetti, and
  Pau]{DBLP:conf/icann/PeregoCAP20}
Riccardo Perego, Antonio Candelieri, Francesco Archetti, and Danilo Pau.
\newblock Tuning deep neural network's hyperparameters constrained to
  deployability on tiny systems.
\newblock In Igor Farkas, Paolo Masulli, and Stefan Wermter, editors,
  \emph{Artificial Neural Networks and Machine Learning - {ICANN} 2020 - 29th
  International Conference on Artificial Neural Networks, Bratislava, Slovakia,
  September 15-18, 2020, Proceedings, Part {II}}, volume 12397 of \emph{Lecture
  Notes in Computer Science}, pages 92--103. Springer, 2020.
\newblock \doi{10.1007/978-3-030-61616-8\_8}.
\newblock URL \url{https://doi.org/10.1007/978-3-030-61616-8\_8}.

\bibitem[Loni et~al.(2020)Loni, Sinaei, Zoljodi, Daneshtalab, and
  Sj{\"{o}}din]{DBLP:journals/mam/LoniSZDS20}
Mohammad Loni, Sima Sinaei, Ali Zoljodi, Masoud Daneshtalab, and Mikael
  Sj{\"{o}}din.
\newblock Deepmaker: {A} multi-objective optimization framework for deep neural
  networks in embedded systems.
\newblock \emph{Microprocess. Microsystems}, 73:\penalty0 102989, 2020.
\newblock \doi{10.1016/j.micpro.2020.102989}.
\newblock URL \url{https://doi.org/10.1016/j.micpro.2020.102989}.

\bibitem[Sener and Koltun(2018)]{DBLP:conf/nips/SenerK18}
Ozan Sener and Vladlen Koltun.
\newblock Multi-task learning as multi-objective optimization.
\newblock In Samy Bengio, Hanna~M. Wallach, Hugo Larochelle, Kristen Grauman,
  Nicol{\`{o}} Cesa{-}Bianchi, and Roman Garnett, editors, \emph{Advances in
  Neural Information Processing Systems 31: Annual Conference on Neural
  Information Processing Systems 2018, NeurIPS 2018, December 3-8, 2018,
  Montr{\'{e}}al, Canada}, pages 525--536, 2018.
\newblock URL
  \url{https://proceedings.neurips.cc/paper/2018/hash/432aca3a1e345e339f35a30c8f65edce-Abstract.html}.

\bibitem[Lin et~al.(2019)Lin, Zhen, Li, Zhang, and
  Kwong]{DBLP:conf/nips/LinZ0ZK19}
Xi~Lin, Hui{-}Ling Zhen, Zhenhua Li, Qingfu Zhang, and Sam Kwong.
\newblock Pareto multi-task learning.
\newblock In Hanna~M. Wallach, Hugo Larochelle, Alina Beygelzimer, Florence
  d'Alch{\'{e}}{-}Buc, Emily~B. Fox, and Roman Garnett, editors, \emph{Advances
  in Neural Information Processing Systems 32: Annual Conference on Neural
  Information Processing Systems 2019, NeurIPS 2019, December 8-14, 2019,
  Vancouver, BC, Canada}, pages 12037--12047, 2019.
\newblock URL
  \url{https://proceedings.neurips.cc/paper/2019/hash/685bfde03eb646c27ed565881917c71c-Abstract.html}.

\bibitem[Zuluaga et~al.(2013)Zuluaga, Sergent, Krause, and
  P{\"{u}}schel]{DBLP:conf/icml/ZuluagaSKP13}
Marcela Zuluaga, Guillaume Sergent, Andreas Krause, and Markus P{\"{u}}schel.
\newblock Active learning for multi-objective optimization.
\newblock In \emph{Proceedings of the 30th International Conference on Machine
  Learning, {ICML} 2013, Atlanta, GA, USA, 16-21 June 2013}, volume~28 of
  \emph{{JMLR} Workshop and Conference Proceedings}, pages 462--470. JMLR.org,
  2013.
\newblock URL \url{http://proceedings.mlr.press/v28/zuluaga13.html}.

\bibitem[Zuluaga et~al.(2016)Zuluaga, Krause, and
  P{\"{u}}schel]{DBLP:journals/jmlr/ZuluagaKP16}
Marcela Zuluaga, Andreas Krause, and Markus P{\"{u}}schel.
\newblock e-pal: An active learning approach to the multi-objective
  optimization problem.
\newblock \emph{J. Mach. Learn. Res.}, 17:\penalty0 104:1--104:32, 2016.
\newblock URL \url{http://jmlr.org/papers/v17/15-047.html}.

\bibitem[Shah and Ghahramani(2016)]{DBLP:conf/icml/ShahG16}
Amar Shah and Zoubin Ghahramani.
\newblock Pareto frontier learning with expensive correlated objectives.
\newblock In Maria{-}Florina Balcan and Kilian~Q. Weinberger, editors,
  \emph{Proceedings of the 33nd International Conference on Machine Learning,
  {ICML} 2016, New York City, NY, USA, June 19-24, 2016}, volume~48 of
  \emph{{JMLR} Workshop and Conference Proceedings}, pages 1919--1927.
  JMLR.org, 2016.
\newblock URL \url{http://proceedings.mlr.press/v48/shahc16.html}.

\bibitem[Lin et~al.(2020)Lin, Yang, Zhang, and
  Kwong]{DBLP:journals/corr/abs-2010-06313}
Xi~Lin, Zhiyuan Yang, Qingfu Zhang, and Sam Kwong.
\newblock Controllable pareto multi-task learning.
\newblock \emph{CoRR}, abs/2010.06313, 2020.
\newblock URL \url{https://arxiv.org/abs/2010.06313}.

\bibitem[Archetti and Candelieri(2019)]{archetti2019bayesian}
Francesco Archetti and Antonio Candelieri.
\newblock \emph{Bayesian optimization and data science}.
\newblock Springer, 2019.

\bibitem[Frazier(2018)]{DBLP:journals/corr/abs-1807-02811}
Peter~I. Frazier.
\newblock A tutorial on bayesian optimization.
\newblock \emph{CoRR}, abs/1807.02811, 2018.
\newblock URL \url{http://arxiv.org/abs/1807.02811}.

\bibitem[Belakaria et~al.(2019)Belakaria, Deshwal, and
  Doppa]{DBLP:conf/nips/BelakariaDD19}
Syrine Belakaria, Aryan Deshwal, and Janardhan~Rao Doppa.
\newblock Max-value entropy search for multi-objective bayesian optimization.
\newblock In Hanna~M. Wallach, Hugo Larochelle, Alina Beygelzimer, Florence
  d'Alch{\'{e}}{-}Buc, Emily~B. Fox, and Roman Garnett, editors, \emph{Advances
  in Neural Information Processing Systems 32: Annual Conference on Neural
  Information Processing Systems 2019, NeurIPS 2019, December 8-14, 2019,
  Vancouver, BC, Canada}, pages 7823--7833, 2019.
\newblock URL
  \url{https://proceedings.neurips.cc/paper/2019/hash/82edc5c9e21035674d481640448049f3-Abstract.html}.

\bibitem[Garrido{-}Merch{\'{a}}n and
  Hern{\'{a}}ndez{-}Lobato(2019)]{DBLP:journals/ijon/Garrido-Merchan19}
Eduardo~C. Garrido{-}Merch{\'{a}}n and Daniel Hern{\'{a}}ndez{-}Lobato.
\newblock Predictive entropy search for multi-objective bayesian optimization
  with constraints.
\newblock \emph{Neurocomputing}, 361:\penalty0 50--68, 2019.
\newblock \doi{10.1016/j.neucom.2019.06.025}.
\newblock URL \url{https://doi.org/10.1016/j.neucom.2019.06.025}.

\bibitem[Rahat et~al.(2017)Rahat, Everson, and
  Fieldsend]{DBLP:conf/gecco/RahatEF17}
Alma As{-}Aad~Mohammad Rahat, Richard~M. Everson, and Jonathan~E. Fieldsend.
\newblock Alternative infill strategies for expensive multi-objective
  optimisation.
\newblock In Peter A.~N. Bosman, editor, \emph{Proceedings of the Genetic and
  Evolutionary Computation Conference, {GECCO} 2017, Berlin, Germany, July
  15-19, 2017}, pages 873--880. {ACM}, 2017.
\newblock \doi{10.1145/3071178.3071276}.
\newblock URL \url{https://doi.org/10.1145/3071178.3071276}.

\bibitem[Knowles(2006)]{DBLP:journals/tec/Knowles06}
Joshua~D. Knowles.
\newblock Parego: a hybrid algorithm with on-line landscape approximation for
  expensive multiobjective optimization problems.
\newblock \emph{{IEEE} Trans. Evol. Comput.}, 10\penalty0 (1):\penalty0 50--66,
  2006.
\newblock \doi{10.1109/TEVC.2005.851274}.
\newblock URL \url{https://doi.org/10.1109/TEVC.2005.851274}.

\bibitem[Zhang et~al.(2010)Zhang, Liu, Tsang, and
  Virginas]{DBLP:journals/tec/ZhangLTV10}
Qingfu Zhang, Wudong Liu, Edward P.~K. Tsang, and Botond Virginas.
\newblock Expensive multiobjective optimization by {MOEA/D} with gaussian
  process model.
\newblock \emph{{IEEE} Trans. Evol. Comput.}, 14\penalty0 (3):\penalty0
  456--474, 2010.
\newblock \doi{10.1109/TEVC.2009.2033671}.
\newblock URL \url{https://doi.org/10.1109/TEVC.2009.2033671}.

\bibitem[Zhang et~al.(2019)Zhang, Sun, Liu, Zhang, and
  Zhang]{DBLP:journals/isci/ZhangSLZZ19}
Hu~Zhang, Jianyong Sun, Tonglin Liu, Ke~Zhang, and Qingfu Zhang.
\newblock Balancing exploration and exploitation in multiobjective evolutionary
  optimization.
\newblock \emph{Inf. Sci.}, 497:\penalty0 129--148, 2019.
\newblock \doi{10.1016/j.ins.2019.05.046}.
\newblock URL \url{https://doi.org/10.1016/j.ins.2019.05.046}.

\bibitem[Beume et~al.(2007)Beume, Naujoks, and
  Emmerich]{DBLP:journals/eor/BeumeNE07}
Nicola Beume, Boris Naujoks, and Michael T.~M. Emmerich.
\newblock {SMS-EMOA:} multiobjective selection based on dominated hypervolume.
\newblock \emph{Eur. J. Oper. Res.}, 181\penalty0 (3):\penalty0 1653--1669,
  2007.
\newblock \doi{10.1016/j.ejor.2006.08.008}.
\newblock URL \url{https://doi.org/10.1016/j.ejor.2006.08.008}.

\bibitem[Rodr{\'\i}guez~Villalobos and Coello~Coello(2012)]{rodriguez2012new}
Cynthia~A Rodr{\'\i}guez~Villalobos and Carlos~A Coello~Coello.
\newblock A new multi-objective evolutionary algorithm based on a performance
  assessment indicator.
\newblock In \emph{Proceedings of the 14th annual conference on Genetic and
  evolutionary computation}, pages 505--512, 2012.

\bibitem[Zhang and Li(2007)]{DBLP:journals/tec/ZhangL07}
Qingfu Zhang and Hui Li.
\newblock {MOEA/D:} {A} multiobjective evolutionary algorithm based on
  decomposition.
\newblock \emph{{IEEE} Trans. Evol. Comput.}, 11\penalty0 (6):\penalty0
  712--731, 2007.
\newblock \doi{10.1109/TEVC.2007.892759}.
\newblock URL \url{https://doi.org/10.1109/TEVC.2007.892759}.

\bibitem[Trivedi et~al.(2017)Trivedi, Srinivasan, Sanyal, and
  Ghosh]{DBLP:journals/tec/TrivediSSG17}
Anupam Trivedi, Dipti Srinivasan, Krishnendu Sanyal, and Abhiroop Ghosh.
\newblock A survey of multiobjective evolutionary algorithms based on
  decomposition.
\newblock \emph{{IEEE} Trans. Evol. Comput.}, 21\penalty0 (3):\penalty0
  440--462, 2017.
\newblock \doi{10.1109/TEVC.2016.2608507}.
\newblock URL \url{https://doi.org/10.1109/TEVC.2016.2608507}.

\bibitem[Li et~al.(2019)Li, Deb, Zhang, Suganthan, and
  Chen]{DBLP:journals/swevo/LiDZSC19}
Hui Li, Kalyanmoy Deb, Qingfu Zhang, Ponnuthurai~N. Suganthan, and Lei Chen.
\newblock Comparison between {MOEA/D} and {NSGA-III} on a set of novel many and
  multi-objective benchmark problems with challenging difficulties.
\newblock \emph{Swarm Evol. Comput.}, 46:\penalty0 104--117, 2019.
\newblock \doi{10.1016/j.swevo.2019.02.003}.
\newblock URL \url{https://doi.org/10.1016/j.swevo.2019.02.003}.

\bibitem[Sun et~al.(2019)Sun, Zhang, Zhou, Zhang, and
  Zhang]{DBLP:journals/swevo/SunZZZZ19}
Jianyong Sun, Hu~Zhang, Aimin Zhou, Qingfu Zhang, and Ke~Zhang.
\newblock A new learning-based adaptive multi-objective evolutionary algorithm.
\newblock \emph{Swarm Evol. Comput.}, 44:\penalty0 304--319, 2019.
\newblock \doi{10.1016/j.swevo.2018.04.009}.
\newblock URL \url{https://doi.org/10.1016/j.swevo.2018.04.009}.

\bibitem[Deb and Myburgh(2017)]{DBLP:journals/eor/DebM17}
Kalyanmoy Deb and Christie Myburgh.
\newblock A population-based fast algorithm for a billion-dimensional resource
  allocation problem with integer variables.
\newblock \emph{Eur. J. Oper. Res.}, 261\penalty0 (2):\penalty0 460--474, 2017.
\newblock \doi{10.1016/j.ejor.2017.02.015}.
\newblock URL \url{https://doi.org/10.1016/j.ejor.2017.02.015}.

\bibitem[Blank and Deb(2020{\natexlab{a}})]{DBLP:conf/cec/BlankD20}
Julian Blank and Kalyanmoy Deb.
\newblock A running performance metric and termination criterion for evaluating
  evolutionary multi- and many-objective optimization algorithms.
\newblock In \emph{{IEEE} Congress on Evolutionary Computation, {CEC} 2020,
  Glasgow, United Kingdom, July 19-24, 2020}, pages 1--8. {IEEE},
  2020{\natexlab{a}}.
\newblock \doi{10.1109/CEC48606.2020.9185546}.
\newblock URL \url{https://doi.org/10.1109/CEC48606.2020.9185546}.

\bibitem[Li and Yao(2019)]{DBLP:journals/csur/LiY19}
Miqing Li and Xin Yao.
\newblock Quality evaluation of solution sets in multiobjective optimisation:
  {A} survey.
\newblock \emph{{ACM} Comput. Surv.}, 52\penalty0 (2):\penalty0 26:1--26:38,
  2019.
\newblock \doi{10.1145/3300148}.
\newblock URL \url{https://doi.org/10.1145/3300148}.

\bibitem[Jones et~al.(1998)Jones, Schonlau, and
  Welch]{DBLP:journals/jgo/JonesSW98}
Donald~R. Jones, Matthias Schonlau, and William~J. Welch.
\newblock Efficient global optimization of expensive black-box functions.
\newblock \emph{J. Glob. Optim.}, 13\penalty0 (4):\penalty0 455--492, 1998.
\newblock \doi{10.1023/A:1008306431147}.
\newblock URL \url{https://doi.org/10.1023/A:1008306431147}.

\bibitem[Emmerich et~al.(2006)Emmerich, Giannakoglou, and
  Naujoks]{DBLP:journals/tec/EmmerichGN06}
Michael T.~M. Emmerich, Kyriakos~C. Giannakoglou, and Boris Naujoks.
\newblock Single- and multiobjective evolutionary optimization assisted by
  gaussian random field metamodels.
\newblock \emph{{IEEE} Trans. Evol. Comput.}, 10\penalty0 (4):\penalty0
  421--439, 2006.
\newblock \doi{10.1109/TEVC.2005.859463}.
\newblock URL \url{https://doi.org/10.1109/TEVC.2005.859463}.

\bibitem[Deb et~al.(2002)Deb, Agrawal, Pratap, and
  Meyarivan]{DBLP:journals/tec/DebAPM02}
Kalyanmoy Deb, Samir Agrawal, Amrit Pratap, and T.~Meyarivan.
\newblock A fast and elitist multiobjective genetic algorithm: {NSGA-II}.
\newblock \emph{{IEEE} Trans. Evol. Comput.}, 6\penalty0 (2):\penalty0
  182--197, 2002.
\newblock \doi{10.1109/4235.996017}.
\newblock URL \url{https://doi.org/10.1109/4235.996017}.

\bibitem[Boyd and Vandenberghe(2014)]{DBLP:books/cu/BV2014}
Stephen~P. Boyd and Lieven Vandenberghe.
\newblock \emph{Convex Optimization}.
\newblock Cambridge University Press, 2014.
\newblock ISBN 978-0-521-83378-3.
\newblock \doi{10.1017/CBO9780511804441}.
\newblock URL \url{https://web.stanford.edu/\%7Eboyd/cvxbook/}.

\bibitem[Fan et~al.(2019)Fan, Li, Cai, Huang, Fang, You, Mo, Wei, and
  Goodman]{DBLP:journals/soco/FanLCHFYMWG19}
Zhun Fan, Wenji Li, Xinye Cai, Han Huang, Yi~Fang, Yugen You, Jiajie Mo, Caimin
  Wei, and Erik~D. Goodman.
\newblock An improved epsilon constraint-handling method in {MOEA/D} for cmops
  with large infeasible regions.
\newblock \emph{Soft Comput.}, 23\penalty0 (23):\penalty0 12491--12510, 2019.
\newblock \doi{10.1007/s00500-019-03794-x}.
\newblock URL \url{https://doi.org/10.1007/s00500-019-03794-x}.

\bibitem[Rasmussen and Williams(2006)]{DBLP:books/lib/RasmussenW06}
Carl~Edward Rasmussen and Christopher K.~I. Williams.
\newblock \emph{Gaussian processes for machine learning}.
\newblock Adaptive computation and machine learning. {MIT} Press, 2006.
\newblock ISBN 026218253X.
\newblock URL \url{https://www.worldcat.org/oclc/61285753}.

\bibitem[Kushner(1964)]{kushner1964new}
Harold~J Kushner.
\newblock A new method of locating the maximum point of an arbitrary multipeak
  curve in the presence of noise.
\newblock 1964.

\bibitem[Mockus et~al.(1978)Mockus, Tiesis, and
  Zilinskas]{mockus1978application}
Jonas Mockus, Vytautas Tiesis, and Antanas Zilinskas.
\newblock The application of bayesian methods for seeking the extremum.
\newblock \emph{Towards global optimization}, 2\penalty0 (117-129):\penalty0 2,
  1978.

\bibitem[Astudillo and Frazier(2019)]{DBLP:conf/icml/AstudilloF19}
Raul Astudillo and Peter~I. Frazier.
\newblock Bayesian optimization of composite functions.
\newblock In Kamalika Chaudhuri and Ruslan Salakhutdinov, editors,
  \emph{Proceedings of the 36th International Conference on Machine Learning,
  {ICML} 2019, 9-15 June 2019, Long Beach, California, {USA}}, volume~97 of
  \emph{Proceedings of Machine Learning Research}, pages 354--363. {PMLR},
  2019.
\newblock URL \url{http://proceedings.mlr.press/v97/astudillo19a.html}.

\bibitem[Auer(2002)]{DBLP:journals/jmlr/Auer02}
Peter Auer.
\newblock Using confidence bounds for exploitation-exploration trade-offs.
\newblock \emph{J. Mach. Learn. Res.}, 3:\penalty0 397--422, 2002.
\newblock URL \url{http://jmlr.org/papers/v3/auer02a.html}.

\bibitem[Srinivas et~al.(2012)Srinivas, Krause, Kakade, and
  Seeger]{DBLP:journals/tit/SrinivasKKS12}
Niranjan Srinivas, Andreas Krause, Sham~M. Kakade, and Matthias~W. Seeger.
\newblock Information-theoretic regret bounds for gaussian process optimization
  in the bandit setting.
\newblock \emph{{IEEE} Trans. Inf. Theory}, 58\penalty0 (5):\penalty0
  3250--3265, 2012.
\newblock \doi{10.1109/TIT.2011.2182033}.
\newblock URL \url{https://doi.org/10.1109/TIT.2011.2182033}.

\bibitem[Swersky et~al.(2013)Swersky, Snoek, and
  Adams]{DBLP:conf/nips/SwerskySA13}
Kevin Swersky, Jasper Snoek, and Ryan~Prescott Adams.
\newblock Multi-task bayesian optimization.
\newblock In Christopher J.~C. Burges, L{\'{e}}on Bottou, Zoubin Ghahramani,
  and Kilian~Q. Weinberger, editors, \emph{Advances in Neural Information
  Processing Systems 26: 27th Annual Conference on Neural Information
  Processing Systems 2013. Proceedings of a meeting held December 5-8, 2013,
  Lake Tahoe, Nevada, United States}, pages 2004--2012, 2013.
\newblock URL
  \url{https://proceedings.neurips.cc/paper/2013/hash/f33ba15effa5c10e873bf3842afb46a6-Abstract.html}.

\bibitem[Bardenet et~al.(2013)Bardenet, Brendel, K{\'{e}}gl, and
  Sebag]{DBLP:conf/icml/BardenetBKS13}
R{\'{e}}mi Bardenet, M{\'{a}}ty{\'{a}}s Brendel, Bal{\'{a}}zs K{\'{e}}gl, and
  Mich{\`{e}}le Sebag.
\newblock Collaborative hyperparameter tuning.
\newblock In \emph{Proceedings of the 30th International Conference on Machine
  Learning, {ICML} 2013, Atlanta, GA, USA, 16-21 June 2013}, volume~28 of
  \emph{{JMLR} Workshop and Conference Proceedings}, pages 199--207. JMLR.org,
  2013.
\newblock URL \url{http://proceedings.mlr.press/v28/bardenet13.html}.

\bibitem[Gelbart et~al.(2014)Gelbart, Snoek, and
  Adams]{DBLP:conf/uai/GelbartSA14}
Michael~A. Gelbart, Jasper Snoek, and Ryan~P. Adams.
\newblock Bayesian optimization with unknown constraints.
\newblock In Nevin~L. Zhang and Jin Tian, editors, \emph{Proceedings of the
  Thirtieth Conference on Uncertainty in Artificial Intelligence, {UAI} 2014,
  Quebec City, Quebec, Canada, July 23-27, 2014}, pages 250--259. {AUAI} Press,
  2014.
\newblock URL
  \url{https://dslpitt.org/uai/displayArticleDetails.jsp?mmnu=1\&smnu=2\&article\_id=2460\&proceeding\_id=30}.

\bibitem[Belakaria et~al.(2020{\natexlab{a}})Belakaria, Deshwal, and
  Doppa]{DBLP:conf/aaai/BelakariaDD20}
Syrine Belakaria, Aryan Deshwal, and Janardhan~Rao Doppa.
\newblock Multi-fidelity multi-objective bayesian optimization: An output space
  entropy search approach.
\newblock In \emph{The Thirty-Fourth {AAAI} Conference on Artificial
  Intelligence, {AAAI} 2020, The Thirty-Second Innovative Applications of
  Artificial Intelligence Conference, {IAAI} 2020, The Tenth {AAAI} Symposium
  on Educational Advances in Artificial Intelligence, {EAAI} 2020, New York,
  NY, USA, February 7-12, 2020}, pages 10035--10043. {AAAI} Press,
  2020{\natexlab{a}}.
\newblock URL \url{https://aaai.org/ojs/index.php/AAAI/article/view/6560}.

\bibitem[Belakaria et~al.(2020{\natexlab{b}})Belakaria, Deshwal, Jayakodi, and
  Doppa]{DBLP:conf/aaai/BelakariaDJD20}
Syrine Belakaria, Aryan Deshwal, Nitthilan~Kannappan Jayakodi, and
  Janardhan~Rao Doppa.
\newblock Uncertainty-aware search framework for multi-objective bayesian
  optimization.
\newblock In \emph{The Thirty-Fourth {AAAI} Conference on Artificial
  Intelligence, {AAAI} 2020, The Thirty-Second Innovative Applications of
  Artificial Intelligence Conference, {IAAI} 2020, The Tenth {AAAI} Symposium
  on Educational Advances in Artificial Intelligence, {EAAI} 2020, New York,
  NY, USA, February 7-12, 2020}, pages 10044--10052. {AAAI} Press,
  2020{\natexlab{b}}.
\newblock URL \url{https://aaai.org/ojs/index.php/AAAI/article/view/6561}.

\bibitem[Suzuki et~al.(2020)Suzuki, Takeno, Tamura, Shitara, and
  Karasuyama]{DBLP:conf/icml/SuzukiTTSK20}
Shinya Suzuki, Shion Takeno, Tomoyuki Tamura, Kazuki Shitara, and Masayuki
  Karasuyama.
\newblock Multi-objective bayesian optimization using pareto-frontier entropy.
\newblock In \emph{Proceedings of the 37th International Conference on Machine
  Learning, {ICML} 2020, 13-18 July 2020, Virtual Event}, volume 119 of
  \emph{Proceedings of Machine Learning Research}, pages 9279--9288. {PMLR},
  2020.
\newblock URL \url{http://proceedings.mlr.press/v119/suzuki20a.html}.

\bibitem[Ishibuchi et~al.(2018)Ishibuchi, Imada, Masuyama, and
  Nojima]{DBLP:conf/cec/IshibuchiIMN18}
Hisao Ishibuchi, Ryo Imada, Naoki Masuyama, and Yusuke Nojima.
\newblock Dynamic specification of a reference point for hypervolume
  calculation in {SMS-EMOA}.
\newblock In \emph{2018 {IEEE} Congress on Evolutionary Computation, {CEC}
  2018, Rio de Janeiro, Brazil, July 8-13, 2018}, pages 1--8. {IEEE}, 2018.
\newblock \doi{10.1109/CEC.2018.8477903}.
\newblock URL \url{https://doi.org/10.1109/CEC.2018.8477903}.

\bibitem[Ponti et~al.(2021{\natexlab{a}})Ponti, Candelieri, and
  Archetti]{ponti2021new}
Andrea Ponti, Antonio Candelieri, and Francesco Archetti.
\newblock A new evolutionary approach to optimal sensor placement in water
  distribution networks.
\newblock \emph{Water}, 13\penalty0 (12):\penalty0 1625, 2021{\natexlab{a}}.

\bibitem[Ponti et~al.(2021{\natexlab{b}})Ponti, Candelieri, and
  Archetti]{ponti2021wasserstein}
Andrea Ponti, Antonio Candelieri, and Francesco Archetti.
\newblock A wasserstein distance based multiobjective evolutionary algorithm
  for the risk aware optimization of sensor placement.
\newblock \emph{Intelligent Systems with Applications}, 10:\penalty0 200047,
  2021{\natexlab{b}}.

\bibitem[Leskovec et~al.(2007)Leskovec, Krause, Guestrin, Faloutsos,
  VanBriesen, and Glance]{DBLP:conf/kdd/LeskovecKGFVG07}
Jure Leskovec, Andreas Krause, Carlos Guestrin, Christos Faloutsos, Jeanne~M.
  VanBriesen, and Natalie~S. Glance.
\newblock Cost-effective outbreak detection in networks.
\newblock In Pavel Berkhin, Rich Caruana, and Xindong Wu, editors,
  \emph{Proceedings of the 13th {ACM} {SIGKDD} International Conference on
  Knowledge Discovery and Data Mining, San Jose, California, USA, August 12-15,
  2007}, pages 420--429. {ACM}, 2007.
\newblock \doi{10.1145/1281192.1281239}.
\newblock URL \url{https://doi.org/10.1145/1281192.1281239}.

\bibitem[Galuzzi et~al.(2020)Galuzzi, Giordani, Candelieri, Perego, and
  Archetti]{DBLP:journals/cms/GaluzziGCPA20}
Bruno~G. Galuzzi, Ilaria Giordani, Antonio Candelieri, Riccardo Perego, and
  Francesco Archetti.
\newblock Hyperparameter optimization for recommender systems through bayesian
  optimization.
\newblock \emph{Comput. Manag. Sci.}, 17\penalty0 (4):\penalty0 495--515, 2020.
\newblock \doi{10.1007/s10287-020-00376-3}.
\newblock URL \url{https://doi.org/10.1007/s10287-020-00376-3}.

\bibitem[Gillis et~al.(2019)Gillis, Hien, Leplat, and
  Tan]{DBLP:journals/corr/abs-1901-10757}
Nicolas Gillis, Le~Thi~Khanh Hien, Valentin Leplat, and Vincent Y.~F. Tan.
\newblock Distributionally robust and multi-objective nonnegative matrix
  factorization.
\newblock \emph{CoRR}, abs/1901.10757, 2019.
\newblock URL \url{http://arxiv.org/abs/1901.10757}.

\bibitem[Zhu and Honeine(2016)]{DBLP:journals/tgrs/ZhuH16}
Fei Zhu and Paul Honeine.
\newblock Biobjective nonnegative matrix factorization: Linear versus
  kernel-based models.
\newblock \emph{{IEEE} Trans. Geosci. Remote. Sens.}, 54\penalty0 (7):\penalty0
  4012--4022, 2016.
\newblock \doi{10.1109/TGRS.2016.2535298}.
\newblock URL \url{https://doi.org/10.1109/TGRS.2016.2535298}.

\bibitem[Lin et~al.(2018)Lin, Wang, Hu, Ma, Chen, Li, and
  Coello]{DBLP:journals/complexity/LinWHMCLC18}
Qiuzhen Lin, Xiaozhou Wang, Bishan Hu, Lijia Ma, Fei Chen, Jianqiang Li, and
  Carlos A.~Coello Coello.
\newblock Multiobjective personalized recommendation algorithm using extreme
  point guided evolutionary computation.
\newblock \emph{Complex.}, 2018:\penalty0 1716352:1--1716352:18, 2018.
\newblock \doi{10.1155/2018/1716352}.
\newblock URL \url{https://doi.org/10.1155/2018/1716352}.

\bibitem[Monge(1781)]{monge1781memoire}
Gaspard Monge.
\newblock M{\'e}moire sur la th{\'e}orie des d{\'e}blais et des remblais.
\newblock \emph{Histoire de l'Acad{\'e}mie Royale des Sciences de Paris}, 1781.

\bibitem[Kantorovitch(1958)]{custom:journals/Kantorovitch}
Lev Kantorovitch.
\newblock On the translocation of masses.
\newblock \emph{Management Science}, 1958.
\newblock \doi{10.1287/mnsc.5.1.1}.
\newblock URL \url{https://pubsonline.informs.org/doi/abs/10.1287/mnsc.5.1.1}.

\bibitem[Bonneel et~al.(2016)Bonneel, Peyr{\'{e}}, and
  Cuturi]{DBLP:journals/tog/BonneelPC16}
Nicolas Bonneel, Gabriel Peyr{\'{e}}, and Marco Cuturi.
\newblock Wasserstein barycentric coordinates: histogram regression using
  optimal transport.
\newblock \emph{{ACM} Trans. Graph.}, 35\penalty0 (4):\penalty0 71:1--71:10,
  2016.
\newblock \doi{10.1145/2897824.2925918}.
\newblock URL \url{https://doi.org/10.1145/2897824.2925918}.

\bibitem[Huang et~al.(2016)Huang, Guo, Kusner, Sun, Sha, and
  Weinberger]{DBLP:conf/nips/HuangGKSSW16}
Gao Huang, Chuan Guo, Matt~J. Kusner, Yu~Sun, Fei Sha, and Kilian~Q.
  Weinberger.
\newblock Supervised word mover's distance.
\newblock In Daniel~D. Lee, Masashi Sugiyama, Ulrike von Luxburg, Isabelle
  Guyon, and Roman Garnett, editors, \emph{Advances in Neural Information
  Processing Systems 29: Annual Conference on Neural Information Processing
  Systems 2016, December 5-10, 2016, Barcelona, Spain}, pages 4862--4870, 2016.
\newblock URL
  \url{https://proceedings.neurips.cc/paper/2016/hash/10c66082c124f8afe3df4886f5e516e0-Abstract.html}.

\bibitem[Arjovsky et~al.(2017)Arjovsky, Chintala, and
  Bottou]{DBLP:conf/icml/ArjovskyCB17}
Mart{\'{\i}}n Arjovsky, Soumith Chintala, and L{\'{e}}on Bottou.
\newblock Wasserstein generative adversarial networks.
\newblock In Doina Precup and Yee~Whye Teh, editors, \emph{Proceedings of the
  34th International Conference on Machine Learning, {ICML} 2017, Sydney, NSW,
  Australia, 6-11 August 2017}, volume~70 of \emph{Proceedings of Machine
  Learning Research}, pages 214--223. {PMLR}, 2017.
\newblock URL \url{http://proceedings.mlr.press/v70/arjovsky17a.html}.

\bibitem[Meng et~al.(2020)Meng, Yan, Liu, Wu, and
  Cheng]{DBLP:conf/um/MengYLWC20}
Yitong Meng, Xiao Yan, Weiwen Liu, Huanhuan Wu, and James Cheng.
\newblock Wasserstein collaborative filtering for item cold-start
  recommendation.
\newblock In Tsvi Kuflik, Ilaria Torre, Robin Burke, and Cristina Gena,
  editors, \emph{Proceedings of the 28th {ACM} Conference on User Modeling,
  Adaptation and Personalization, {UMAP} 2020, Genoa, Italy, July 12-18, 2020},
  pages 318--322. {ACM}, 2020.
\newblock \doi{10.1145/3340631.3394870}.
\newblock URL \url{https://doi.org/10.1145/3340631.3394870}.

\bibitem[Backurs et~al.(2020)Backurs, Dong, Indyk, Razenshteyn, and
  Wagner]{DBLP:conf/icml/BackursDIRW20}
Arturs Backurs, Yihe Dong, Piotr Indyk, Ilya~P. Razenshteyn, and Tal Wagner.
\newblock Scalable nearest neighbor search for optimal transport.
\newblock In \emph{Proceedings of the 37th International Conference on Machine
  Learning, {ICML} 2020, 13-18 July 2020, Virtual Event}, volume 119 of
  \emph{Proceedings of Machine Learning Research}, pages 497--506. {PMLR},
  2020.
\newblock URL \url{http://proceedings.mlr.press/v119/backurs20a.html}.

\bibitem[Chen et~al.(2021)Chen, Ma, Song, and
  Wang]{DBLP:journals/corr/abs-2102-03450}
Zhixian Chen, Tengfei Ma, Yangqiu Song, and Yang Wang.
\newblock Wasserstein diffusion on graphs with missing attributes.
\newblock \emph{CoRR}, abs/2102.03450, 2021.
\newblock URL \url{https://arxiv.org/abs/2102.03450}.

\bibitem[Kandasamy et~al.(2018)Kandasamy, Neiswanger, Schneider, P{\'{o}}czos,
  and Xing]{DBLP:conf/nips/KandasamyNSPX18}
Kirthevasan Kandasamy, Willie Neiswanger, Jeff Schneider, Barnab{\'{a}}s
  P{\'{o}}czos, and Eric~P. Xing.
\newblock Neural architecture search with bayesian optimisation and optimal
  transport.
\newblock In Samy Bengio, Hanna~M. Wallach, Hugo Larochelle, Kristen Grauman,
  Nicol{\`{o}} Cesa{-}Bianchi, and Roman Garnett, editors, \emph{Advances in
  Neural Information Processing Systems 31: Annual Conference on Neural
  Information Processing Systems 2018, NeurIPS 2018, December 3-8, 2018,
  Montr{\'{e}}al, Canada}, pages 2020--2029, 2018.
\newblock URL
  \url{https://proceedings.neurips.cc/paper/2018/hash/f33ba15effa5c10e873bf3842afb46a6-Abstract.html}.

\bibitem[Villani(2008)]{villani2008optimal}
C{\'e}dric Villani.
\newblock \emph{Optimal transport: old and new}, volume 338.
\newblock Springer Science \& Business Media, 2008.

\bibitem[Peyr{\'{e}} and Cuturi(2019)]{DBLP:journals/ftml/PeyreC19}
Gabriel Peyr{\'{e}} and Marco Cuturi.
\newblock Computational optimal transport.
\newblock \emph{Found. Trends Mach. Learn.}, 11\penalty0 (5-6):\penalty0
  355--607, 2019.
\newblock \doi{10.1561/2200000073}.
\newblock URL \url{https://doi.org/10.1561/2200000073}.

\bibitem[Li et~al.(2020)Li, Qian, Du, Zhao, and Zhang]{li2020collaborative}
Rui Li, Fulan Qian, Xiuquan Du, Shu Zhao, and Yanping Zhang.
\newblock A collaborative filtering recommendation framework based on
  wasserstein gan.
\newblock In \emph{Journal of Physics: Conference Series}, volume 1684, page
  012057. IOP Publishing, 2020.

\bibitem[Zhang et~al.(2020)Zhang, Li, Zhu, and
  Hu]{DBLP:journals/prl/ZhangLZH20}
Yuhong Zhang, Yuling Li, Yi~Zhu, and Xuegang Hu.
\newblock Wasserstein {GAN} based on autoencoder with back-translation for
  cross-lingual embedding mappings.
\newblock \emph{Pattern Recognit. Lett.}, 129:\penalty0 311--316, 2020.
\newblock \doi{10.1016/j.patrec.2019.11.033}.
\newblock URL \url{https://doi.org/10.1016/j.patrec.2019.11.033}.

\bibitem[{\"O}cal et~al.(2019){\"O}cal, Grima, and
  Sanguinetti]{ocal2019parameter}
Kaan {\"O}cal, Ramon Grima, and Guido Sanguinetti.
\newblock Parameter estimation for biochemical reaction networks using
  wasserstein distances.
\newblock \emph{Journal of Physics A: Mathematical and Theoretical},
  53\penalty0 (3):\penalty0 034002, 2019.

\bibitem[Indyk and Thaper(2003)]{indyk2003fast}
Piotr Indyk and Nitin Thaper.
\newblock Fast image retrieval via embeddings.
\newblock In \emph{3rd international workshop on statistical and computational
  theories of vision}, volume~2, page~5, 2003.

\bibitem[Atasu and Mittelholzer(2019)]{DBLP:conf/icml/AtasuM19}
Kubilay Atasu and Thomas Mittelholzer.
\newblock Linear-complexity data-parallel earth mover's distance
  approximations.
\newblock In Kamalika Chaudhuri and Ruslan Salakhutdinov, editors,
  \emph{Proceedings of the 36th International Conference on Machine Learning,
  {ICML} 2019, 9-15 June 2019, Long Beach, California, {USA}}, volume~97 of
  \emph{Proceedings of Machine Learning Research}, pages 364--373. {PMLR},
  2019.
\newblock URL \url{http://proceedings.mlr.press/v97/atasu19a.html}.

\bibitem[Fantozzi et~al.(2014)Fantozzi, Popescu, Farnham, Archetti, Mogre,
  Tsouchnika, Chiesa, Tsertou, Gama, and Bimpas]{fantozzi2014ict}
M~Fantozzi, I~Popescu, T~Farnham, F~Archetti, P~Mogre, E~Tsouchnika, C~Chiesa,
  A~Tsertou, M~Castro Gama, and M~Bimpas.
\newblock Ict for efficient water resources management: the icewater energy
  management and control approach.
\newblock \emph{Procedia Engineering}, 70:\penalty0 633--640, 2014.

\bibitem[Schieber et~al.(2017)Schieber, Carpi, D{\'\i}az-Guilera, Pardalos,
  Masoller, and Ravetti]{schieber2017quantification}
Tiago~A Schieber, Laura Carpi, Albert D{\'\i}az-Guilera, Panos~M Pardalos,
  Cristina Masoller, and Mart{\'\i}n~G Ravetti.
\newblock Quantification of network structural dissimilarities.
\newblock \emph{Nature communications}, 8\penalty0 (1):\penalty0 1--10, 2017.

\bibitem[Ponti et~al.(2021{\natexlab{c}})Ponti, Candelieri, Giordani, and
  Archetti]{ponti2021novel}
Andrea Ponti, Antonio Candelieri, Ilaria Giordani, and Francesco Archetti.
\newblock A novel graph-based vulnerability metric in urban network
  infrastructures: The case of water distribution networks.
\newblock \emph{Water}, 13\penalty0 (11):\penalty0 1502, 2021{\natexlab{c}}.

\bibitem[Klise et~al.(2018)Klise, Murray, and Haxton]{klise2018overview}
Katherine~A Klise, Regan Murray, and Terra Haxton.
\newblock An overview of the water network tool for resilience (wntr).
\newblock 2018.

\bibitem[Deb et~al.(2007)Deb, Sindhya, and Okabe]{DBLP:conf/gecco/DebSO07}
Kalyanmoy Deb, Karthik Sindhya, and Tatsuya Okabe.
\newblock Self-adaptive simulated binary crossover for real-parameter
  optimization.
\newblock In Hod Lipson, editor, \emph{Genetic and Evolutionary Computation
  Conference, {GECCO} 2007, Proceedings, London, England, UK, July 7-11, 2007},
  pages 1187--1194. {ACM}, 2007.
\newblock \doi{10.1145/1276958.1277190}.
\newblock URL \url{https://doi.org/10.1145/1276958.1277190}.

\bibitem[Tak{\'{a}}cs et~al.(2009)Tak{\'{a}}cs, Pil{\'{a}}szy, N{\'{e}}meth,
  and Tikk]{DBLP:journals/jmlr/TakacsPNT09}
G{\'{a}}bor Tak{\'{a}}cs, Istv{\'{a}}n Pil{\'{a}}szy, Botty{\'{a}}n
  N{\'{e}}meth, and Domonkos Tikk.
\newblock Scalable collaborative filtering approaches for large recommender
  systems.
\newblock \emph{J. Mach. Learn. Res.}, 10:\penalty0 623--656, 2009.
\newblock URL \url{https://dl.acm.org/citation.cfm?id=1577091}.

\bibitem[Harper and Konstan(2016)]{DBLP:journals/tiis/HarperK16}
F.~Maxwell Harper and Joseph~A. Konstan.
\newblock The movielens datasets: History and context.
\newblock \emph{{ACM} Trans. Interact. Intell. Syst.}, 5\penalty0 (4):\penalty0
  19:1--19:19, 2016.
\newblock \doi{10.1145/2827872}.
\newblock URL \url{https://doi.org/10.1145/2827872}.

\bibitem[Zuo et~al.(2015)Zuo, Gong, Zeng, Ma, and
  Jiao]{DBLP:journals/cim/ZuoGZMJ15}
Yi~Zuo, Maoguo Gong, Jiulin Zeng, Lijia Ma, and Licheng Jiao.
\newblock Personalized recommendation based on evolutionary multi-objective
  optimization [research frontier].
\newblock \emph{{IEEE} Comput. Intell. Mag.}, 10\penalty0 (1):\penalty0 52--62,
  2015.
\newblock \doi{10.1109/MCI.2014.2369894}.
\newblock URL \url{https://doi.org/10.1109/MCI.2014.2369894}.

\bibitem[Blank and Deb(2020{\natexlab{b}})]{DBLP:journals/access/BlankD20}
Julian Blank and Kalyanmoy Deb.
\newblock Pymoo: Multi-objective optimization in python.
\newblock \emph{{IEEE} Access}, 8:\penalty0 89497--89509, 2020{\natexlab{b}}.
\newblock \doi{10.1109/ACCESS.2020.2990567}.
\newblock URL \url{https://doi.org/10.1109/ACCESS.2020.2990567}.

\bibitem[Daulton et~al.(2020)Daulton, Balandat, and
  Bakshy]{DBLP:conf/nips/DaultonBB20}
Samuel Daulton, Maximilian Balandat, and Eytan Bakshy.
\newblock Differentiable expected hypervolume improvement for parallel
  multi-objective bayesian optimization.
\newblock In Hugo Larochelle, Marc'Aurelio Ranzato, Raia Hadsell,
  Maria{-}Florina Balcan, and Hsuan{-}Tien Lin, editors, \emph{Advances in
  Neural Information Processing Systems 33: Annual Conference on Neural
  Information Processing Systems 2020, NeurIPS 2020, December 6-12, 2020,
  virtual}, 2020.
\newblock URL
  \url{https://proceedings.neurips.cc/paper/2020/hash/6fec24eac8f18ed793f5eaad3dd7977c-Abstract.html}.

\bibitem[Balandat et~al.(2020)Balandat, Karrer, Jiang, Daulton, Letham, Wilson,
  and Bakshy]{DBLP:conf/nips/BalandatKJDLWB20}
Maximilian Balandat, Brian Karrer, Daniel~R. Jiang, Samuel Daulton, Benjamin
  Letham, Andrew~Gordon Wilson, and Eytan Bakshy.
\newblock Botorch: {A} framework for efficient monte-carlo bayesian
  optimization.
\newblock In Hugo Larochelle, Marc'Aurelio Ranzato, Raia Hadsell,
  Maria{-}Florina Balcan, and Hsuan{-}Tien Lin, editors, \emph{Advances in
  Neural Information Processing Systems 33: Annual Conference on Neural
  Information Processing Systems 2020, NeurIPS 2020, December 6-12, 2020,
  virtual}, 2020.
\newblock URL
  \url{https://proceedings.neurips.cc/paper/2020/hash/f5b1b89d98b7286673128a5fb112cb9a-Abstract.html}.

\bibitem[Virtanen et~al.(2020)Virtanen, Gommers, Oliphant, Haberland, Reddy,
  Cournapeau, Burovski, Peterson, Weckesser, Bright, {van der Walt}, Brett,
  Wilson, Millman, Mayorov, Nelson, Jones, Kern, Larson, Carey, Polat, Feng,
  Moore, {VanderPlas}, Laxalde, Perktold, Cimrman, Henriksen, Quintero, Harris,
  Archibald, Ribeiro, Pedregosa, {van Mulbregt}, and {SciPy 1.0
  Contributors}]{2020SciPy-NMeth}
Pauli Virtanen, Ralf Gommers, Travis~E. Oliphant, Matt Haberland, Tyler Reddy,
  David Cournapeau, Evgeni Burovski, Pearu Peterson, Warren Weckesser, Jonathan
  Bright, St{\'e}fan~J. {van der Walt}, Matthew Brett, Joshua Wilson, K.~Jarrod
  Millman, Nikolay Mayorov, Andrew R.~J. Nelson, Eric Jones, Robert Kern, Eric
  Larson, C~J Carey, {\.I}lhan Polat, Yu~Feng, Eric~W. Moore, Jake
  {VanderPlas}, Denis Laxalde, Josef Perktold, Robert Cimrman, Ian Henriksen,
  E.~A. Quintero, Charles~R. Harris, Anne~M. Archibald, Ant{\^o}nio~H. Ribeiro,
  Fabian Pedregosa, Paul {van Mulbregt}, and {SciPy 1.0 Contributors}.
\newblock {{SciPy} 1.0: Fundamental Algorithms for Scientific Computing in
  Python}.
\newblock \emph{Nature Methods}, 17:\penalty0 261--272, 2020.
\newblock \doi{10.1038/s41592-019-0686-2}.

\bibitem[Maria et~al.(2014)Maria, Boissonnat, Glisse, and
  Yvinec]{DBLP:conf/icms/MariaBGY14}
Cl{\'{e}}ment Maria, Jean{-}Daniel Boissonnat, Marc Glisse, and Mariette
  Yvinec.
\newblock The gudhi library: Simplicial complexes and persistent homology.
\newblock In Hoon Hong and Chee Yap, editors, \emph{Mathematical Software -
  {ICMS} 2014 - 4th International Congress, Seoul, South Korea, August 5-9,
  2014. Proceedings}, volume 8592 of \emph{Lecture Notes in Computer Science},
  pages 167--174. Springer, 2014.
\newblock \doi{10.1007/978-3-662-44199-2\_28}.
\newblock URL \url{https://doi.org/10.1007/978-3-662-44199-2\_28}.

\bibitem[Flamary et~al.(2021)Flamary, Courty, Gramfort, Alaya, Boisbunon,
  Chambon, Chapel, Corenflos, Fatras, Fournier, et~al.]{flamary2021pot}
R{\'e}mi Flamary, Nicolas Courty, Alexandre Gramfort, Mokhtar~Zahdi Alaya,
  Aur{\'e}lie Boisbunon, Stanislas Chambon, Laetitia Chapel, Adrien Corenflos,
  Kilian Fatras, Nemo Fournier, et~al.
\newblock Pot: Python optimal transport.
\newblock \emph{Journal of Machine Learning Research}, 22\penalty0
  (78):\penalty0 1--8, 2021.

\bibitem[Cuturi(2013)]{DBLP:conf/nips/Cuturi13}
Marco Cuturi.
\newblock Sinkhorn distances: Lightspeed computation of optimal transport.
\newblock In Christopher J.~C. Burges, L{\'{e}}on Bottou, Zoubin Ghahramani,
  and Kilian~Q. Weinberger, editors, \emph{Advances in Neural Information
  Processing Systems 26: 27th Annual Conference on Neural Information
  Processing Systems 2013. Proceedings of a meeting held December 5-8, 2013,
  Lake Tahoe, Nevada, United States}, pages 2292--2300, 2013.
\newblock URL
  \url{https://proceedings.neurips.cc/paper/2013/hash/af21d0c97db2e27e13572cbf59eb343d-Abstract.html}.

\bibitem[Bonneel et~al.(2015)Bonneel, Rabin, Peyr{\'{e}}, and
  Pfister]{DBLP:journals/jmiv/BonneelRPP15}
Nicolas Bonneel, Julien Rabin, Gabriel Peyr{\'{e}}, and Hanspeter Pfister.
\newblock Sliced and radon wasserstein barycenters of measures.
\newblock \emph{J. Math. Imaging Vis.}, 51\penalty0 (1):\penalty0 22--45, 2015.
\newblock \doi{10.1007/s10851-014-0506-3}.
\newblock URL \url{https://doi.org/10.1007/s10851-014-0506-3}.

\bibitem[Agueh and Carlier(2011)]{DBLP:journals/siamma/AguehC11}
Martial Agueh and Guillaume Carlier.
\newblock Barycenters in the wasserstein space.
\newblock \emph{{SIAM} J. Math. Anal.}, 43\penalty0 (2):\penalty0 904--924,
  2011.
\newblock \doi{10.1137/100805741}.
\newblock URL \url{https://doi.org/10.1137/100805741}.

\bibitem[Cuturi and Doucet(2014)]{DBLP:conf/icml/CuturiD14}
Marco Cuturi and Arnaud Doucet.
\newblock Fast computation of wasserstein barycenters.
\newblock In \emph{Proceedings of the 31th International Conference on Machine
  Learning, {ICML} 2014, Beijing, China, 21-26 June 2014}, volume~32 of
  \emph{{JMLR} Workshop and Conference Proceedings}, pages 685--693. JMLR.org,
  2014.
\newblock URL \url{http://proceedings.mlr.press/v32/cuturi14.html}.

\bibitem[Shannon et~al.(2003)Shannon, Markiel, Ozier, Baliga, Wang, Ramage,
  Amin, Schwikowski, and Ideker]{shannon2003cytoscape}
Paul Shannon, Andrew Markiel, Owen Ozier, Nitin~S Baliga, Jonathan~T Wang,
  Daniel Ramage, Nada Amin, Benno Schwikowski, and Trey Ideker.
\newblock Cytoscape: a software environment for integrated models of
  biomolecular interaction networks.
\newblock \emph{Genome research}, 13\penalty0 (11):\penalty0 2498--2504, 2003.

\bibitem[Morris et~al.(2011)Morris, Apeltsin, Newman, Baumbach, Wittkop, Su,
  Bader, and Ferrin]{DBLP:journals/bmcbi/MorrisANBWSBF11}
John~H. Morris, Leonard Apeltsin, Aaron~M. Newman, Jan Baumbach, Tobias
  Wittkop, Gang Su, Gary~D. Bader, and Thomas~E. Ferrin.
\newblock clustermaker: a multi-algorithm clustering plugin for cytoscape.
\newblock \emph{{BMC} Bioinform.}, 12:\penalty0 436, 2011.
\newblock \doi{10.1186/1471-2105-12-436}.
\newblock URL \url{https://doi.org/10.1186/1471-2105-12-436}.

\end{thebibliography}
